\documentclass[lettersize,journal]{IEEEtran}
\usepackage{amsmath,amsfonts}
\usepackage{algorithmic}
\usepackage{array}
\usepackage{textcomp}
\usepackage{stfloats}
\usepackage{url}
\usepackage{verbatim}
\usepackage{graphicx}
\def\BibTeX{{\rm B\kern-.05em{\sc i\kern-.025em b}\kern-.08em
    T\kern-.1667em\lower.7ex\hbox{E}\kern-.125emX}}
\usepackage{balance}
\usepackage{graphics}
\usepackage{subfigure}
\usepackage{booktabs}
\usepackage{multirow}
\usepackage{cite}
\usepackage{tikz,xcolor,hyperref}

\usepackage{algorithm}
\definecolor{lime}{HTML}{A6CE39}
\DeclareRobustCommand{\orcidicon}{%
	\begin{tikzpicture}
	\draw[lime, fill=lime] (0,0) 
	circle [radius=0.16] 
	node[white] {{\fontfamily{qag}\selectfont \tiny ID}};
	\draw[white, fill=white] (-0.0625,0.095) 
	circle [radius=0.007];
	\end{tikzpicture}
	\hspace{-2mm}
}

\foreach \x in {A, ..., Z}{%
	\expandafter\xdef\csname orcid\x\endcsname{\noexpand\href{https://orcid.org/\csname orcidauthor\x\endcsname}{\noexpand\orcidicon}}
}

\begin{document}
\title{Neuro-Planner: A 3D Visual Navigation Method for MAV with Depth Camera based on Neuromorphic Reinforcement Learning}
\author{Junjie Jiang\orcidA{}, Delei Kong\orcidB{}, Kuanxu Hou\orcidC{}, Xinjie Huang\orcidD{}, Hao Zhuang\orcidE{} and Zheng Fang\orcidF{}, \emph{Member, IEEE}
% , \\ Sonya Coleman\orcidG{}, \emph{Member, IEEE}, and Dermot Kerr\orcidH{}
\thanks{
Manuscript created December 23, 2021. This work was supported by National Natural Science Foundation of China (62073066, U20A20197), Intel Neuromorphic Research Community (NRC) Grant Award (RV2.137.Fang), Science and Technology on Near-Surface Detection Laboratory (6142414200208), the Fundamental Research Funds for the Central Universities (N2226001), and Aeronautical Science Foundation of China (No.201941050001). \emph{(Corresponding author: Zheng Fang.)}

Junjie Jiang, Kuanxu Hou, Xinjie Huang and Zheng Fang are with Faculty of Robot Science and Engineering, Northeastern University, Shenyang, China (e-mail: 2001998@stu.neu.edu.cn, 2001995@stu.neu.edu.cn, 2101979@stu.neu.edu.cn, fangzheng@mail.neu.edu.cn).

Delei Kong and Hao Zhuang are with College of Information Science and Engineering, Northeastern University, Shenyang, China (e-mail: kong.delei.neu@gmail.com, 2100922@stu.neu.edu.cn).

% Sonya Coleman and Dermot Kerr are with Faculty of Computing, Engineering and Built Environment, Ulster University, Northern Ireland, UK (e-mail: sa.coleman@ulster.ac.uk, d.kerr@ulster.ac.uk).
}}

\markboth{IEEE Robotics and Automation Letters, Vol. ?, No. ?, May 2022}%
{How to Use the IEEEtran \LaTeX \ Templates}

\maketitle
\begin{abstract}
Traditional visual navigation methods of micro aerial vehicle (MAV) usually calculate a passable path that satisfies the constraints depending on a prior map. However, these methods have issues such as high demand for computing resources and poor robustness in face of unfamiliar environments. Aiming to solve the above problems, we propose a neuromorphic reinforcement learning method (Neuro-Planner) that combines spiking neural network (SNN) and deep reinforcement learning (DRL) to realize MAV 3D visual navigation with depth camera. Specifically, we design spiking actor network based on two-state LIF (TS-LIF) neurons and its encoding-decoding schemes for efficient inference. Then 
our improved hybrid deep deterministic policy gradient (HDDPG) and TS-LIF-based spatio-temporal back propagation (STBP) algorithms are used as the training framework for actor-critic network architecture. To verify the effectiveness of the proposed Neuro-Planner, we carry out detailed comparison experiments with various SNN training algorithm (STBP, BPTT and SLAYER) in the software-in-the-loop (SITL) simulation framework. The navigation success rate of our HDDPG-STBP is 4.3\% and 5.3\% higher than that of the original DDPG in the two evaluation environments. To the best of our knowledge, this is the first work combining neuromorphic computing and deep reinforcement learning for MAV 3D visual navigation task.
\end{abstract}


\section{Introduction}
\label{sec:introduction}

\begin{figure}[htbp]
\vspace{-0mm}
\centering
\includegraphics[width=\columnwidth]{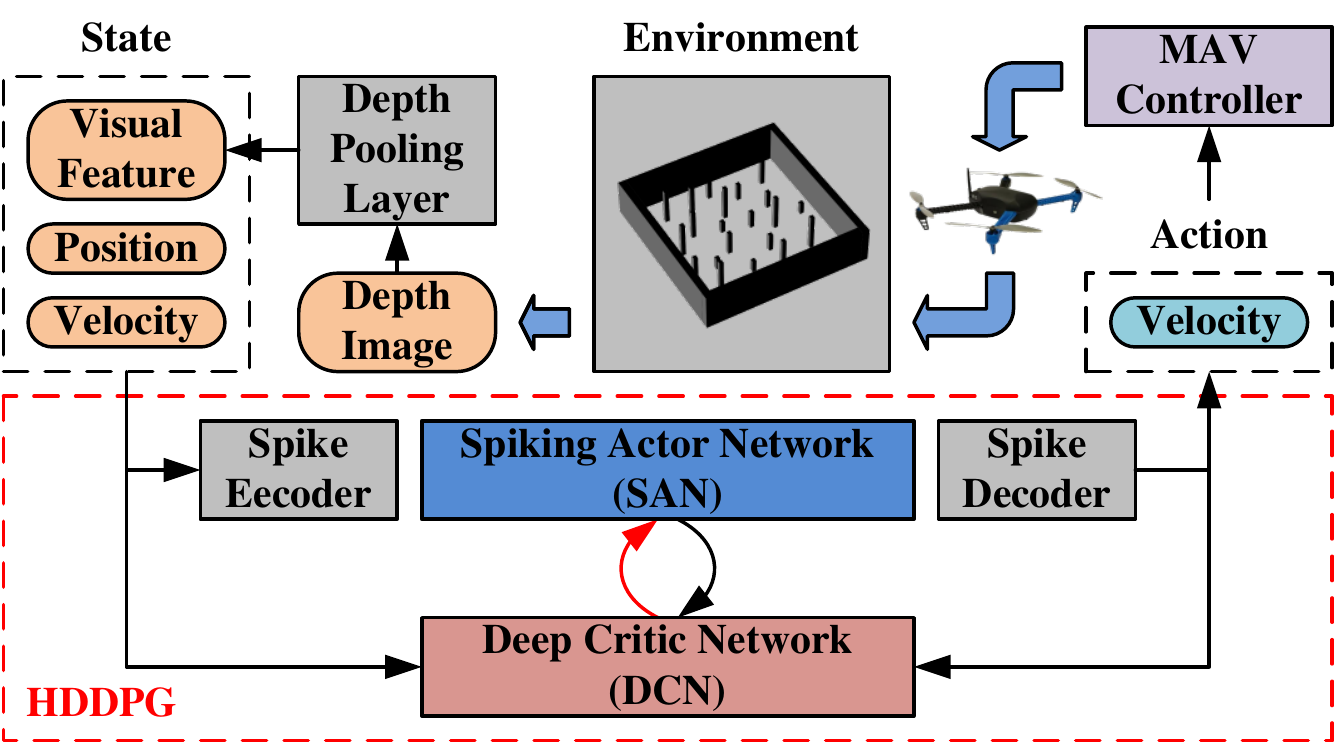}
\caption{Overview of the proposed Neuro-Planner architecture. In the training part, the hybrid deep deterministic policy gradient (HDDPG) algorithm is used to train the entire actor-critic network. In the testing part, only spiking actor network is used for inference to realize the MAV autonomous navigation.}
\label{fig:1}
\vspace{-0mm}
\end{figure}

\IEEEPARstart{A}{utonomous} navigation in GPS-denied environment is one of the most core issues in the field of mobile robots, especially for the autonomous navigation of micro aerial vehicle (MAV). Due to the limited load capacity of MAV, only lightweight sensor modules can be carried. Therefore, vision sensors with small size, lightweight, low power consumption and rich information are very suitable for MAV navigation \cite{lu2018survey} \cite{ross2013learning}. Visual autonomous navigation of MAV means that MAV system can perceive the surrounding environment and independently complete obstacle avoidance and navigation from the starting point to the goal point. At present, the main visual autonomous navigation framework of MAV usually relies on mapping technology. Firstly, a 3D map of the environment are constructed by visual sensors, then the path that can honor traversability constraints is calculated by global and local planners. Finally, the generated paths are tracked by controllers \cite{fang2017robust} \cite{lin2018autonomous} \cite{mohta2018fast} \cite{oleynikova2018complete} \cite{gao2019flying} \cite{zhou2020ego} \cite{zhou2022swarm}. In essence, this autonomous navigation paradigm calculates the passable path through certain constraints, which is difficult to adapt to the complex and changeable real environments. Therefore, it is still far behind the expected intelligence of humans.

In comparison, the processing mechanism of biological visual navigation system in nature is completely different from that of traditional autonomous navigation system. This is mainly reflected in the following two aspects: First, biological visual navigation system does not need to accurately calculate and model the ontology and the surrounding environment but uses the experience of interaction with environment previously to determine the current action. And it further improves the later action through the feedback of current action. Second, biological visual navigation system processes information hierarchically, parallelly, cyclically and asynchronously in a dynamic neural network composed of hundreds of millions of spiking neurons.

These phenomena arises great interest of researchers, and gives birth to two research fields: reinforcement learning (RL) \cite{silver2021reward} and spiking neural network (SNN) \cite{roy2019towards}. To gain experience as animals, RL is proposed. RL imitates the behavior paradigm of interaction between biology and environment. It makes agents interact with the environment continuously to gain experience to improve the navigation policy by setting reasonable reward function to realize the independent decision-making ability of agents. Autonomous navigation based on RL shows better success rate and robustness than traditional methods when facing unknown environments. To imitate the biological processing of brains, SNN is proposed. Spiking neurons in SNN execute asynchronously and independently, and thus have greater flexibility as well as the ability to learn temporal information. Compared with artificial neural network (ANN), SNN has advantages of more complex spatio-temporal dynamics, rich spiking coding mechanisms and high energy efficiency.

In this paper, we aim to combine deep reinforcement learning and spiking neural network to solve the visual autonomous navigation problem of MAV facing unknown environments \footnote{Supplementary Material: An accompanying video for this work is available at \url{https://youtu.be/P1GFOx9mWTU}.}. The main innovations and contributions of this paper are as follows:
\begin{itemize}
\item Aiming at the MAV 3D visual navigation task, a spiking actor network (SAN) based on two-state LIF (TS-LIF) spiking neurons and its encoding-decoding schemes (including state normalization, uniform encoding, rate decoding, and setting velocity mapping) are proposed. By processing the encoded state and visual observation, the action is obtained and decoded into control commands sent to low-level MAV controller.
\item We propose a hybrid deep deterministic policy gradient (HDDPG) reinforcement learning algorithm based on actor-critic networks (ACN), and improve spatio-temporal back propagation (STBP) for SAN training based on TS-LIF spiking neurons.
\item A software in-the-loop (SITL) simulation system of MAV is built based on ROS, Gazebo, PX4 and CUDA, and the simulation training process of MAV 3D visual navigation is designed. And we propose a method for judging whether the MAV pass the obstacles.
\item We compare and discuss the navigation performances of various spiking network training frameworks (STBP, BPTT, SLAYER) and various time steps in detail. The experiment results show that HDDPG-STBP has higher success rate than original DDPG. 
\end{itemize}

The rest of this paper is organized as follows: First, related work is presented in Section \ref{sec:relatedwork}. Then, the overall framework and pipeline of our algorithm are described in Section \ref{sec:methodology}, where the simulation system, HDDPG algorithm, and network training process are introduced in detail respectively. Next, the experimental results in the simulation environment are presented in Section \ref{sec:experiments}.
Finally, Section \ref{sec:conclusions} summarizes the paper. 

\section{Related Work}
\label{sec:relatedwork}
Reinforcement learning aims to learn a policy through the interaction and feedback between agent and environment. By combining deep network with reinforcement learning, deep reinforcement learning algorithm arises. For example, the deep Q network (DQN) proposed by Mnil et al. \cite{mnih2015human} and the deep deterministic policy gradient (DDPG) proposed by Lillicrap et al. \cite{lillicrap2015continuous}. In recent years, it has become popular to apply deep reinforcement learning to visual autonomous navigation task of mobile robots. For instance, Tai et al. \cite{tai2016robot} firstly deployed DQN on wheeled robot based on a RGB-D camera and controlled the robot to explore autonomously in a simulation corridor environment by discrete actions. Aiming at the problems of overestimation of Q value and slow convergence of training in DQN, Xie et al. \cite{xie2017towards} proposed dueling double DQN (D3QN) algorithm, which realized twice accelerated training, and deployed it to an actual wheeled robot to realize autonomous exploration. Since DQN and its variants can not output continuous control variables, it is difficult to achieve higher control accuracy. Tai et al. \cite{tai2017virtual} further proposed asynchronous DDPG algorithm, which uses 2D laser sensors to realize point-to-point autonomous navigation on virtual and actual wheeled robots. To collect training data faster, a separate thread is used for data collection. Furthermore, Mirowski et al. \cite{mirowski2016learning} used asynchronous advantage actor-critic (A3C) algorithm \cite{mnih2016asynchronous} as a training framework to realize the navigation of agents carrying a RGB camera in a 3D maze. They added two auxiliary tasks (deep prediction and closed-loop prediction) and took the loss of auxiliary tasks as additional supervision signals, which significantly improved the data efficiency and task performance when training agents. In addition, unlike point-to-point autonomous navigation tasks, Zhu et al. \cite{zhu2017target} proposed a target-driven visual deep reinforcement learning method. They extract features from the input target image and the image of the robot's view, projecting them to the same embedded space, and complete autonomous navigation by distinguishing the spatial relationship between the current state and the target.

All the above works were realized on wheeled robots, and later some works deployed on unmanned aerial vehicles appeared. Grando et al. \cite{grando2020deep} implemented mapless autonomous navigation of MAV using DDPG and soft actor-critic (SAC) \cite{haarnoja2018soft} respectively. However, due to the use of 2D laser sensors, the MAV's movement is limited to 2D plane. They subsequently introduced recurrent neural network into the twin delayed deep deterministic policy gradient (TD3) algorithm to realize map-free navigation and obstacle avoidance of MAV \cite{grando2022double}. Wang et al. \cite{wang2019autonomous} formulates the MAV navigation problem as partially observable Markov decision process, and proposed fast recursive deterministic policy gradient (Fast-RDPG) algorithm, which successfully enabled MAV to complete autonomous navigation at a fixed speed and altitude. In order to solve the sparse reward problem in reinforcement learning, they introduced a prior policy and achieved better navigation performance than using sparse reward directly for training \cite{wang2020deep}. He et al. \cite{he2020explainable} used a depth camera combined with DDPG to realize autonomous navigation of MAV in 3D environment, and the success rate achieved about 75\% in simulation environments, and also verified the effectiveness of its method in real environments. Lee et al. \cite{lee2021deep} introduced hindsight experience replay mechanism \cite{andrychowicz2017hindsight} into SAC algorithm, adding reward feedback of state, action and target to MAV during training, which improved the learning speed and success rate. In this paper, we use neuromorphic reinforcement learning method to realize the visual autonomous navigation of MAV in the 3D environment for the first time, and achieve a comparable success rate. 

% In recent years, spiking neural networks with rich dynamic characteristics have attracted more and more researchers' attention in the field of robot learning. For example, Tang et al. \cite{tang2020reinforcement} combined SNN and DDPG algorithm for the first time, enabling wheeled robot with 2D laser sensor can stably learn navigation policy and achieved a success rate comparable to artificial neural network method in simulation and real environment. At the same time, compared with DDPG algorithm deployed on Jetson TX2, SDDPG algorithm deployed on Loihi achieves more than 75\% energy saving. This proves that the combination of deep reinforcement learning and spiking neural network has the advantage of high energy efficiency in robot autonomous navigation task.

\section{Methodology}
\label{sec:methodology}
In this section, we will describe the network architecture of the proposed neuromorphic reinforcement learning method for MAV 3D autonomous navigation (Neuro-Planner) in detail, including the various parts of our framework, procedures of network training, and components of simulation system.

\subsection{Problem and Pipeline}
\subsubsection{Problem Formulation}
This paper mainly focuses on the mapless autonomous navigation and local obstacle avoidance of MAV from a specified starting point to a goal point in 3D structured environment, where MAV only relies on its onboard state estimation and visual observations. Therefore, the point-to-point 3D visual autonomous navigation problem of MAV can be formally defined as follows. In a specified 3D environment, the starting point $\boldsymbol{P}_\text{start}=\left[x_{0}, y_{0}, z_{0}\right]^{\top}$ and the goal point $\boldsymbol{P}_\text{goal}=\left[x_*,y_*,z_*\right]^{\top}$ are given. Knowing the current position $\boldsymbol{P}_k=\left[x_k,y_k,z_k\right]^{\top}$, current velocity $\boldsymbol{v}_k=\left[v_{\text{xy}, k}, v_{\text{yaw}, k}, v_{\text{z}, k}\right]^{\top}$ of the MAV and visual observation $\boldsymbol{D}_k$ of onboard depth camera in the current step $k$, we can solve for current setting velocity command  $\bar{\boldsymbol{v}}_k=\left[\bar{v}_{\text{xy}, k}, \bar{v}_{\text{yaw}, k}, \bar{v}_{\text{z}, k}\right]^{\top}$ of MAV. Among them, $v_{\text{xy}, k}, v_{\text{yaw}, k}, v_{\text{z}, k}$ are current horizontal linear velocity, horizontal angular velocity and vertical linear velocity of MAV respectively. MAV can reach the end point $\boldsymbol{P}_\text{end}=\left[x_\mathrm{K}, y_\mathrm{K}, z_\mathrm{K}\right]^{\top}$ near the specified goal point $\boldsymbol{P}_\text{goal}$ from the starting point $\boldsymbol{P}_\text{start}$ within a limited number of steps $\mathrm{K}$ ($k=1, \cdots, \mathrm{K}$) according to the actions, so that the Euclidean distance between the end point and the goal point is less than the setting threshold, that is, $d_\text{euclid}(\boldsymbol{P}_\text{end}, \boldsymbol{P}_\text{goal}) < \varepsilon_\text{goal,th}$. To this purpose, we need to design a navigation policy that infers the current action based on the current state to perform the task of 3D visual autonomous navigation. The specific algorithm pipeline is described as follows.

\subsubsection{Algorithm Pipeline} 
The proposed Neuro-Planner is combined of deep reinforcement learning and spiking neural network. The implementation process of the algorithm can be roughly divided into the following two parts:
\begin{itemize}
\item Hybrid Reinforcement Training. We propose a hybrid actor-critic network architecture for neuromorphic reinforcement learning for MAV 3D visual autonomous navigation. Here, we use a spiking neuron-based MLP network to form an actor network. Firstly, the current state (position, velocity and visual features) is normalized and spiking encoded, and then spike trains of the state are sent to the spiking actor network to obtain spike trains of the action, which are then decoded into the current action. In addition, a critic network (a deep MLP network) evaluates the current state and action. We use the hybrid deep deterministic policy gradient (HDDPG) algorithm to train the entire actor-critic networks.
\item Spiking Network Inference. We only use the trained spiking actor network for action inference. Firstly, we normalize and spike the real-time state, and feed the spike trains of the state  to the spiking neural network to get the spiking trains of the action in real time. We then decode it to the speed control command to realize 3D visual autonomous navigation of the MAV.
\end{itemize}

For systematic discussion, we first introduce the spiking network inference, followed by the hybrid reinforcement training.

\subsection{Spiking Network Inference}
We design a spiking actor network (SAN) $\boldsymbol{f}_\text{SAN}(\cdot)$, which infers the output spike trains from the input spike trains and decodes them into an output action to complete the mapping from the state space to the action space.

\begin{figure}[htbp]
\vspace{-0mm}
\centering
\includegraphics[width=0.8\columnwidth]{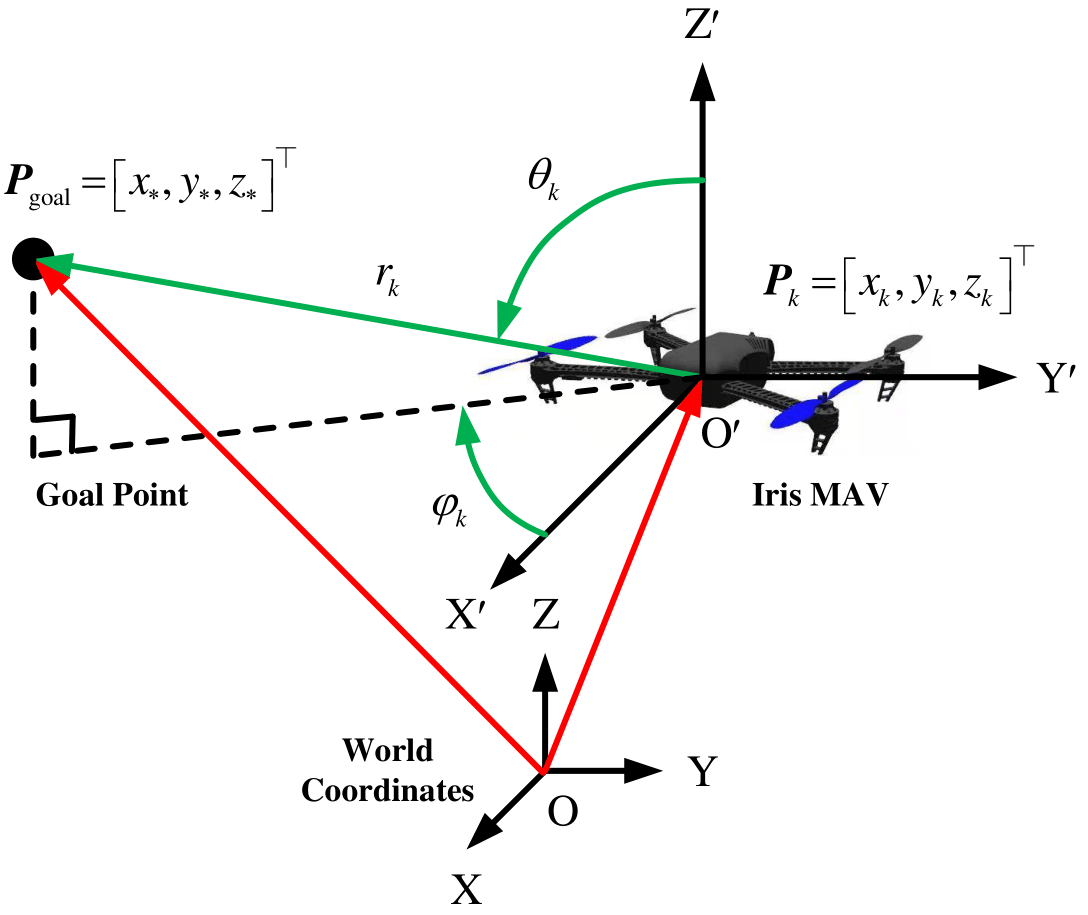}
\caption{Spherical coordinate system between the MAV and goal point.}
\label{fig:2}
\vspace{-0mm}
\end{figure}

\subsubsection{State and Action Selection}
For state space, the current state is composed of the relative position between MAV and goal point, the velocity of the MAV and the visual features of the depth camera. First, in order to express the relative positional relationship between the current position $\boldsymbol{P}_k=\left[x_k, y_k, z_k\right]^{\top}$ of the MAV and the goal point $\boldsymbol{P}_\text{goal}=\left[x_*,y_*,z_*\right]^{\top}$, we use the current position of the MAV $\boldsymbol{P}_k$ as the origin to establish a local spherical coordinate system to obtain the spherical coordinates of the goal point $\left[r_k, \theta_k, \varphi_k\right]^{\top}$, where $r_{k} \in \mathbb{R}^{+}$ is the radial distance, $\theta_{k} \in[0, \pi]$ is the polar angle, and $\varphi_{k} \in[-\pi, \pi]$ is the azimuth angle, as shown in Fig \ref{fig:2}. The x-y plane linear velocity and angular velocity of the MAV in the global coordinate system are $v_{\text{xy}, k}$ and $v_{\text{yaw}, k}$ respectively, and the z-axis linear velocity is $v_{\text{z}, k}$. In addition, since the depth image of the depth camera $\boldsymbol{D}_k$ is a high-dimensional tensor, it is necessary to extract visual features from it. We use the pooling operation $\boldsymbol{f}_\text{AP}(\cdot)$ to divide the depth map into several image patches and calculate the average depth of effective pixels in each patch as visual features: 
\begin{equation}
\label{eq:1}
\boldsymbol{d}_k=\boldsymbol{f}_{\text{AP}}\left(\boldsymbol{D}_k\right).
\end{equation} 
Compared with using the convolutional network to extract features, directly using pooling operation can effectively stabilize training and save memory. Therefore, the state at step $k$ can be summarized as $\boldsymbol{s}_k=\left[r_k, \theta_k, \varphi_k, v_{\text{xy}, k}, v_{\text{yaw}, k}, v_{\text{z}, k}, \boldsymbol{d}_k\right]^{\top}$. For the action space, we map the current action $\boldsymbol{a}_k$ output by the spiking actor network $\boldsymbol{f}_\text{SAN}(\boldsymbol{\phi})$ to the setting velocity $\bar{\boldsymbol{v}}_k = \left[\bar{v}_{\text{xy}, k}, \bar{v}_{\text{yaw}, k}, \bar{v}_{\text{z}, k}\right]^{\top}$, and use the speed control method to control the MAV.

\subsubsection{State Normalization and Uniform Encoding} Since the value range of each element in the current state $\boldsymbol{s}_k$ is different, it needs to be normalized. Here, for elements in the state with bipolarity, we use dual channels to represent its positive and negative polarities respectively (since we use unipolar spiking neurons). In this way, the normalized current state is expressed as $\tilde{\boldsymbol{s}}_k=[\tilde{r}_k, \tilde{\theta}_k, \tilde{\boldsymbol{\varphi}}_k, \tilde{v}_{\text{xy}, k}, \tilde{\boldsymbol{v}}_{\text{yaw}, k}, \tilde{\boldsymbol{v}}_{\text{z}, k}, \tilde{\boldsymbol{d}}_k]^{\top} \in[0,1]$, and the specific normalization for each element is as follows:
\begin{equation}
\label{eq:2}
\begin{aligned}
\tilde{r}_k&=\frac{r_\text{min}}{r_k},  \\
\tilde{\theta}_k&=\frac{\theta_k}{\pi}, \\
\tilde{\boldsymbol{\varphi}}_k&= \begin{cases}{\left[\frac{\left|\varphi_k\right|}{\pi}, 0\right]^{\top},} & \varphi_k \geq 0 \\ {\left[0, \frac{\left|\varphi_k\right|}{\pi}\right]^{\top},} & \varphi_k<0\end{cases},  \\
\tilde{v}_{\text{xy}, k}&=\frac{v_{\text{xy}, k}}{v_\text{xy-max}}, \\
\tilde{\boldsymbol{v}}_{\text{yaw}, k}&= \begin{cases}{\left[\frac{\left|v_{\text{yaw}, k}\right|}{v_\text{yaw-max}}, 0\right]^{\top},} & v_{\text{yaw}, k} \geq 0 \\ {\left[0, \frac{\left|v_{\text{yaw}, k}\right|}{v_\text{yaw-max}}\right]^{\top},} & v_{\text{yaw}, k}<0\end{cases}, \\
\tilde{\boldsymbol{v}}_{\text{z}, k}&= \begin{cases}{\left[\frac{\left|v_{\text{z}, k}\right|}{v_\text{z-max}}, 0\right]^{\top},} & v_{\text{z}, k} \geq 0 \\ {\left[0, \frac{\left|v_{\text{z}, k}\right|}{v_\text{z-max}}\right]^{\top},} & v_{\text{z}, k}<0\end{cases},    \\
\tilde{\boldsymbol{d}}_k&=\frac{d_\text{min}}{\boldsymbol{d}_k},
\end{aligned}
\end{equation}
where $r_\text{min}=0.3, v_\text{xy-max}=0.5, d_\text{min}=0.5, v_\text{yaw-max}=2, v_\text{z-max}=0.2$. The length of the entire normalized state $\tilde{\boldsymbol{s}}_k$ is $\mathrm{C}_{\tilde{\boldsymbol{s}}}=21$. In addition, since the spiking network deals with spike signals, we use uniform coding to encode the normalized state $\tilde{\boldsymbol{s}}_k$ into a set of spike trains. According to the time step $\mathrm{T}$, the generated spike array $\boldsymbol{O}_{\text{in}, k}$ is represented as follows: 
\begin{equation}
\label{eq:3}
\boldsymbol{O}_{\text{in}, k}^{t}[i]=\left\{\begin{array}{lc}
1, & \tilde{\boldsymbol{s}}_k[i]>\boldsymbol{M}_\text{rand}[i, t] \\
0, & \text{otherwise}
\end{array}\right.,
\end{equation} 
where $\boldsymbol{M}_\text{rand}$ is a uniformly distributed random matrix of $\mathrm{C}_{\tilde{\boldsymbol{s}}} \times \mathrm{T}$ size, whose element satisfies $\boldsymbol{M}_\text{rand}[i, t] \sim \mathrm{U}(0, 1)$. We denote the above state normalization and spike encoding procedures as
$\boldsymbol{O}_{\text{in}, k}=\boldsymbol{f}_\text{EC}\left(\boldsymbol{s}_k, \mathrm{T}\right) \doteq \boldsymbol{f}_\text{EC}\left(\boldsymbol{s}_k\right)$.

\subsubsection{Spiking Actor Network}
We process the encoded spike trains using the spiking actor network (SAN), which is a spiking fully connected network (SFCN) of $\mathrm{N}$ layers. Its input is a spike array $\boldsymbol{O}_{\text{in}, k}$ which is normalized and spike-encoded from the current state $\boldsymbol{s}_k$, and its output $\boldsymbol{O}_{\text{out}, k}$ is a spike array (decoded into the current action $\boldsymbol{a}_k$), expressed as follows:
\begin{equation}
\label{eq:4}
\boldsymbol{O}_{\text{out}, k}=\boldsymbol{f}_\text{SAN}\left(\boldsymbol{O}_{\text{in}, k} \mid \boldsymbol{\phi}\right).
\end{equation} 
For the specific implementation of each layer in the spiking actor network $\boldsymbol{f}_\text{SAN}(\boldsymbol{\phi})$, we use the two-state leaky integrate-and-fire (TS-LIF) spiking neuron model, which is a variant of the LIF neuron model \cite{maass1997networks}. The iterative update equation for the layer $n \in[1, \mathrm{N}]$ is given by:
\begin{equation}
\label{eq:5}
\begin{aligned} 
\boldsymbol{C}_{n}^{t} &=\delta_{\text{curr}, n} \boldsymbol{C}_{n}^{t-1} + \boldsymbol{W}_{n} \boldsymbol{O}_{n-1}^{t}+\boldsymbol{b}_{n}, \\
\boldsymbol{U}_{n}^{t} &=\delta_{\text{volt}, n} \boldsymbol{U}_{n}^{t-1} \boldsymbol{g}\left(\boldsymbol{O}_{n}^{t-1}\right)+\boldsymbol{C}_{n}^{t},    \\ 
\boldsymbol{O}_{n}^{t} &=\boldsymbol{h}\left(\boldsymbol{U}_{n}^{t}\right).
\end{aligned}
\end{equation}
where $t \in[1, \mathrm{T}]$ denotes the discrete time, $\boldsymbol{O}_{1}^{t}=\boldsymbol{O}_{\text{in}, k}^{t}$ is the input spike trains, $\boldsymbol{O}_{\text{out},k}^{t}=\boldsymbol{O}_\mathrm{N}^{t}$ is the output spike trains, $\boldsymbol{W}_n$ is the synaptic weight matrix, $\boldsymbol{b}_n$ is the synaptic bias vector. $\boldsymbol{C}_{n}^{t}$ and $\boldsymbol{U}_{n}^{t}$ are the membrane current and membrane voltage respectively. $\delta_{\text{curr}, n}$ and $\delta_{\text{volt}, n}$ are the decay coefficients of the current and voltage respectively. $\boldsymbol{g}\left(\boldsymbol{O}_{n}^{t-1}\right)=\mathbf{1}-\boldsymbol{O}_{n}^{t-1}$ is the reset gate. $\boldsymbol{h}\left(\boldsymbol{U}_{n}^{t}\right)=\boldsymbol{H}\left(\boldsymbol{U}_{n}^{t}-u_{\text{th}, n}\right)$ is the fire gate. $u_{\text{th}, n}$ is the spike-triggered threshold. This iterative update equation incorporates all behaviors (integration, firing, decay, and reset) of TS-LIF neurons. It can be seen that different from the activation functions of analog neurons such as ReLU, TS-LIF spiking neurons have obvious time dependencies. Finally, we use a spike count (SC) layer to decode the output spike array $\boldsymbol{O}_{\text{out}, k}$ to get the frequency of spikes as the action $\boldsymbol{a}_k$ of the SAN output:
\begin{equation}
\label{eq:6}
\boldsymbol{a}_k=\frac{1}{\mathrm{T}} \sum_{t=1}^{\mathrm{T}} \boldsymbol{O}_\text{out}^{t},
\end{equation}
where $\boldsymbol{a}_{k}[i] \in[0,1]$. We denote the above spike decoding as $\boldsymbol{a}_{k}=\boldsymbol{f}_\text{DC}\left(\boldsymbol{O}_{\text{out}, k}, \mathrm{T}\right) \doteq \boldsymbol{f}_\text{DC}\left(\boldsymbol{O}_{\text{out}, k}\right)$.
%Here, in our spiking actor network, we set the number of layers $\mathrm{L}=3$, the number of neurons in each layer $\mathrm{N}_l = 512$, the decay coefficients of the current and voltage states to be $d_{\text {current }}^{(l)} = 0.5 $and $d_{\text {volt }}^{(l)} = 0.75$, respectively, and the spiking trigger threshold $v_{\text {th }}^{(l)} = 0.5$. Input dimension $\mathrm{N}_{\tilde{s}}=21$, output dimension $\mathrm{N}_{\tilde{a}}=4$.

\begin{figure}[htbp]
\centering
\includegraphics[width=\columnwidth]{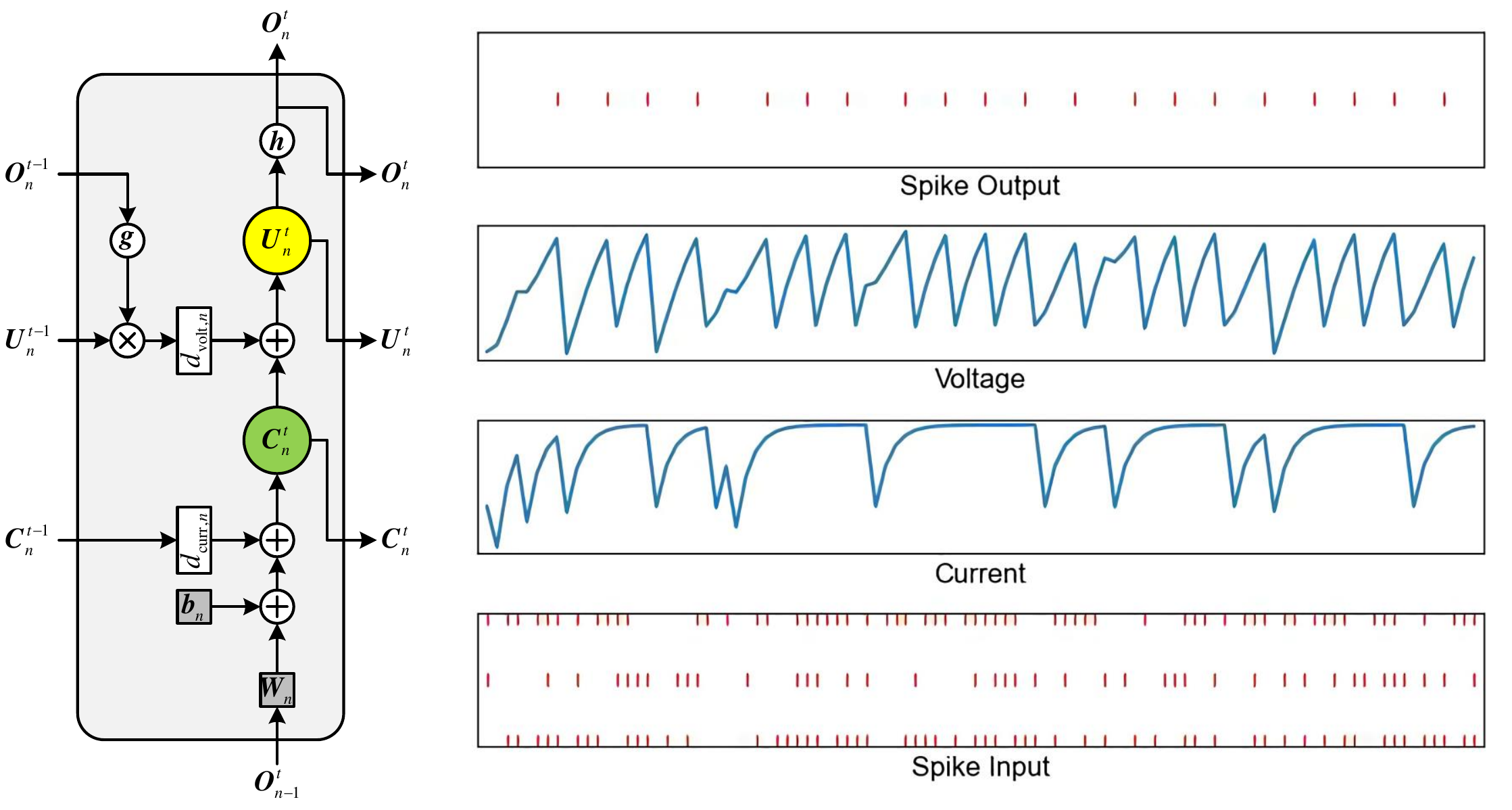}
\caption{Two-state leaky integrate-and-fire (TS-LIF) spiking neuron model. Different from analog neurons such as ReLU, since TS-LIF contains two states of current and voltage, it has rich spatio-temporal dynamics.}
\label{fig:3}
\vspace{-0mm}
\end{figure}

\subsubsection{Setting Velocity Mapping}
The current action $\boldsymbol{a}_{k}[i]\in[0,1]$ needs to be mapped to an appropriate setting velocity. Since the output spike trains are not distributed uniformly in the time step $\mathrm{T}$, in order to ensure that the training can converge, we need to make the horizontal angular velocity $\bar{v}_{\text{yaw}, k}$ and vertical linear velocity $\bar{v}_{\text{z}, k}$ distributed uniformly at the beginning of the training. Thus, we use the difference between two channels to calculate $\bar{v}_{\text{yaw}, k}$ and $\bar{v}_{\text{z}, k}$. Therefore, we use the current action $\boldsymbol{a}_{k}$ with four channels to solve for three setting velocities. We map $\boldsymbol{a}_{k}$ to get the setting velocities:
\begin{equation}
\label{eq:7}
\begin{aligned}
\bar{v}_{\text{xy},k} &=\alpha_{1} \cdot\left(\boldsymbol{a}_{k}[1]+\boldsymbol{a}_{k}[2]\right)+v_\text{min},  \\
\bar{v}_{\text{yaw},k} &=\alpha_{2} \cdot\left(\boldsymbol{a}_{k}[2]-\boldsymbol{a}_{k}[1]\right),  \\
\bar{v}_{\text{z},k} &=\alpha_{3} \cdot\left(\boldsymbol{a}_{k}[4]-\boldsymbol{a}_{k}[3]\right),
\end{aligned}
\end{equation}
where $\alpha_{1}=0.225$, $\alpha_{2}=1.8$ and $\alpha_{3}=0.18$. $v_\text{min}=0.05$ is the minimum horizontal linear velocity that enables the MAV to explore.

\begin{algorithm}[htbp]
\caption{SAN Forward Propagation}
\label{algo:1}
\renewcommand{\algorithmicrequire}{\textbf{Input:}}
\renewcommand{\algorithmicensure}{\textbf{Output:}}
\begin{algorithmic}[1]
\REQUIRE
State $\boldsymbol{s}_k$.   \\
Weights $\boldsymbol{W}_n$ and biases $\boldsymbol{b}_n$.   \\
Maximum time step $\mathrm{T}$.     \\
Network depth $\mathrm{N}$.     \\
Minimum plane linear velocity $v_\text{min}$.   \\
Velocity mapping factor $\alpha_{1},\alpha_{2},\alpha_{3}$.
\ENSURE
Setting velocity of MAV $\bar{\boldsymbol{v}}_k=\left[\bar{v}_{\text{xy}, k}, \bar{v}_{\text{yaw}, k}, \bar{v}_{\text{z}, k}\right]^{\top}$.
%Linear velocity in the $x\text{-}y$ plane $v_\text{xy-set}$, angular velocity in the $x\text{-}y$ plane $v_{\theta\text{-set}}$, linear velocity on the $z$-axis $v_\text{z-set}$.
\STATE Normalize and uniformly encode the state $\boldsymbol{s}_k$ to get the input spike train $\boldsymbol{O}_{\text{in}, k}=\boldsymbol{f}_\text{EC}\left(\boldsymbol{s}_{k}\right)$.
\STATE Propagate SAN with time $t$ and layer $n$ recurrently:
\FOR{$t=1,\cdots,\mathrm{T}$}
\STATE $\boldsymbol{O}_{1}^{t} = \boldsymbol{O}_{\text{in},k}^{t}$.
\FOR{$l=1,\cdots,\mathrm{L}$}
\STATE $\boldsymbol{C}_{n}^{t} = \delta_{\text{curr},n} \boldsymbol{C}_{n}^{t-1} + \boldsymbol{W}_{n} \boldsymbol{O}_{n-1}^{t} + \boldsymbol{b}_{n}$.
\STATE $\boldsymbol{U}_{n}^{t} = \delta_{\text{volt},n} \boldsymbol{U}_{n}^{t-1} g(\boldsymbol{O}_{n}^{t-1})+ \boldsymbol{C}_{n}^{t}$.
\STATE $\boldsymbol{O}_{n}^{t} = h(\boldsymbol{U}_{n}^{t})$.
\ENDFOR
\STATE $\boldsymbol{O}_{\text{out},k}^{t} = \boldsymbol{O}_\mathrm{N}^{t}$.
\ENDFOR
\STATE Decode the output spike train $\boldsymbol{O}_{\text{out},k}$ to get the action $\boldsymbol{a}_{k}=\boldsymbol{f}_\text{DC}\left(\boldsymbol{O}_{\text{out}, k}\right)$.
\STATE Map action $\boldsymbol{a}_k$ to actual setting velocity $\bar{\boldsymbol{v}}_k$:
\STATE $\bar{v}_{\text{xy},k} = \alpha_{1} \cdot (\boldsymbol{a}_{k}[1] + \boldsymbol{a}_{k}[2]) + v_\text{min}$.
\STATE $\bar{v}_{\text{yaw},k} = \alpha_{2} \cdot (\boldsymbol{a}_{k}[2] - \boldsymbol{a}_{k}[1])$.
\STATE $\bar{v}_{\text{z},k} = \alpha_{3} \cdot (\boldsymbol{a}_{k}[4] - \boldsymbol{a}_{k}[3])$.
\end{algorithmic}
\end{algorithm}

\subsection{Hybrid Reinforcement Training}
For the spiking actor network (SAN)  $\boldsymbol{f}_\text{SAN}(\boldsymbol{\phi})$ to learn an appropriate navigation policy, we employ the reinforcement learning paradigm.
% Reinforcement learning aims to learn a policy through the continuous interaction between the agent and environment. It can learn how to infer the optimal action according to the current state so that the agent can obtain autonomous decision-making ability.
In this paper, we propose a neuromorphic reinforcement learning method (HDDPG) based on the deep deterministic policy gradient (DDPG) algorithm \cite{lillicrap2015continuous} as the training framework.

\subsubsection{Hybrid Deep Deterministic Policy Gradient (HDDPG)}
\begin{figure}[htbp]
\centering
\subfigure[]{\label{fig:4a}\includegraphics[width=\columnwidth]{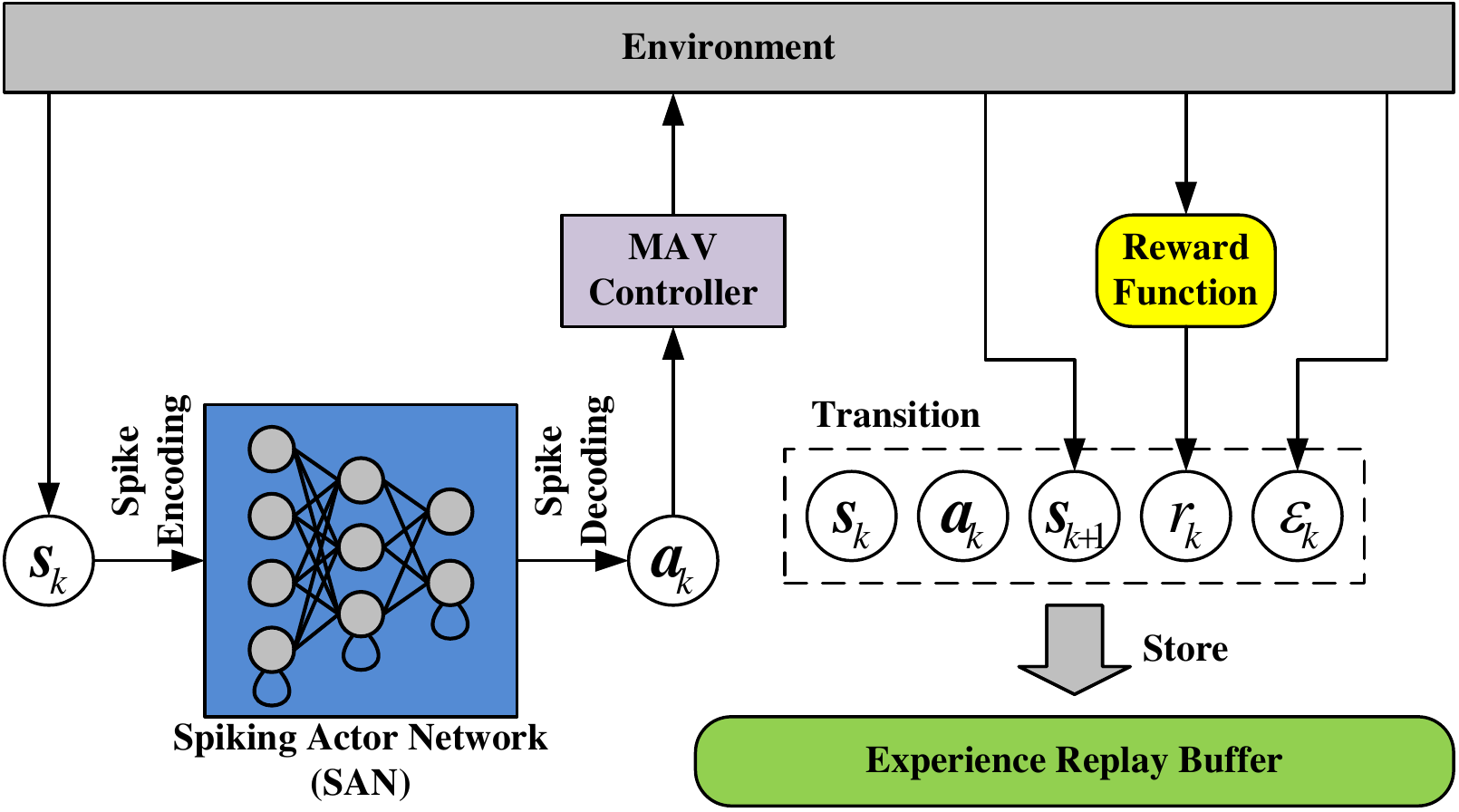}}
\subfigure[]{\label{fig:4b}\includegraphics[width=\columnwidth]{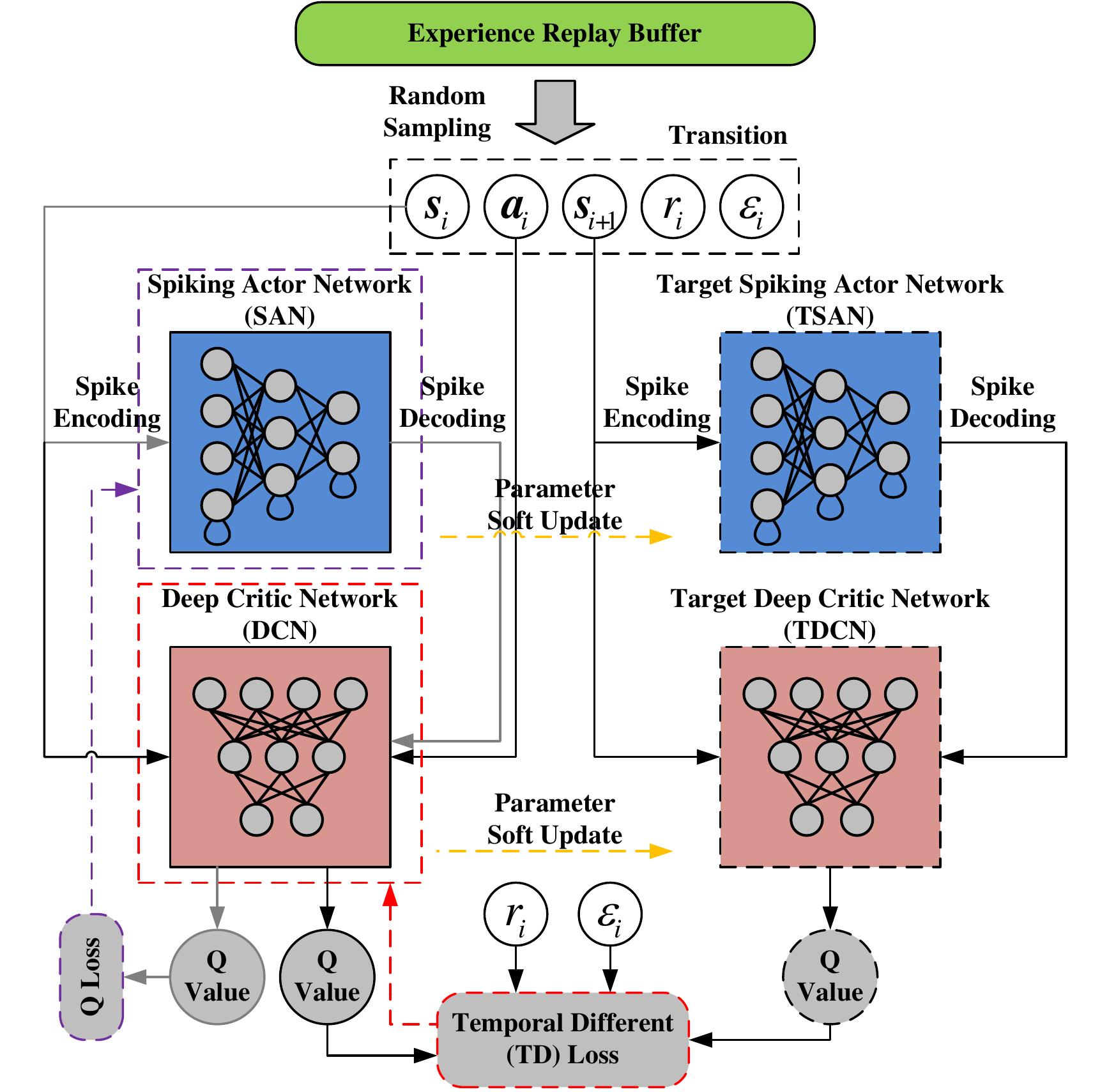}}
\caption{Overview of the proposed hybrid deep deterministic policy gradient (HDDPG) algorithm. (a) Experience replay mechanism \cite{mnih2015human}. (b) Temporal different (TD) loss and Q loss.}
\label{fig:4}
\vspace{-0mm}
\end{figure}

The DDPG algorithm is a deep reinforcement learning framework that can effectively optimize the expected return and estimate value function. On the basis of the DDPG \cite{lillicrap2015continuous} algorithm, our proposed hybrid deep deterministic policy gradient (HDDPG) algorithm combining the SNN dynamics and actor-critic network architecture, as shown in Fig. \ref{fig:4}. First, $\boldsymbol{f}_\text{SAN}(\boldsymbol{\phi})$ maps the current state $\boldsymbol{s}_k$ to the current action $\boldsymbol{a}_k(\boldsymbol{s}_k\mid\boldsymbol{\phi})$ with its encoding function $\boldsymbol{f}_\text{EC}(\cdot)$ and decoding function $\boldsymbol{f}_\text{DC}(\cdot)$. In order to improve the robustness of the policy, we add random noises $\kappa$ to the output of $\boldsymbol{f}_\text{SAN}(\boldsymbol{\phi})$ in the training phase, and get:
\begin{equation}
\label{eq:8}
\boldsymbol{a}_k=\boldsymbol{f}_\text{DC}\left(\boldsymbol{f}_\text{SAN}\left(\boldsymbol{f}_\text{EC}\left(\boldsymbol{s}_k\right) \mid \boldsymbol{\phi}\right)\right)+\kappa.
\end{equation}
Then, the deep critic network (DCN) $f_\text{DCN}(\boldsymbol{\psi})$ formed by the deep network evaluates $\boldsymbol{s}_k$ and $\boldsymbol{a}_k$ output by $\boldsymbol{f}_\text{SAN}(\boldsymbol{\phi})$, and obtains the Q value $Q_k$:
\begin{equation}
\label{eq:9}
Q_k=f_\text{DCN}(\boldsymbol{s}_k,\boldsymbol{a}_k \mid \boldsymbol{\psi}).
\end{equation}
In other words, $\boldsymbol{f}_\text{SAN}(\boldsymbol{\phi})$ aims to find appropriate $\boldsymbol{a}_k$, which maximizes $Q_k$ of $f_\text{DCN}(\boldsymbol{\psi})$. In addition, we adopt the experience replay mechanism \cite{mnih2015human} to improve the data utilization during training, as shown in Fig. \ref{fig:4a}. The current state $\boldsymbol{s}_k$, the current action $\boldsymbol{a}_k$, the next state $\boldsymbol{s}_{k+1}$, the reward $r_k$ from the state transition and the current episode end flag $\varepsilon_k$ are stored in the experience replay buffer as a transition, so that MAV can improve the current policy from the previous experience. At the same time, in order to make the networks converge stably, we use the target network mechanism \cite{mnih2015human} to set the conventional networks and the target networks respectively, as shown in Fig. \ref{fig:4b}. The parameters of the conventional networks are updated at each backpropagation. The parameters of target networks are frozen and only updated when the set number of backpropagations is reached. Here, target spiking actor network (TSAN) $\boldsymbol{f}_\text{TSAN}\left(\boldsymbol{\phi}^{\prime}\right)$ and the target deep critic network (TDCN) $f_\text{TDCN}\left(\boldsymbol{\psi}^{\prime}\right)$ are set. We sample $\mathrm{B}$ experiences $\{\boldsymbol{s}_i, \boldsymbol{a}_i, \boldsymbol{s}_{i+1}, r_i, \varepsilon_i\}$ from the experience replay buffer $\boldsymbol{R}$ for batch training. First,  $f_\text{DCN}(\boldsymbol{\psi})$, $f_\text{TDCN}\left(\boldsymbol{\psi}^{\prime}\right)$ and the reward $r_k$ are used to construct temporal-difference (TD) loss function, which is expressed as follows:
\begin{equation}
\label{eq:10}
\begin{aligned}
L_\text{TD}(\boldsymbol{\psi}) &=\left[r_{i}+\gamma Q_{i+1} \cdot\left(1-\varepsilon_{i}\right)-Q_{i}\right]^{2}, \\
Q_{i+1} &=f_\text{TDCN}\left(\boldsymbol{s}_{i+1}, \boldsymbol{a}_{i+1} \mid \boldsymbol{\psi}^{\prime}\right) \\
&=f_\text{TDCN}\left(\boldsymbol{s}_{i+1}, \boldsymbol{f}_\text{DC}\left(\boldsymbol{f}_\text{TSAN}\left(\boldsymbol{f}_\text{EC}\left(\boldsymbol{s}_{i+1}\right) \mid \boldsymbol{\phi}^{\prime}\right)\right) \mid \boldsymbol{\psi}^{\prime}\right),    \\
Q_{i} &=f_\text{DCN}\left(\boldsymbol{s}_{i}, \boldsymbol{a}_{i} \mid \boldsymbol{\psi}\right).
\end{aligned}
\end{equation}
where $\gamma$ is the discount factor. Then, the Q value can be directly used as its loss function for $\boldsymbol{f}_\text{SAN}(\boldsymbol{\phi})$, which is expressed as follows:
\begin{equation}
\label{eq:11}
\begin{aligned}
L_\text{Q}(\boldsymbol{\phi})&=-Q_{i}^{\prime}   \\
&=-f_\text{DCN}\left(\boldsymbol{s}_{i}, \boldsymbol{f}_\text{DC}\left(\boldsymbol{f}_\text{SAN}\left(\boldsymbol{f}_\text{EC}\left(\boldsymbol{s}_{i}\right)\right) \mid \boldsymbol{\phi}\right) \mid \boldsymbol{\psi}\right).
\end{aligned}
\end{equation}
In addition, the parameters of $\boldsymbol{f}_\text{TSAN}(\boldsymbol{\phi}^{\prime})$ and $f_\text{TDCN}(\boldsymbol{\psi}^{\prime})$ are periodically soft-updated by the parameters of $\boldsymbol{f}_\text{SAN}(\boldsymbol{\phi})$ and $f_\text{DCN}(\boldsymbol{\psi})$ to improve the stability of the learning process, which is specifically expressed as follows:
\begin{equation}
\label{eq:12}
\begin{aligned}
\boldsymbol{\phi}^{\prime}&=\eta \boldsymbol{\phi}+(1-\eta) \boldsymbol{\phi}^{\prime},  \\
\boldsymbol{\psi}^{\prime}&=\eta \boldsymbol{\psi}+(1-\eta) \boldsymbol{\psi}^{\prime},
\end{aligned}
\end{equation}
where is $\eta$ a sufficiently small constant. For the reward function $r_k$, its specific form is as follows:
\begin{equation}
\label{eq:13}
\begin{aligned}
r_{k}&=\left\{\begin{array}{cc}
r_\text{goal} & \text{if } d_\text{euclid}(\boldsymbol{P}_k, \boldsymbol{P}_\text{goal})<\varepsilon_\text{goal,th} \\
r_\text{obs} & \text{if } d_\text{euclid}(\boldsymbol{P}_k, \boldsymbol{P}_\text{obs})<\varepsilon_\text{obs,th}  \\
r_\text{other} & \text{otherwise}
\end{array}\right., \\
r_\text{other}&=\sigma \cdot\left(d_\text{euclid}(\boldsymbol{P}_k, \boldsymbol{P}_\text{goal}) - d_\text{euclid}(\boldsymbol{P}_{k-1}, \boldsymbol{P}_\text{goal})\right).
\end{aligned}
\end{equation}
where $r_\text{goal}$ and $r_\text{obstacle}$ are the positive and negative rewards respectively. $d_\text{euclid}(\boldsymbol{P}_k, \boldsymbol{P}_\text{goal})$ is the distance to the goal point, $d_\text{euclid}(\boldsymbol{P}_k, \boldsymbol{P}_\text{obs})$ is the distance to the nearest obstacle, $\sigma$ is the amplification factor, $\varepsilon_\text{goal,th}$ and $\varepsilon_\text{obs,th}$ are the distance threshold to the goal point and the distance threshold to the obstacle respectively. In a word, $r_k$ is designed to guide the MAV to move towards the goal and avoid obstacles during autonomous navigation.

\begin{algorithm}[htbp]
\caption{HDDPG}
\label{algo:2}
\renewcommand{\algorithmicrequire}{\textbf{Input:}}
\renewcommand{\algorithmicensure}{\textbf{Output:}}
\begin{algorithmic}[1]
\REQUIRE
Spiking actor network $\boldsymbol{f}_\text{SAN}(\boldsymbol{\phi})$.    \\
Deep critic network $f_\text{DCN}(\boldsymbol{\psi})$.    \\
Target network $\boldsymbol{f}_\text{TSAN}(\boldsymbol{\phi}^{\prime})$ and $f_\text{TDCN}(\boldsymbol{\psi}^{\prime})$.
\ENSURE
Temporal difference (TD) loss $L_\text{TD}(\boldsymbol{\psi})$ and Q loss $L_\text{Q}(\boldsymbol{\phi})$.
\STATE Randomly initialize spiking actor network $\boldsymbol{f}_\text{SAN}(\boldsymbol{\phi})$ and deep critic network $f_\text{DCN}(\boldsymbol{\psi})$ with weights $\boldsymbol{\phi}$ and $\boldsymbol{\psi}$. \\
\STATE Initialize target network $\boldsymbol{f}_\text{TSAN}(\boldsymbol{\phi}^{\prime})$ and $f_\text{TDCN}(\boldsymbol{\psi}^{\prime})$ with weights $\boldsymbol{\phi}^{\prime} \leftarrow \boldsymbol{\phi}$ and $\boldsymbol{\psi}^{\prime} \leftarrow \boldsymbol{\psi}$.   \\
\STATE Initialize replay buffer $\boldsymbol{R}$.
\FOR{$m=1,\cdots,\mathrm{M}$}
\STATE Receive initial pbservation state $\boldsymbol{s}_{0}$.
\FOR{$k=1,\cdots,\mathrm{K}$}
\STATE Select action $\boldsymbol{a}_k=\boldsymbol{f}_\text{DC}(\boldsymbol{f}_\text{SAN}(\boldsymbol{f}_\text{EC}(\boldsymbol{s}_k)|\boldsymbol{\phi}))+\kappa$ according to the current policy and exploration noise.
\STATE Execute action $\boldsymbol{a}_k$, and get reward $r_k$, new state $\boldsymbol{s}_{k+1}$ and epoch end bit $\varepsilon_k$.
\STATE Store current transition $[\boldsymbol{s}_k,\boldsymbol{a}_k,r_k,\boldsymbol{s}_{k+1},\varepsilon_k]^{\top}$ in experience replay buffer $\boldsymbol{R}$.
\STATE Sample a random minibatch of $\mathrm{N}$ transitions $[\boldsymbol{s}_i,\boldsymbol{a}_i,r_i,\boldsymbol{s}_{i+1},\varepsilon_i]^{\top}$ from experience replay buffer $\boldsymbol{R}$.
\STATE Update critic by minimizing TD error loss:   \\
$L_\text{TD}(\boldsymbol{\psi})=[r_i+\gamma Q_{i+1}\cdot(1-\varepsilon_i)-Q_i]^2$.
\STATE Update action with maximum Q loss:   \\
$L_\text{Q}(\boldsymbol{\phi})=-Q_i^{\prime}$.
\STATE Update target networks:  \\
$\boldsymbol{\phi}^{\prime}=\eta \boldsymbol{\phi}+(1-\eta) \boldsymbol{\phi}^{\prime}, \boldsymbol{\psi}^{\prime}=\eta \boldsymbol{\psi}+(1-\eta) \boldsymbol{\psi}^{\prime}$.
\ENDFOR
\ENDFOR
\end{algorithmic}
\end{algorithm}

\subsubsection{Spatio-Temporal Back Propagation (STBP) for TS-LIF-based SAN}
In our HDDPG algorithm, $f_\text{DCN}(\boldsymbol{\psi})$ is directly trained using the conventional back-propagation algorithm. $\boldsymbol{f}_\text{SAN}(\boldsymbol{\phi})$ has the characteristics of space-time dynamics and has the issue that the spike function is not differentiable. For $\boldsymbol{f}_\text{SAN}(\boldsymbol{\phi})$ based on TS-LIF spiking neurons, we train it using our improved spatio-temporal back-propagation (STBP) \cite{wu2018spatio}. According to Eq. (\ref{eq:5}) and Eq. (\ref{eq:11}), the parameters that $\boldsymbol{f}_\text{SAN}(\boldsymbol{\phi})$ finally needs to update are the weights $\boldsymbol{W}_n$ and the biases $\boldsymbol{b}_n$, so we need to obtain the derivative of the loss $L_\text{Q}$ with respect to $\boldsymbol{W}_n$ and $\boldsymbol{b}_n$. According to Eq. (\ref{eq:5}), we expand $\boldsymbol{f}_\text{SAN}(\boldsymbol{\phi})$ with TS-LIF spiking neurons in the space and time directions to obtain two chains of spatial propagation and temporal propagation, and realize iterative forward propagation. In the propagation chains, $\boldsymbol{W}_n$ and $\boldsymbol{b}_n$ are included in the current recurrence relation, so we calculate the derivative of the loss $L_\text{Q}$ to the currents $\boldsymbol{C}_n^t$ of the current layer $n$ at time $t$, and we can further calculate the derivative of  $L_\text{Q}$ with respect to $\boldsymbol{W}_n$ and $\boldsymbol{b}_n$. According to Eq. (\ref{eq:6}) and Eq. (\ref{eq:11}), the derivative of $L_\text{Q}$ with respect to the action $\boldsymbol{a}$ can be directly obtained according to the back-propagation of $f_\text{DCN}(\boldsymbol{\psi})$:
\begin{equation}
\label{eq:14}
\frac{\partial L_\text{Q}}{\partial \boldsymbol{a}}=-\frac{\partial Q}{\partial \boldsymbol{a}}=-\frac{\partial f_{\mathrm{DCN}}(\boldsymbol{s}, \boldsymbol{a})}{\partial \boldsymbol{a}}.
\end{equation}
According to Eq. (\ref{eq:5}), in order to obtain the derivative of $L_\text{Q}$ to the spikes $\boldsymbol{O}_{n}^{t}$, the voltages $\boldsymbol{U}_{n}^{t}$, and the currents $\boldsymbol{C}_{n}^{t}$, it can be divided into the following four cases depending on the layer $n$ and time $t$:

Case 1: At the time $t=\mathrm{T}$ and the output layer $n=\mathrm{N}$. In this case, only spatial derivatives exists. The derivative of $L_\text{Q}$ with respect to $\boldsymbol{O}_\mathrm{N}^\mathrm{T}$ is obtained by dividing the derivative of $L_\text{Q}$ with respect to $\boldsymbol{a}$ equally over each time:
\begin{equation}
\label{eq:15}
\frac{\partial L_\text{Q}}{\partial \boldsymbol{O}_\mathrm{N}^\mathrm{T}}=\left.\frac{\partial L_\text{Q}}{\partial \boldsymbol{O}_\mathrm{N}^\mathrm{T}}\right|_\text{spatial}=\frac{\partial L_\text{Q}}{\partial \boldsymbol{a}} \frac{\partial \boldsymbol{a}}{\partial \boldsymbol{O}_\mathrm{N}^\mathrm{T}}=\frac{1}{\mathrm{T}} \frac{\partial L_\text{Q}}{\partial \boldsymbol{a}}.
\end{equation}
The derivative of $L_\text{Q}$ with respect to $\boldsymbol{U}_\mathrm{N}^\mathrm{T}$ depends only on the derivative of $L_\text{Q}$ with respect to $\boldsymbol{O}_\mathrm{N}^\mathrm{T}$:
\begin{equation}
\label{eq:16}
\frac{\partial L_\text{Q}}{\partial \boldsymbol{U}_\mathrm{N}^\mathrm{T}}=\left.\frac{\partial L_\text{Q}}{\partial \boldsymbol{U}_\mathrm{N}^\mathrm{T}}\right|_\text{spatial}=\frac{\partial L_\text{Q}}{\partial \boldsymbol{O}_\mathrm{N}^\mathrm{T}} \frac{\partial \boldsymbol{O}_\mathrm{N}^\mathrm{T}}{\partial \boldsymbol{U}_\mathrm{N}^\mathrm{T}}=\frac{\partial L_\text{Q}}{\partial \boldsymbol{O}_\mathrm{N}^\mathrm{T}} \dot{\boldsymbol{h}}\left(\boldsymbol{U}_\mathrm{N}^\mathrm{T}\right).
\end{equation}
The derivative of $L_\text{Q}$ with respect to $\boldsymbol{C}_\mathrm{N}^\mathrm{T}$ depends only on the derivative of $L_\text{Q}$ with respect to $\boldsymbol{U}_\mathrm{N}^\mathrm{T}$:
\begin{equation}
\label{eq:17}
\frac{\partial L_\text{Q}}{\partial \boldsymbol{C}_\mathrm{N}^\mathrm{T}}=\left.\frac{\partial L_\text{Q}}{\partial \boldsymbol{C}_\mathrm{N}^\mathrm{T}}\right|_\text{spatial}=\frac{\partial L_\text{Q}}{\partial \boldsymbol{U}_\mathrm{N}^\mathrm{T}} \frac{\partial \boldsymbol{U}_\mathrm{N}^\mathrm{T}}{\partial \boldsymbol{C}_\mathrm{N}^\mathrm{T}}=\frac{\partial L_\text{Q}}{\partial \boldsymbol{U}_\mathrm{N}^\mathrm{T}}.
\end{equation}

Case2: At the time $t=\mathrm{T}$ and the middle layer $n<\mathrm{N}$. In this case, only spatial derivatives exists, recursively depending on error propagation in the spatial domain at time $\mathrm{T}$. The derivative of $L_\text{Q}$ with respect to $\boldsymbol{O}_\mathrm{n}^\mathrm{T}$ depends only on the derivative of $L_\text{Q}$ with respect to $\boldsymbol{C}_{n+1}^\mathrm{T}$:
\begin{equation}
\label{eq:18}
\frac{\partial L_\text{Q}}{\partial \boldsymbol{O}_{n}^\mathrm{T}}=\left.\frac{\partial L_\text{Q}}{\partial \boldsymbol{O}_{n}^\mathrm{T}}\right|_\text{spatial}=\frac{\partial L_\text{Q}}{\partial \boldsymbol{C}_{n+1}^\mathrm{T}} \frac{\partial \boldsymbol{C}_{n+1}^\mathrm{T}}{\partial \boldsymbol{O}_{n}^\mathrm{T}}=\frac{\partial L_\text{Q}}{\partial \boldsymbol{C}_{n+1}^\mathrm{T}} \boldsymbol{W}_{n}.
\end{equation}
The derivative of $L_\text{Q}$ with respect to $\boldsymbol{U}_{n}^\mathrm{T}$ depends only on the derivative of $L_\text{Q}$ with respect to $\boldsymbol{O}_{n}^\mathrm{T}$:
\begin{equation}
\label{eq:19}
\frac{\partial L_\text{Q}}{\partial \boldsymbol{U}_{n}^\mathrm{T}}=\left.\frac{\partial L_\text{Q}}{\partial \boldsymbol{U}_{n}^\mathrm{T}}\right|_\text{spatial}=\frac{\partial L_\text{Q}}{\partial \boldsymbol{O}_{n}^\mathrm{T}} \frac{\partial \boldsymbol{O}_{n}^\mathrm{T}}{\partial \boldsymbol{U}_{n}^\mathrm{T}}=\frac{\partial L_\text{Q}}{\partial \boldsymbol{O}_{n}^\mathrm{T}} \dot{\boldsymbol{h}}\left(\boldsymbol{U}_{n}^\mathrm{T}\right).
\end{equation}
The derivative of $L_\text{Q}$ with respect to $\boldsymbol{C}_{n}^\text{T}$ depends only on the derivative of $L_\text{Q}$ with respect to $\boldsymbol{U}_{n}^\text{T}$:
\begin{equation}
\label{eq:20}
\frac{\partial L_\text{Q}}{\partial \boldsymbol{C}_{n}^\mathrm{T}}=\left.\frac{\partial L_\text{Q}}{\partial \boldsymbol{C}_{n}^\mathrm{T}}\right|_\text{spatial}=\frac{\partial L_\text{Q}}{\partial \boldsymbol{U}_{n}^\mathrm{T}} \frac{\partial \boldsymbol{U}_{n}^\mathrm{T}}{\partial \boldsymbol{C}_{n}^\mathrm{T}}=\frac{\partial L_\text{Q}}{\partial \boldsymbol{U}_{n}^\mathrm{T}}.
\end{equation}

Case 3: At the time $t<\mathrm{T}$ and the output layer $n=\mathrm{N}$. In this case, the derivative depends on both the temporal and spatial domain directions of error propagation. The derivative of $L_\text{Q}$ with respect to $\boldsymbol{O}_\mathrm{N}^{t}$ depends on both the derivative of $L_\text{Q}$ with respect to $\boldsymbol{a}$ in the space domain and the derivative of $L_\text{Q}$ with respect to $\boldsymbol{U}_\mathrm{N}^{t+1}$ in the time domain:
\begin{equation}
\label{eq:21}
\begin{aligned}
\frac{\partial L_\text{Q}}{\partial \boldsymbol{O}_\mathrm{N}^{t}} &=\left.\frac{\partial L_\text{Q}}{\partial \boldsymbol{O}_\mathrm{N}^{t}}\right|_\text{spatial}+\left.\frac{\partial L_\text{Q}}{\partial \boldsymbol{O}_\mathrm{N}^{t}}\right|_{\text{temporal}} \\
&=\frac{\partial L_\text{Q}}{\partial \boldsymbol{a}} \frac{\partial \boldsymbol{a}}{\partial \boldsymbol{O}_\mathrm{N}^{t}}+\frac{\partial L_\text{Q}}{\partial \boldsymbol{U}_\mathrm{N}^{t+1}} \frac{\partial \boldsymbol{U}_\mathrm{N}^{t+1}}{\partial \boldsymbol{O}_\mathrm{N}^{t}} \\
&=\frac{1}{\mathrm{T}} \frac{\partial L_\text{Q}}{\partial \boldsymbol{a}}+\frac{\partial L_\text{Q}}{\partial \boldsymbol{U}_\mathrm{N}^{t+1}} \delta_{\text{volt}, \mathrm{N}} \boldsymbol{U}_\mathrm{N}^{t} \dot{\boldsymbol{g}}\left(\boldsymbol{O}_\mathrm{N}^{t}\right).
\end{aligned}
\end{equation}
The derivative of $L_\text{Q}$ with respect to $\boldsymbol{U}_\mathrm{N}^{t}$ depends on both the derivative of $L_\text{Q}$ with respect to $\boldsymbol{O}_\mathrm{N}^{t}$ in the space domain and the derivative of $L_\text{Q}$ with respect to $\boldsymbol{U}_\mathrm{N}^{t+1}$ in the time domain:
\begin{equation}
\label{eq:22}
\begin{aligned}
\frac{\partial L_\text{Q}}{\partial \boldsymbol{U}_\mathrm{N}^{t}} &=\left.\frac{\partial L_\text{Q}}{\partial \boldsymbol{U}_\mathrm{N}^{t}}\right|_\text{spatial}+\left.\frac{\partial L_\text{Q}}{\partial \boldsymbol{U}_\mathrm{N}^{t}}\right|_\text{temporal} \\
&=\frac{\partial L_\text{Q}}{\partial \boldsymbol{O}_\mathrm{N}^{t}} \frac{\partial \boldsymbol{O}_\mathrm{N}^{t}}{\partial \boldsymbol{U}_\mathrm{N}^{t}}+\frac{\partial L_\text{Q}}{\partial \boldsymbol{U}_\mathrm{N}^{t+1}} \frac{\partial \boldsymbol{U}_\mathrm{N}^{t+1}}{\partial \boldsymbol{U}_\mathrm{N}^{t}} \\
&=\frac{\partial L_\mathrm{Q}}{\partial \boldsymbol{O}_\mathrm{N}^{t}} \dot{\boldsymbol{h}}\left(\boldsymbol{U}_\mathrm{N}^{t}\right)+\frac{\partial L_\text{Q}}{\partial \boldsymbol{U}_\mathrm{N}^{t+1}} \delta_{\text{volt},\mathrm{N}} \boldsymbol{g}\left(\boldsymbol{O}_\mathrm{N}^{t}\right).
\end{aligned}
\end{equation}
The derivative of $L_\text{Q}$ with respect to $\boldsymbol{C}_\mathrm{N}^{t}$ depends both on the derivative of $L_\text{Q}$ with respect to $\boldsymbol{U}_\mathrm{N}^{t}$ in the space domain, and the derivative of $L_\text{Q}$ with respect to $\boldsymbol{C}_\mathrm{N}^{t+1}$ in the time domain:
\begin{equation}
\label{eq:23}
\begin{aligned}
\frac{\partial L_\text{Q}}{\partial \boldsymbol{C}_\mathrm{N}^{t}} &=\left.\frac{\partial L_\text{Q}}{\partial \boldsymbol{C}_\mathrm{N}^{t}}\right|_\text{spatial}+\left.\frac{\partial L_\text{Q}}{\partial \boldsymbol{C}_\mathrm{N}^{t}}\right|_\text{temporal} \\
&=\frac{\partial L_\text{Q}}{\partial \boldsymbol{U}_\mathrm{N}^{t}} \frac{\partial \boldsymbol{U}_\mathrm{N}^{t}}{\partial \boldsymbol{C}_\mathrm{N}^{t}}+\frac{\partial L_\text{Q}}{\partial \boldsymbol{C}_\mathrm{N}^{t+1}} \frac{\partial \boldsymbol{C}_\mathrm{N}^{t+1}}{\partial \boldsymbol{U}_\mathrm{N}^{t}} \\
&=\frac{\partial L_\text{Q}}{\partial \boldsymbol{U}_\mathrm{N}^{t}}+\frac{\partial L_\text{Q}}{\partial \boldsymbol{C}_\mathrm{N}^{t+1}} \delta_{\text{curr}, \mathrm{N}}.
\end{aligned}
\end{equation}

Case 4: At the time $t<\mathrm{T}$ and the middle layer $n<\mathrm{N}$. In this case, the derivative depends on both the temporal and spatial domain directions of error propagation. The derivative of $L_\text{Q}$ with respect to $\boldsymbol{O}_{n}^{t}$ depends on both the derivative of $L_\text{Q}$ with respect to $\boldsymbol{C}_{n+1}^{t}$ in the space domain and the derivative of $L_\text{Q}$ with respect to $\boldsymbol{U}_{n+1}^{t}$ in the time domain:
\begin{equation}
\label{eq:24}
\begin{aligned}
\frac{\partial L_\text{Q}}{\partial \boldsymbol{O}_{n}^{t}} &=\left.\frac{\partial L_\text{Q}}{\partial \boldsymbol{O}_{n}^{t}}\right|_\text{spatial}+\left.\frac{\partial L_\text{Q}}{\partial \boldsymbol{O}_{n}^{t}}\right|_\text{temporal} \\
&=\frac{\partial L_\text{Q}}{\partial \boldsymbol{C}_{n+1}^{t}} \frac{\partial \boldsymbol{C}_{n+1}^{t}}{\partial \boldsymbol{O}_{n}^{t}}+\frac{\partial L_\text{Q}}{\partial \boldsymbol{U}_{n}^{t+1}} \frac{\partial \boldsymbol{U}_{n}^{t+1}}{\partial \boldsymbol{O}_{n}^{t}} \\
&=\frac{\partial L_\text{Q}}{\partial \boldsymbol{C}_{n+1}^{t}} \boldsymbol{W}_{n+1}+\frac{\partial L_\text{Q}}{\partial \boldsymbol{U}_{n}^{t+1}} \delta_{\text{volt}, n} \boldsymbol{U}_{n}^{t} \dot{\boldsymbol{g}}\left(\boldsymbol{O}_{n}^{t}\right).
\end{aligned}
\end{equation}
The derivative of $L_\text{Q}$ with respect to $\boldsymbol{U}_{n}^{t}$ depends on both the derivative of $L_\text{Q}$ with respect to $\boldsymbol{O}_{n}^{t}$ in the space domain and the derivative of $L_\text{Q}$ with respect to $\boldsymbol{U}_{n}^{t+1}$ in the time domain:
\begin{equation}
\label{eq:25}
\begin{aligned}
\frac{\partial L_\text{Q}}{\partial \boldsymbol{U}_{n}^{t}} &=\left.\frac{\partial L_\text{Q}}{\partial \boldsymbol{U}_{n}^{t}}\right|_\text{spatial}+\left.\frac{\partial L_\text{Q}}{\partial \boldsymbol{U}_{n}^{t}}\right|_\text{temporal} \\
&=\frac{\partial L_\text{Q}}{\partial \boldsymbol{O}_{n}^{t}} \frac{\partial \boldsymbol{O}_{n}^{t}}{\partial \boldsymbol{U}_{n}^{t}}+\frac{\partial L_\text{Q}}{\partial \boldsymbol{U}_{n}^{t+1}} \frac{\partial \boldsymbol{U}_{n}^{t+1}}{\partial \boldsymbol{U}_{n}^{t}} \\
&=\frac{\partial L_\text{Q}}{\partial \boldsymbol{O}_{n}^{t}} \dot{\boldsymbol{h}}\left(\boldsymbol{U}_{n}^{t}\right)+\frac{\partial L_\text{Q}}{\partial \boldsymbol{U}_{n}^{t+1}} \delta_{\text{volt}, n} \boldsymbol{g}\left(\boldsymbol{O}_{n}^{t}\right).
\end{aligned}
\end{equation}
The derivative of $L_\text{Q}$ with respect to $\boldsymbol{C}_{n}^{t}$ depends both on the derivative of $L_\text{Q}$ with respect to $\boldsymbol{U}_{n}^{t}$ in the space domain, and the derivative of $L_\text{Q}$ with respect to $\boldsymbol{C}_{n}^{t+1}$ in the time domain:
\begin{equation}
\label{eq:26}
\begin{aligned}
\frac{\partial L_\text{Q}}{\partial \boldsymbol{C}_{n}^{t}} &=\left.\frac{\partial L_\text{Q}}{\partial \boldsymbol{C}_{n}^{t}}\right|_\text{spatial}+\left.\frac{\partial L_\text{Q}}{\partial \boldsymbol{C}_{n}^{t}}\right|_\text{temporal} \\
&=\frac{\partial L_\text{Q}}{\partial \boldsymbol{U}_{n}^{t}} \frac{\partial \boldsymbol{U}_{n}^{t}}{\partial \boldsymbol{C}_{n}^{t}}+\frac{\partial L_\text{Q}}{\partial \boldsymbol{C}_{n}^{t+1}} \frac{\partial \boldsymbol{C}_{n}^{t+1}}{\partial \boldsymbol{U}_{n}^{t}} \\
&=\frac{\partial L_\text{Q}}{\partial \boldsymbol{U}_{n}^{t}}+\frac{\partial L_\text{Q}}{\partial \boldsymbol{C}_{n}^{t+1}} \delta_{\text{curr}, n}.
\end{aligned}
\end{equation}

\begin{figure*}[htbp]
\centering
\includegraphics[width=0.8\textwidth]{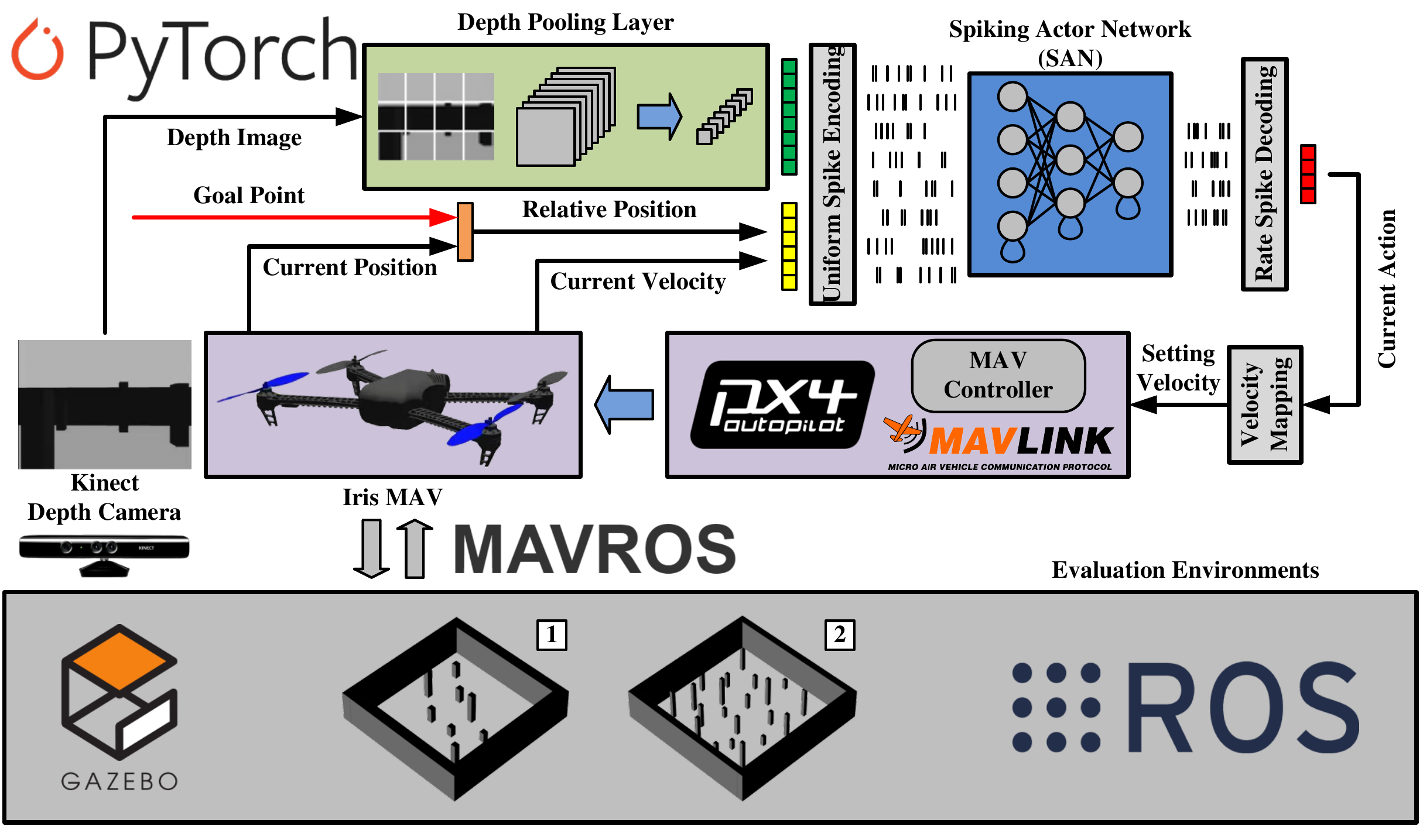}
\caption{MAV simulation system framework through the robot operating system (ROS).}
\label{fig:5}
\vspace{-0mm}
\end{figure*}

Based on the above four cases, we can get the derivative of $L_\text{Q}$ to $\boldsymbol{O}_{n}^{t}$, $\boldsymbol{U}_{n}^{t}$, $\boldsymbol{C}_{n}^{t}$ at any layer $n$ and any time $t$, and then get the derivative of $L_\text{Q}$ to $\boldsymbol{W}_n$ and $\boldsymbol{b}_n$ as follows:
\begin{equation}
\label{eq:27}
\begin{aligned}
&\frac{\partial L_\text{Q}}{\partial \boldsymbol{W}_{n}}=\sum_{t=1}^\mathrm{T} \frac{\partial L_\text{Q}}{\partial \boldsymbol{C}_{n}^{t}} \frac{\partial \boldsymbol{C}_{n}^{t}}{\boldsymbol{W}_{n}}=\sum_{t=1}^\mathrm{T} \frac{\partial L_\text{Q}}{\partial \boldsymbol{C}_{n}^{t}} \boldsymbol{O}_{n-1}^{t}, \\
&\frac{\partial L_\text{Q}}{\partial \boldsymbol{b}_{n}}=\sum_{t=1}^\mathrm{T} \frac{\partial L_\text{Q}}{\partial \boldsymbol{C}_{n}^{t}} \frac{\partial \boldsymbol{C}_{n}^{t}}{\boldsymbol{b}_{n}}=\sum_{t=1}^\mathrm{T} \frac{\partial L_\text{Q}}{\partial \boldsymbol{C}_{n}^{t}},
\end{aligned}
\end{equation}
where $\frac{\partial L_\text{Q}}{\partial \boldsymbol{C}_{n}^{t}}$ can be obtained from Eq. (\ref{eq:14}) to Eq. (\ref{eq:26}). In addition, since the spike generating function $\boldsymbol{h}(\cdot)$ of the TS-LIF spiking neurons is not differentiable, here we use the rectangular function as the surrogate gradient function to approximate the derivative of $\boldsymbol{h}(\cdot)$:
\begin{equation}
\label{eq:28}
\dot{\boldsymbol{h}}\left(\boldsymbol{U}_{n}^{t}\right)=\frac{1}{m} \boldsymbol{H}\left(\frac{m}{2}-\left|\boldsymbol{U}_{n}^{t}-u_{\text{th}, n}\right|\right).
\end{equation}
Among them, coefficient $m=1.0$ and spike trigger threshold $u_{\text{th}, n}=0.5$. Based on the STBP framework for the above TS-LIF spiking neurons, we can use gradient descent to efficiently train $\boldsymbol{f}_\text{SAN}(\boldsymbol{\phi})$. In experiments, we also compare other two SNN training frameworks, back-propagation through time (BPTT) \cite{werbos1990backpropagation} and spike layer error reassignment (SLAYER) \cite{shrestha2018slayer}.

\subsection{MAV Simulation System}
\subsubsection{Software-in-the-Loop Simulation Framework}
In order to perform hybrid reinforcement training and spiking network inference in the simulation environment, we built a complete set of MAV software-in-the-loop (SITL) simulation system based on the Robot Operating System (ROS), as shown in Fig. \ref{fig:5}. It mainly includes the following components:
\begin{itemize}
\item Simulation Environment. We use Gazebo \footnote{Gazebo Website: \url{https://gazebosim.org}.} simulator to set up the training environment and the evaluation environment respectively. We learn the insight of curriculum learning and set up four training environments from simple to complex in total. In order to evaluate the generalization of our algorithm, we set up a total of two evaluation environments. Evaluation environment \#1 is similar in size to the training environment, and Evaluation environment \#2 is larger, more complex and more challenging.
\item MAV with Depth Camera. In the Gazebo simulation framework, we choose the Iris MAV and the Kinect depth camera, and the depth camera is fixed at $(0, 0, 0.14\text{m})$ the MAV body.
\item Flight Control. In order to realize the speed control of the MAV, we choose PX4 \footnote{PX4 Website: \url{https://px4.io/}.} \cite{meier2015px4} as the flight controller of the MAV. PX4 is an open-source professional flight controller that supports a variety of MAV models and simulation platforms, providing a wealth of application programming interfaces (APIs) to facilitate the deployment of actual MAVs. It is widely used in the current robotics research community around the world.
\item Communication Mechanism. We use ROS nodes for message interaction between algorithm programs, Gazebo simulation environment and PX4 flight controller.
\end{itemize}

\subsubsection{MAV Take-Off Process}
For the training or evaluation, the MAV takes off to a fixed altitude from a random starting point at the beginning of each episode. First, the MAV randomly appears in a specified starting point on the ground by the service in Gazebo. Then, the MAV switches to the off-board mode and enters the idle speed. Finally, the MAV flies to the fixed altitude to perform autonomous navigation after a period of time.

\subsubsection{Judgment of MAV over Obstacles}
\begin{figure}[htbp]
\centering
\subfigure[]{\label{fig:6a}\includegraphics[width=0.49\columnwidth]{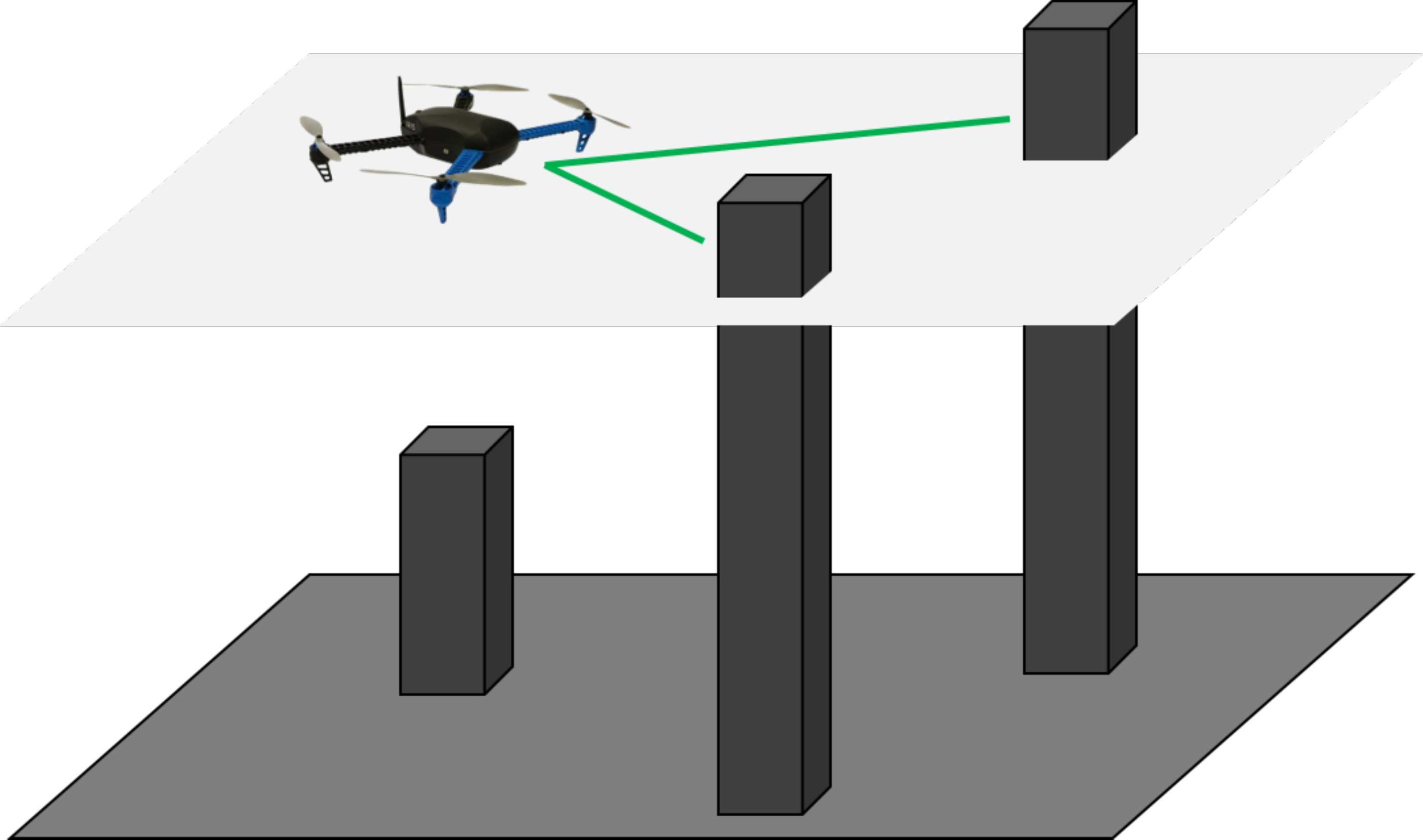}}
\subfigure[]{\label{fig:6b}\includegraphics[width=0.49\columnwidth]{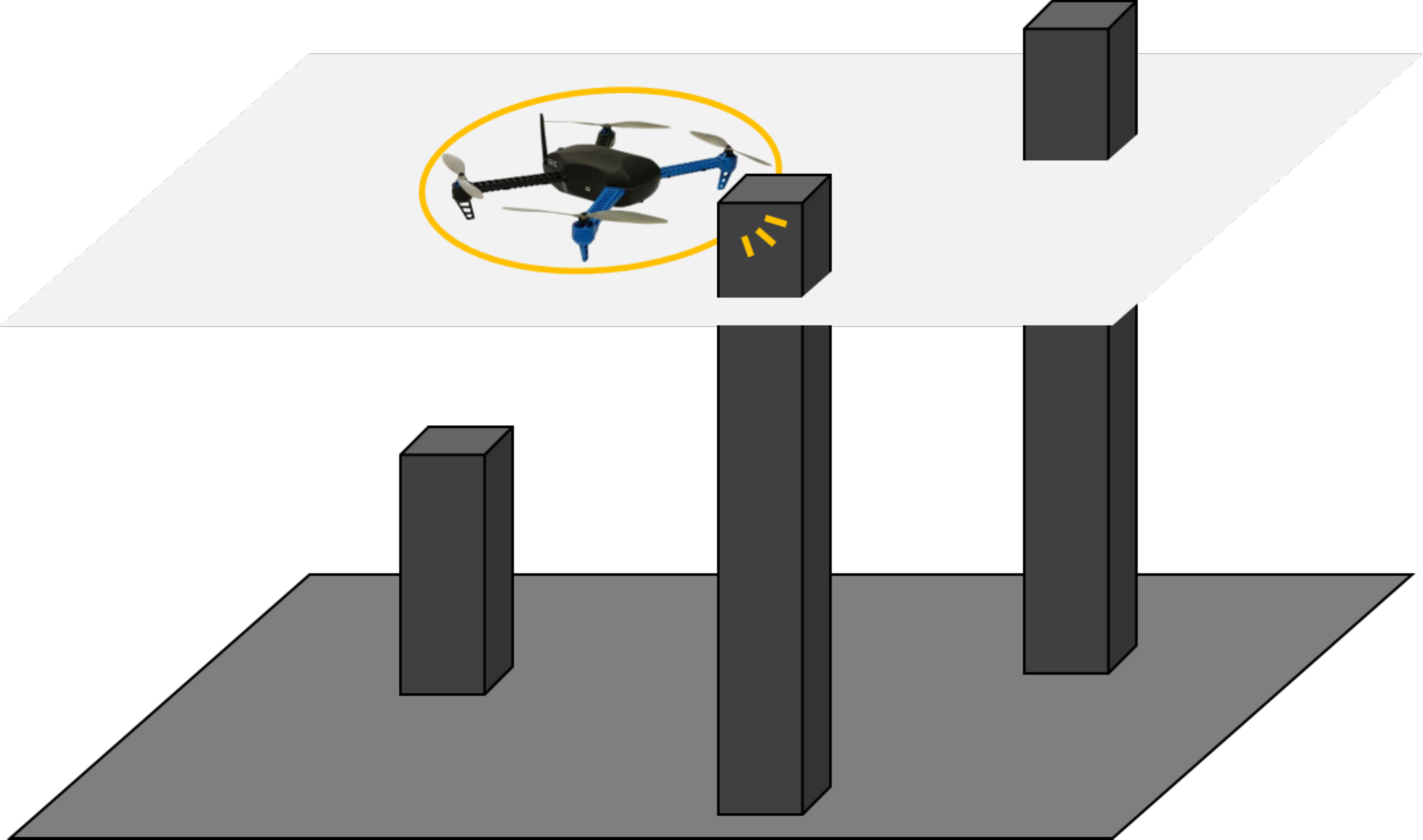}}
\caption{Judgment of MAV over obstacles. (a) Only distance from obstacles higher than the current flight altitude is calculated. (b) If the distance between the MAV and the obstacle is less than the threshold, a collision is considered to have occurred.}
\label{fig:6}
\vspace{-0mm}
\end{figure}

During the training process, it is necessary to determine whether the MAV has pass the obstacle. We propose an appropriate judgment of MAV over obstacles as shown in Fig. \ref{fig:6}. For obstacles lower than the MAV current height, it is considered that the MAV can pass directly. For obstacles higher than the MAV current height, the distance in xy-plane between the MAV and obstacle are calculated. If the distance between the MAV and the obstacle in xy-plane is less than the collision threshold, it is considered that the MAV has collided, otherwise the MAV is considered to pass the obstacle safely.

\section{Experiments}
\label{sec:experiments}
In this section, we conduct multiple experiments in the Gazebo simulation environments to verify the effectiveness of the Neuro-Planner proposed in this paper through qualitative and quantitative experimental results. First, we evaluate the impact of original DDPG and HDDPG implemented by three different SNN training frameworks (STBP \cite{wu2018spatio}, BPTT \cite{werbos1990backpropagation}, and SLAYER \cite{shrestha2018slayer}) on the performance during training. Then, the success rate, average distances and average time of successful paths by different training frameworks are evaluated in two significantly different evaluation environments, and success and failure trajectories are shown finally. In the following part, we use HDDPG-STBP, HDDPG-BPTT, HDDPG-SLAYER to represent HDDPG algorithm trained with STBP, BPTT and SLAYER frameworks respectively.

\subsection{Experiment Setup}

\begin{figure}[htbp]
\vspace{-0mm}
\centering
\subfigure[]{\label{fig:7a}\includegraphics[width=\columnwidth]{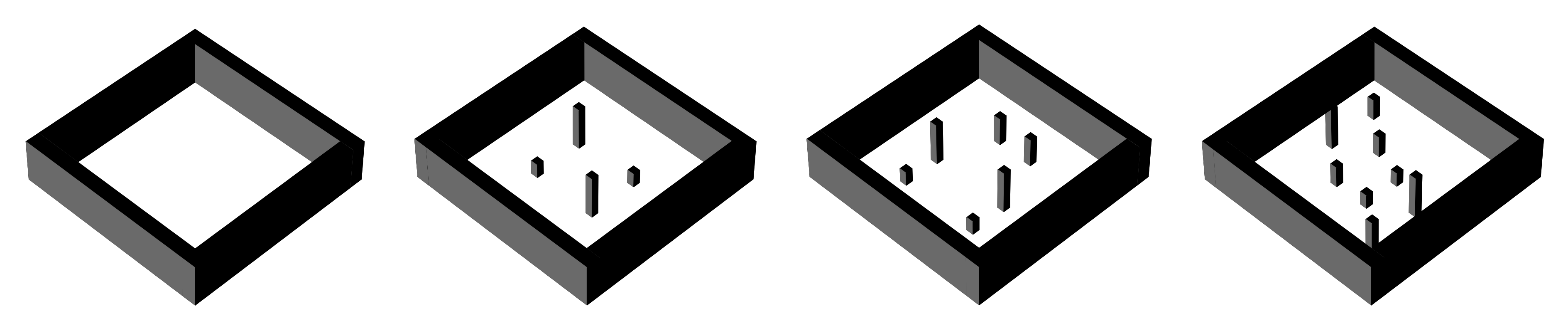}}
\subfigure[]{\label{fig:7b}\includegraphics[width=\columnwidth]{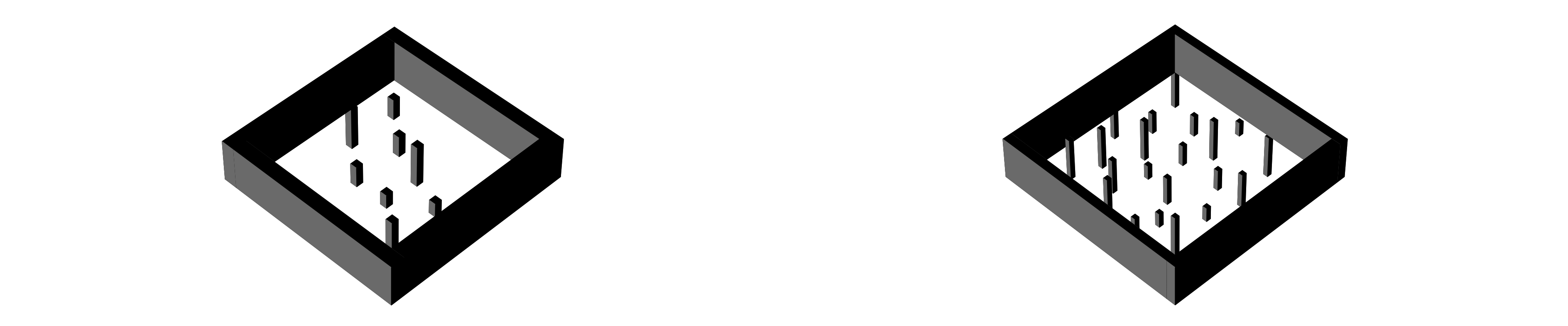}}
\caption{Training and evaluation environments in Gazebo simulator. We adopt the insight of curriculum learning in setting training environments. (a) Training environments \#1-\#4: Four training environments $(12\mathrm{m}\times12\mathrm{m}\times3\mathrm{m})$ of increasing complexity with 0, 4, 6 and 8 cuboid pillars. \cite{bengio2009curriculum} \cite{graves2017automated} \cite{soviany2021curriculum}. (b) Evaluation environments \#1-\#2: Two evaluation environments. Evaluation environment \#1 is similar to training environment \#4, and evaluation environment \#2 is $20\mathrm{m}\times20\mathrm{m}\times3\mathrm{m}$ with 21 cuboid pillars.}
\label{fig:7}
\vspace{0mm}
\end{figure}

\begin{table}[htbp]
\vspace{-0mm}
\begin{center}
\caption{Hyper-Parameters for Neuro-Planner Training}
\setlength{\tabcolsep}{1pt}
\newcommand{\tabincell}[2]{\begin{tabular}{@{}#1@{}}#2\end{tabular}}
\small
%\begin{tabular*}{0.95\linewidth}{@{}@{\extracolsep{\fill}}c|c|cccp{1cm}@{}}
\setlength{\tabcolsep}{0.02\linewidth}
\begin{tabular}{c|cp{1cm}}
\toprule
Parameters          &Values     \\
\midrule
Channels of state $\mathrm{C}_{\boldsymbol{s}}$                         &18     \\
Channels of normalized state $\mathrm{C}_{\tilde{\boldsymbol{s}}}$      &21     \\
Channels of action $\mathrm{C}_{\boldsymbol{a}}$                        &4      \\
\midrule
Spike trigger threshold $u_{\text{th},n}$                               &0.5    \\
Decay factor of current $\delta_{\text{curr},n}$                        &0.5    \\
Decay factor of voltage $\delta_{\text{volt},n}$                        &0.75   \\
\midrule
Velocity mapping factors $\alpha_{1},\alpha_{2},\alpha_{3}$             &0.225, 1.8, 0.18   \\
Minimum plane linear velocity $v_\text{min}$                            &0.05               \\
\midrule
Hidden layers of SAN     &3 \\
Hidden layers of DCN     &3 \\
Neurons in each hidden layer of SAN                                     &512    \\
Neurons in each hidden layer of DCN                                     &512    \\
\midrule
Reward for reaching the goal point $r_\text{goal}$                    &30     \\
Reward for collision with obstacle $r_\text{obs}$                       &-30    \\
Amplified factor of rewards $\sigma$                                    &15     \\
Distance threshold to goal point $\varepsilon_\text{goal,th}$         &0.54 m \\
Distance threshold to obstacle $\varepsilon_\text{obs,th}$              &0.5 m  \\
\midrule
Learning rate of SAN                                                    &0.00001    \\
Learning rate of DCN                                                    &0.0001     \\
\midrule
Sizes of experience replay buffer                                       &100000     \\
Soft update factor $\eta$                                               &0.01       \\
Batch size $\text{B}$                                                   &256        \\
\midrule
Height restrictions                                                     &0.3m, 3.1m \\
\bottomrule
%\end{tabular*}
\end{tabular}
\label{tab:1}
\vspace{0mm}
\end{center}
\end{table}

In this paper, we use a hybrid reinforcement training method based on curriculum learning \cite{bengio2009curriculum} \cite{graves2017automated} \cite{soviany2021curriculum} to train the actor-critic network. We set up a total of four training environments, sequentially increasing the difficulty of the environment to speed up training and improve performance, as shown in Fig. \ref{fig:7a}. Four training environments with a size of $12\text{m} \times 12\text{m} \times 3\text{m}$ are set with closed walls as enclosures. In the four training environments, 0, 4, 6, and 8 cuboid pillars are placed as obstacles in sequence. The placements of obstacles in each training environment are shown in Fig. \ref{fig:7}. In order to enable the MAV to fully learn the navigation policy, we set 100, 200, 300, and 400 training episodes in each training environment, and randomly generated starting points and goal points. The training parameters are shown in Table \ref{tab:1}.

For the evaluation environment, we set up two unfamiliar environments, as shown in Fig. \ref{fig:7b}. Evaluation environment \#1 changes the height of some columns compared to training environment \#4. To further verify the robustness of our Neuro-Planner, evaluation environment \#2 is designed more complexity with $20\text{m} \times 20\text{m} \times 3\text{m}$. A total of 21 cuboid pillars are placed in evaluation environment \#2 as obstacles. In each evaluation environment, we randomly generate 100 starting points and goal points for evaluation, and compare the performance of the spiking actor networks trained with different training frameworks and the artificial network. In order to ensure that the evaluation difficulty of each method is consistent, each method will use the same starting points and goal points. In the evaluation phase, we take the success rate as a priority condition. To exclude chance, we repeat the evaluation three times for each different training frameworks, and take the average success rate. Besides the success rate, we also calculated the average distance and average speed when each method successfully reached the goal point. Both training and evaluation in our experiments are performed under Ubuntu 18.04, using Gazebo9 as the simulator. The CPU is Intel i7-7700 and the GPU is Nvidia GTX 1650.

\subsection{Performance Analysis in Training}
Here, we compare the training performance of HDDPG based on three SNN training frameworks including STBP, BPTT and SLAYER and original DDPG. We analyze average reward and training time respectively.

\begin{figure*}[htbp]
\vspace{-0mm}
\centering
\subfigure[]{\label{fig:8a}\includegraphics[width=0.245\textwidth]{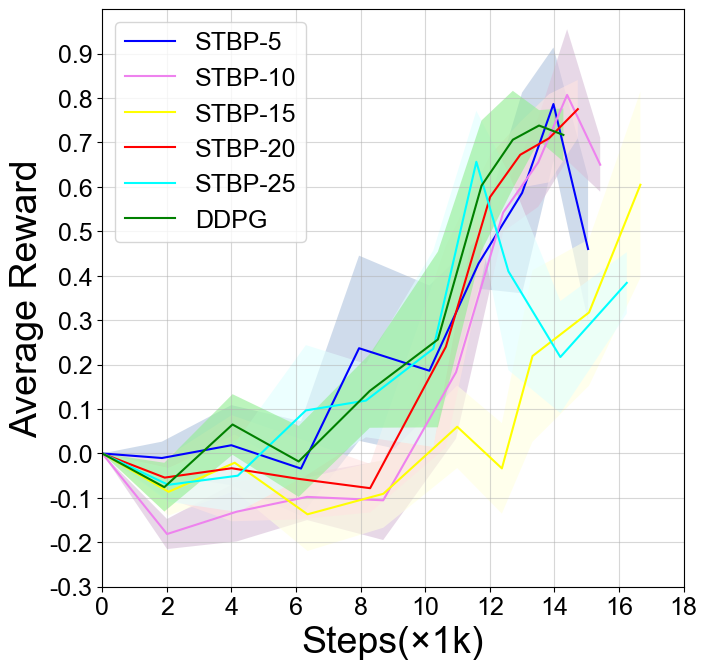}}
\subfigure[]{\label{fig:8b}\includegraphics[width=0.245\textwidth]{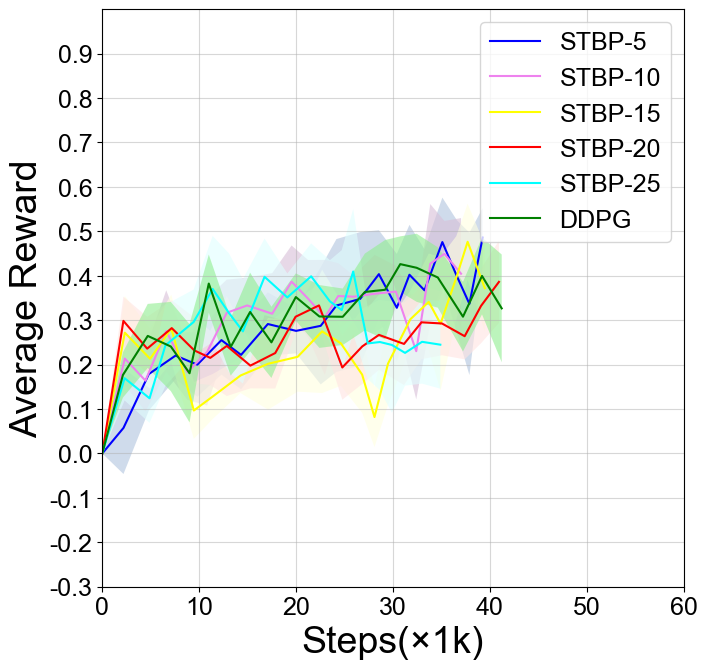}}
\subfigure[]{\label{fig:8c}\includegraphics[width=0.245\textwidth]{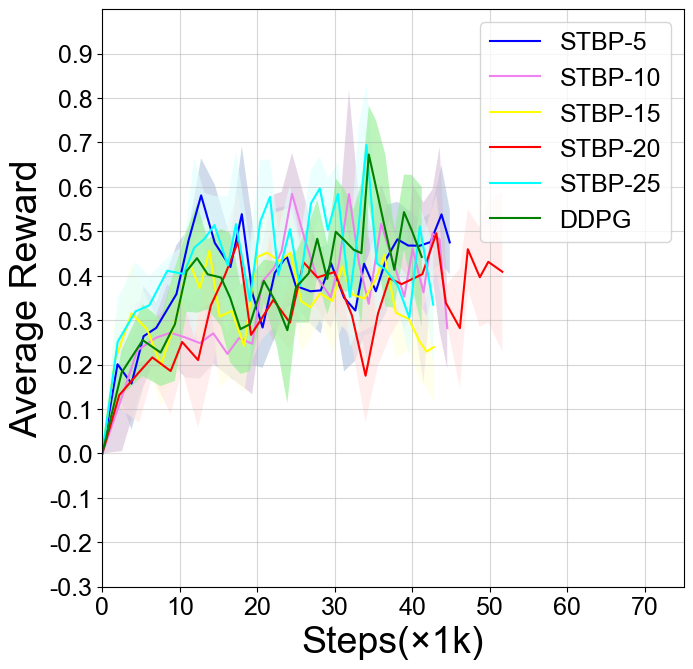}}
\subfigure[]{\label{fig:8d}\includegraphics[width=0.245\textwidth]{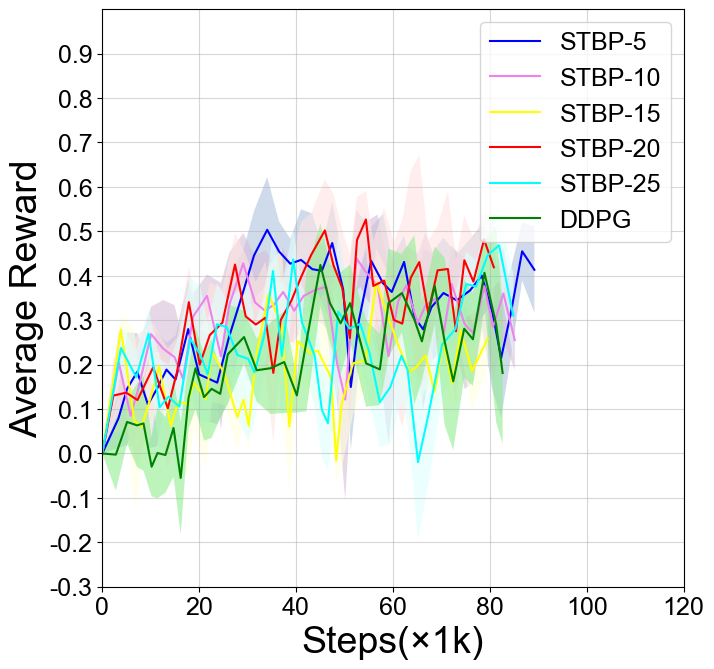}}
\subfigure[]{\label{fig:8e}\includegraphics[width=0.245\textwidth]{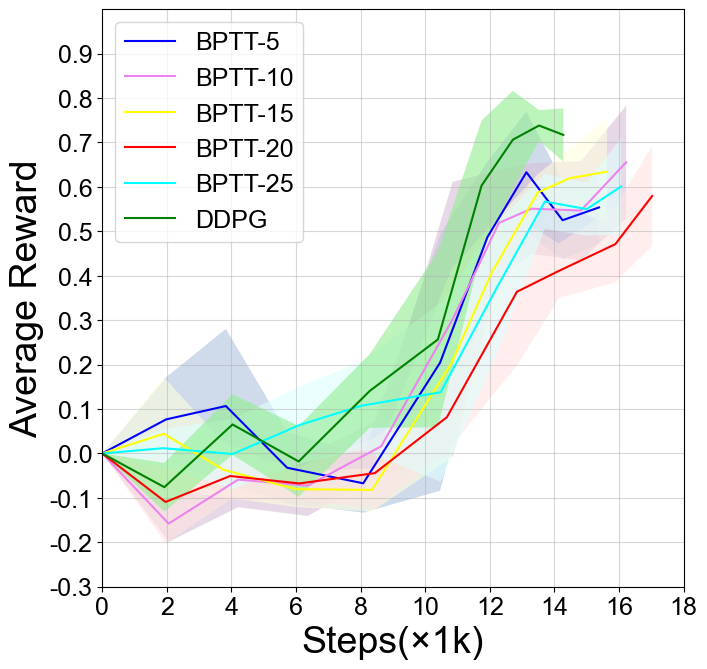}}
\subfigure[]{\label{fig:8f}\includegraphics[width=0.245\textwidth]{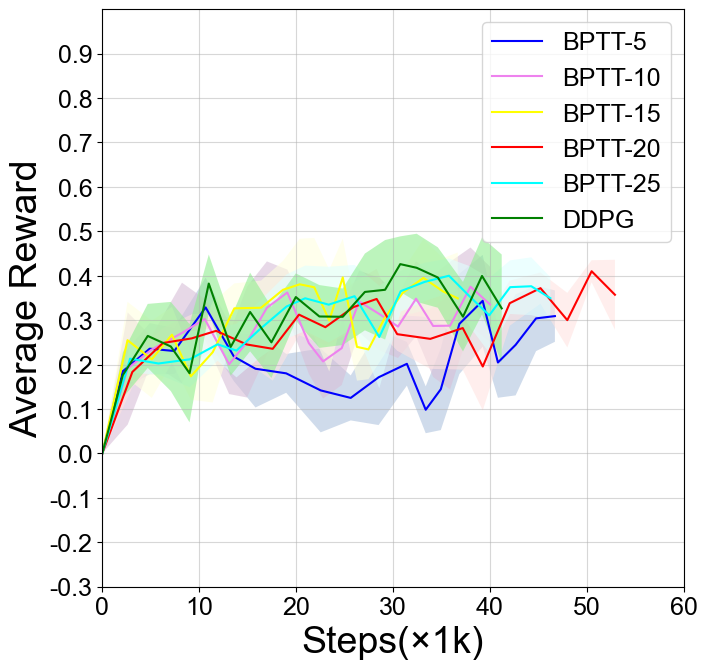}}
\subfigure[]{\label{fig:8g}\includegraphics[width=0.245\textwidth]{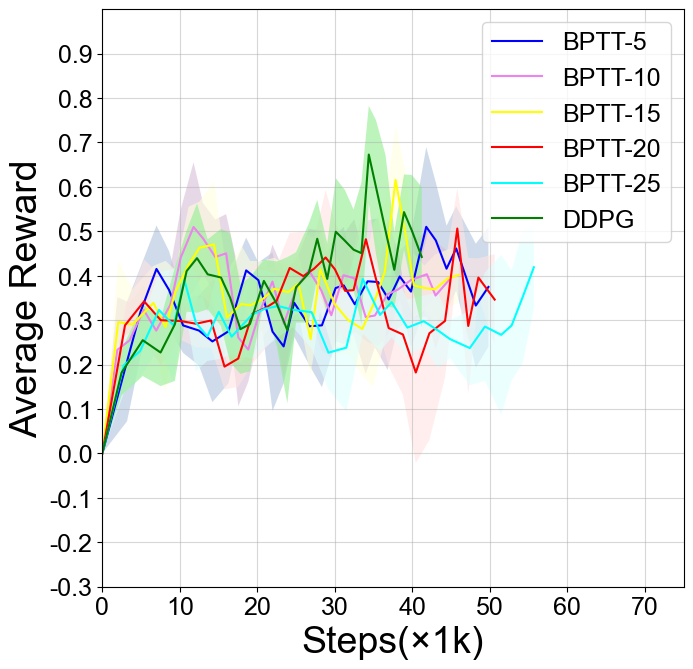}}
\subfigure[]{\label{fig:8h}\includegraphics[width=0.245\textwidth]{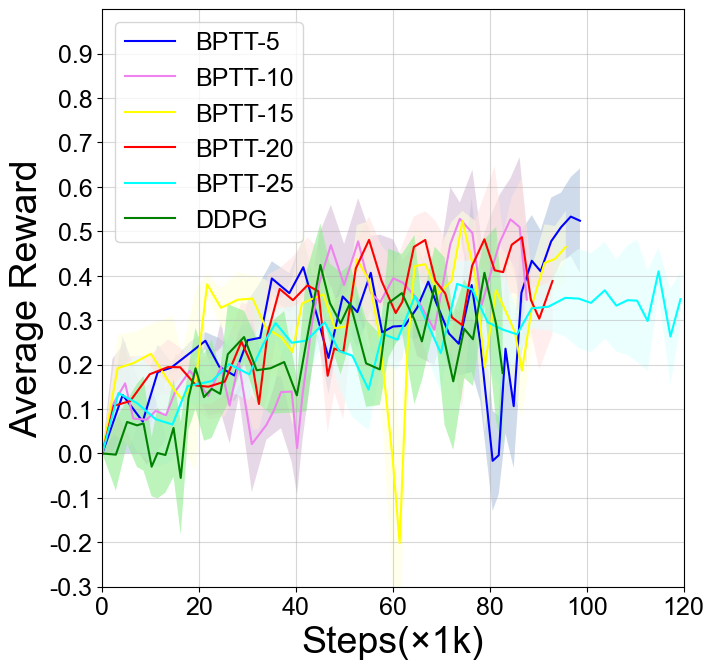}}
\subfigure[]{\label{fig:8i}\includegraphics[width=0.245\textwidth]{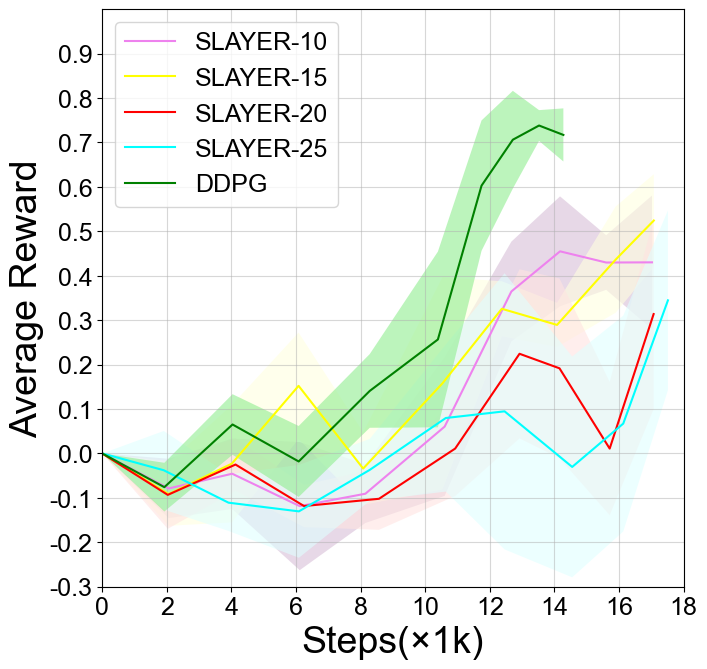}}
\subfigure[]{\label{fig:8j}\includegraphics[width=0.245\textwidth]{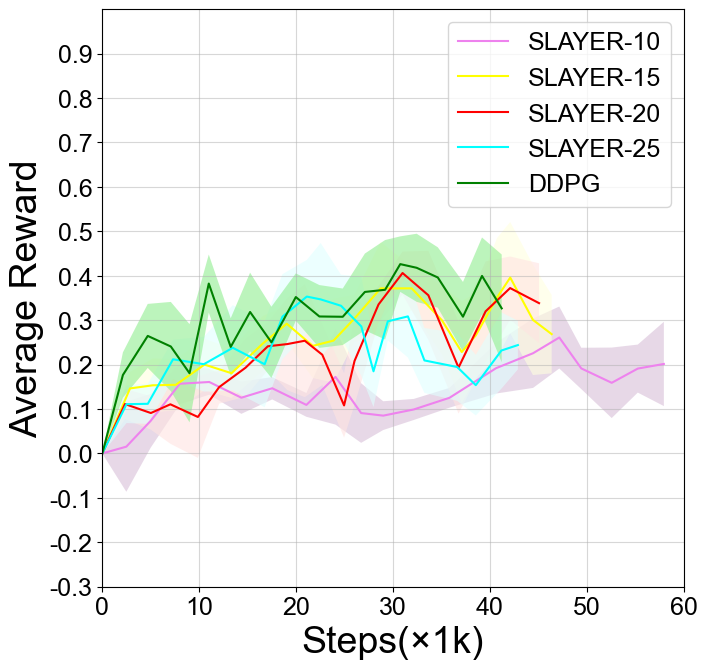}}
\subfigure[]{\label{fig:8k}\includegraphics[width=0.245\textwidth]{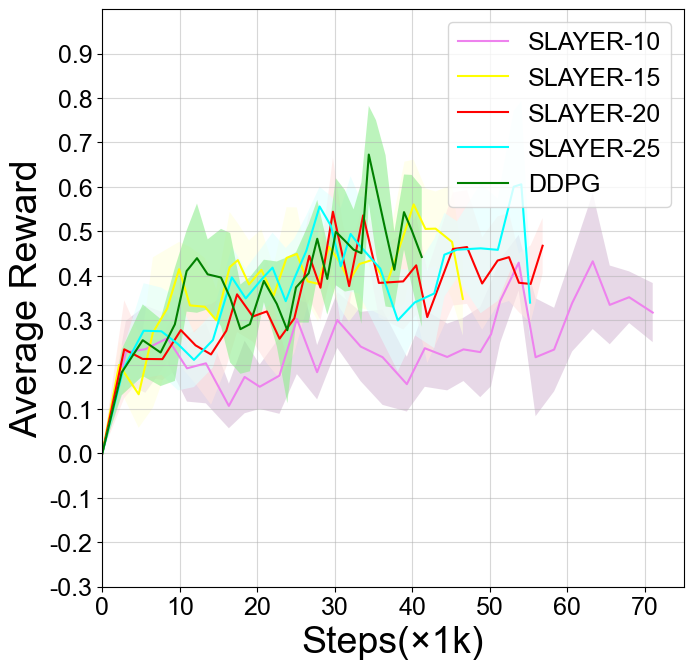}}
\subfigure[]{\label{fig:8l}\includegraphics[width=0.245\textwidth]{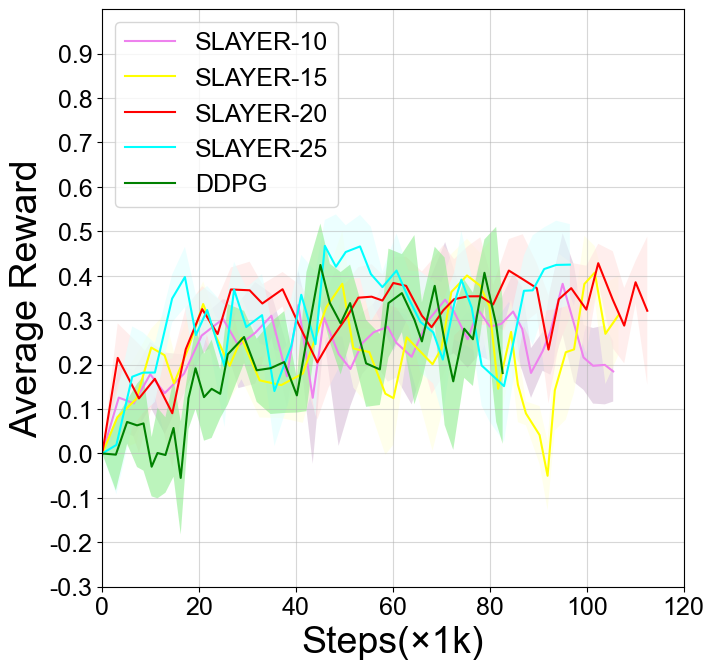}}
\caption{Average reward curves of hybrid actor-critic network of our proposed HDDPG (with various SNN training frameworks and different time steps) and original DDPG. (a) HDDPG-STBP, training environment \#1. (b) HDDPG-STBP, training environment \#2. (c) HDDPG-STBP, training environment \#3. (d) HDDPG-STBP, training environment \#4. (e) HDDPG-BPTT, training environment \#1. (f) HDDPG-BPTT, training environment \#2. (g) HDDPG-BPTT, training environment \#3. (h) HDDPG-BPTT, training environment \#4. (i) HDDPG-SLAYER, training environment \#1. (j) HDDPG-SLAYER, training environment \#2. (k) HDDPG-SLAYER, training environment \#3. (l) HDDPG-SLAYER, training environment \#4.}
\label{fig:8}
\vspace{-0mm}
\end{figure*}

\subsubsection{Average Reward}
We plot the average reward curves of HDDPG based on the three SNN training frameworks with different time steps and original DDPG in the four training environments respectively, as shown in Fig. \ref{fig:8}. The average reward is calculated by averaging all rewards in a period. Specifically, the average reward of HDDPG-STBP in training environment \#1/\#2/\#3 is similar to that of original DDPG, but in training environment \#4, HDDPG-STBP can achieve higher average reward than original DDPG at the beginning, which means that HDDPG-STBP learns the appropriate navigation policy quickly. The average reward of HDDPG-BPTT is slightly worse than that of original DDPG in training environment \#1/\#2, and similar in training environment \#3. In training environment \#4, HDDPG-BPTT also learns an appropriate navigation policy faster than original DDPG and finally achieves similar average reward. HDDPG-SLAYER performs significantly worse than original DDPG in training environments \#1/\#2/\#3, especially when the time step $T$ is 10. But in training environment \#4, HDDPG-SLAYER can also learn the appropriate navigation policy faster, and finally achieve similar average reward to original DDPG. The above situation shows that in the face of a complex environment, the spatio-temporal dynamic characteristics of SNN can better help the actor network learn a suitable navigation policy. In conclusion the average reward of HDDPG-STBP is close to that of original DDPG and is higher to HDDPG-BPTT and HDDPG-SLAYER.

\subsubsection{Training Time}
\begin{figure}[htbp]
\vspace{-0mm}
\centering
\includegraphics[width=\columnwidth]{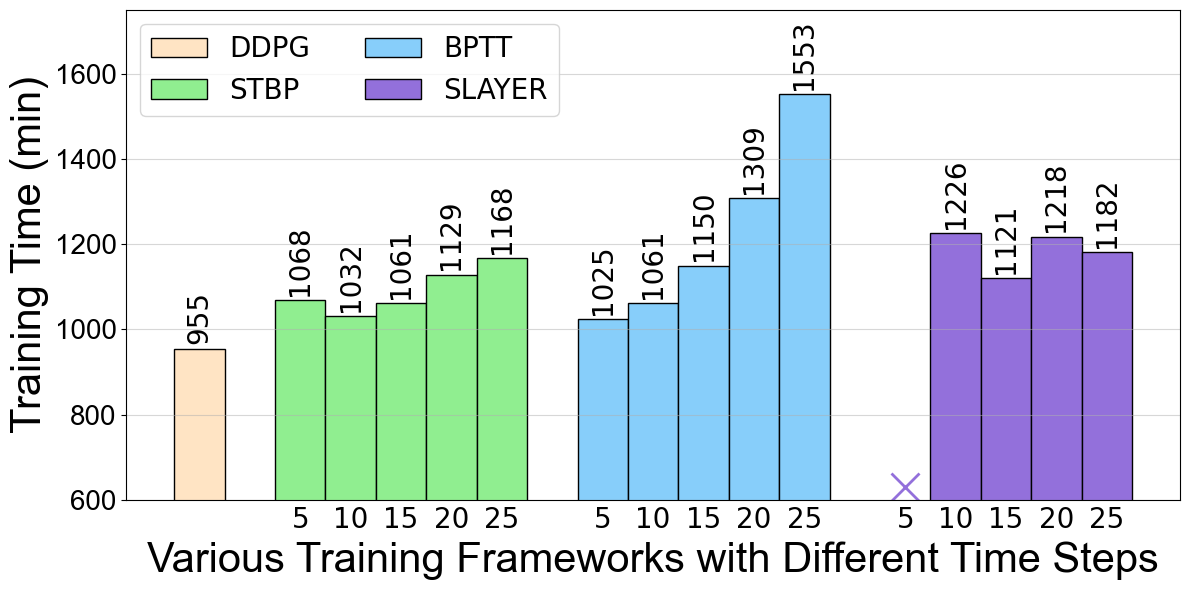}
\caption{Total training time of our proposed HDDPG (with various SNN training frameworks and different time steps) and original DDPG.}
\label{fig:9}
\vspace{-0mm}
\end{figure}

In this experiment, we compare the training time of HDDPG based on the three SNN training frameworks with different steps and original DDPG. The experimental results are shown in Fig. \ref{fig:9}. With the increase of the time step, the training time of HDDPG based on the three SNN training frameworks shows different trends. The total training time of DDPG is the shortest, and the total training time of HDDPG-STBP is the closest to original DDPG. The total training time of HDDPG-STBP decreases with the decrease of the time step, which finally is close to that of original DDPG. For both STBP and BPTT training frameworks, since they use an iterative form similar to recurrent networks in the time dimension, the training time increases with larger time step. Here HDDPG-BPTT is implemented by the Sinabs framework \footnote{Sinabs Website: \url{https://pypi.org/project/sinabs}.}, and the training time increases significantly with the time step. In contrast, the training time of HDDPG-STBP increases less with the time step increasing than that of HDDPG-BPTT. For the SLAYER training framework, as the time step increases, the total training time remains close. This is because the SLAYER framework uses 3D convolution operations to achieve propagation in the time dimension, so the training time of HDDPG-SLAYER is not sensitive to the increase of the time step.

\subsection{Performance Analysis in Evaluation}
In this subsection, we evaluate the navigation success rate, average distance, and average speed of HDDPG based on three SNN training frameworks with different steps and original DDPG in two evaluation environments, and a comprehensive analysis is carried out. In evaluation environment \#1, the maximum allowable height of the MAV is set to 3.1m. In order to adapt to the expansion of the environment, the maximum allowable height of the MAV in the evaluation environment \#2 is set to 3.6m.

\begin{figure}[htbp]
\vspace{-0mm}
\centering
\subfigure[]{\label{fig:10a}\includegraphics[width=\columnwidth]{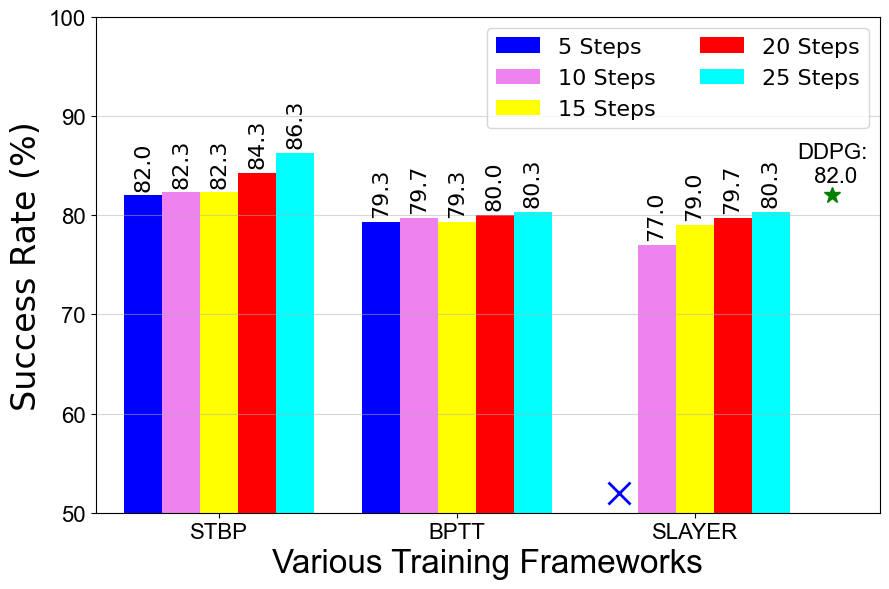}}
\subfigure[]{\label{fig:10b}\includegraphics[width=\columnwidth]{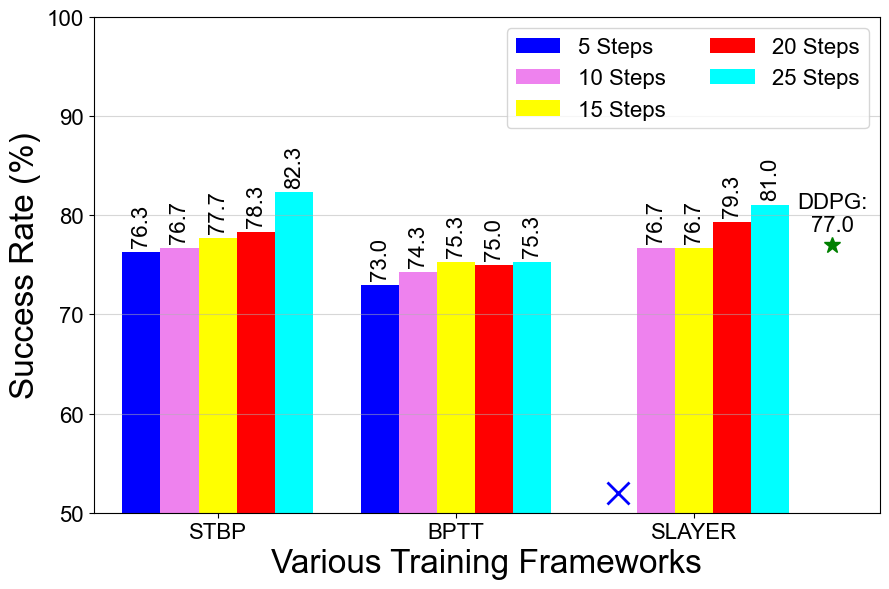}}
\caption{Success rates in evaluation environments of our proposed HDDPG (with various SNN training frameworks and  different time steps) and original DDPG. (a) Evaluation environment \#1. (b) Evaluation environment \#2.}
\label{fig:10}
\vspace{-0mm}
\end{figure}

\begin{figure}[htbp]
\vspace{-0mm}
\centering
\subfigure[]{\label{fig:11a}\includegraphics[width=\columnwidth]{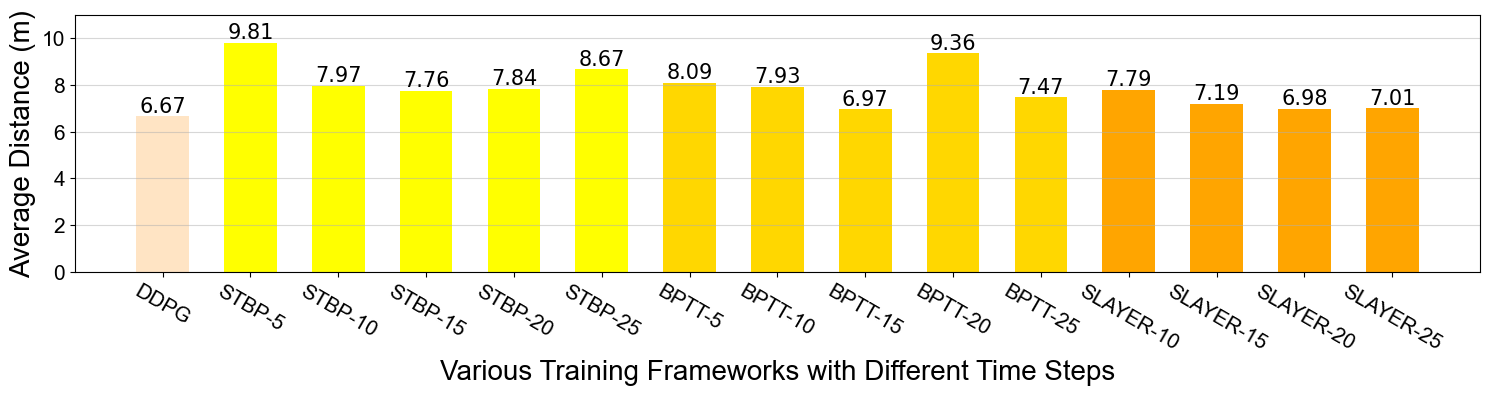}}
\subfigure[]{\label{fig:11b}\includegraphics[width=\columnwidth]{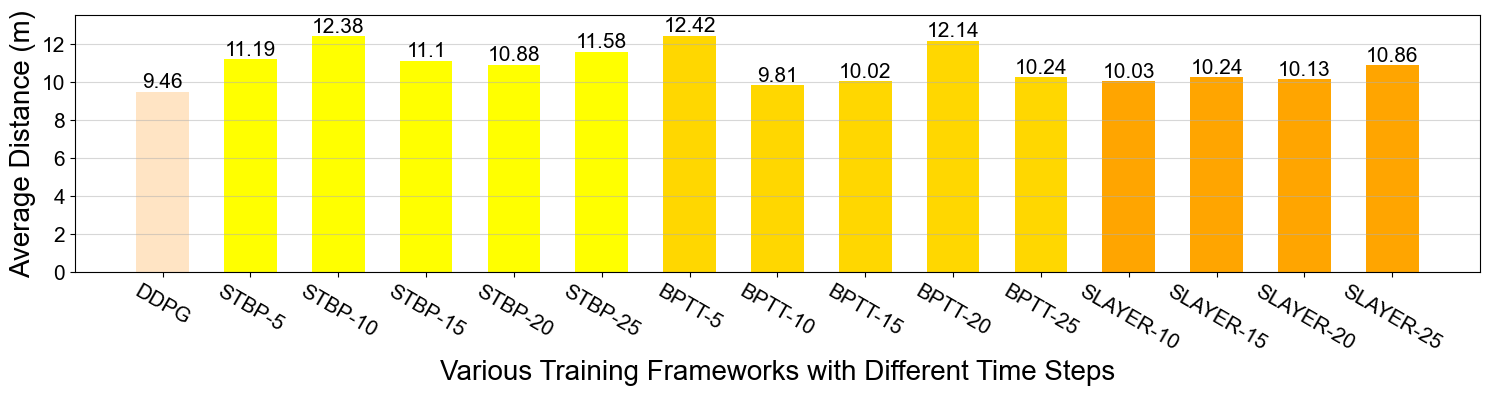}}
\subfigure[]{\label{fig:11c}\includegraphics[width=\columnwidth]{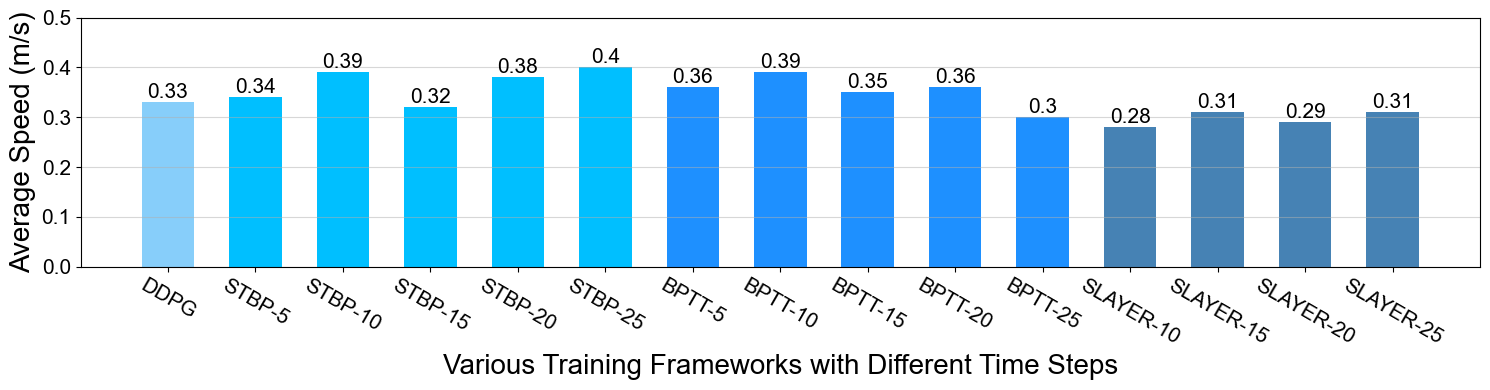}}
\subfigure[]{\label{fig:11d}\includegraphics[width=\columnwidth]{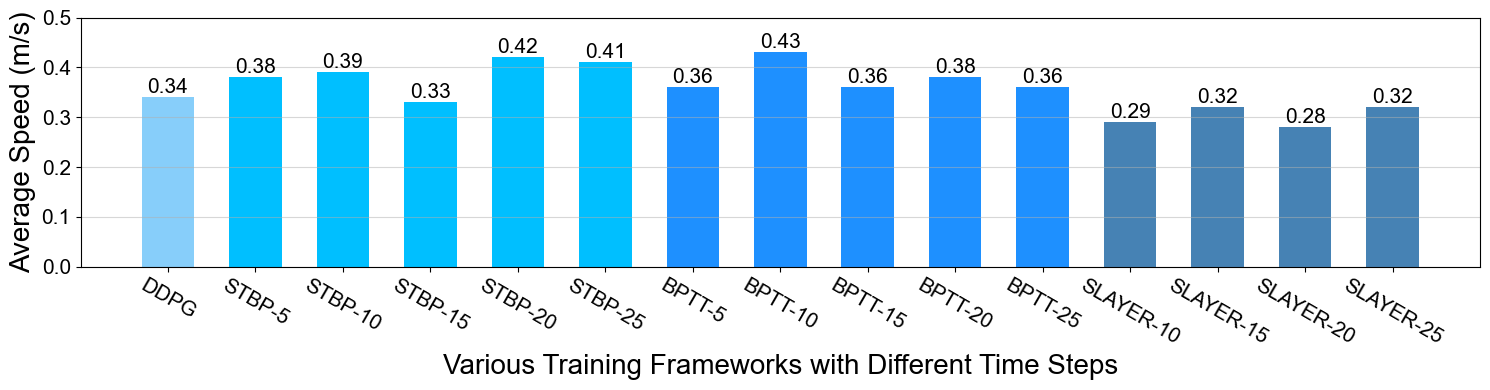}}
\caption{Average distance and average velocity of HDDPG and DDPG by various training frameworks with different time steps. (a) Average distance in evaluation environment \#1. (b) Average distance in evaluation environment \#2. (c) Average speed in evaluation environment \#1. (d) Average speed in evaluation environment \#2.}
\label{fig:11}
\vspace{-0mm}
\end{figure}

\subsubsection{Success Rate}
For the HDDPG algorithm, we select the time steps $\mathrm{T}=5,10,15,20,25$ respectively. The experimental results are shown in Fig. \ref{fig:10}. In order to exclude chance, we repeat the experiment three times for each method and take the average value. The results show that the success rate of HDDPG-STBP, HDDPG-BPTT and HDDPG-SLAYER show a slight increase with the increase of time step in both evaluation environment \#1 and evaluation environment \#2. This is due to higher time step can bring higher control accuracy. In evaluation environment \#1 (as shown in Fig. \ref{fig:10a}), the success rate of DDPG is 82.0\%. Taking this as a benchmark, the success rate of HDDPG-STBP can be significantly higher than that of original DDPG when the time steps are $\mathrm{T}=10,15,20,25$, with a maximum increase of about 4.3\%. Both HDDPG-BPTT and HDDPG-SLAYER are overall slightly lower than the original DDPG benchmark. In evaluation environment \#2, the success rate of original DDPG is 77.0\%. Since evaluation environment \#2 is relatively more complex, original DDPG drops by about 5.0\% compared to evaluation environment \#1 (as shown in Fig. \ref{fig:10b}). Taking this as a benchmark, the success rate of HDDPG-STBP can be significantly higher than that of original DDPG when the time steps are $\mathrm{T}=15,20,25$, with a maximum improvement of about 5.3\%. The success rate of HDDPG-SLAYER is slightly higher than that of original DDPG at the time steps $\mathrm{T}=20,25$, and the highest improvement is about 4.0\%. HDDPG-BPTT is slightly lower than original DDPG as a whole.

\begin{figure*}[htbp]
\vspace{-0mm}
\centering
\subfigure[]{\label{fig:12a}\includegraphics[width=0.245\textwidth]{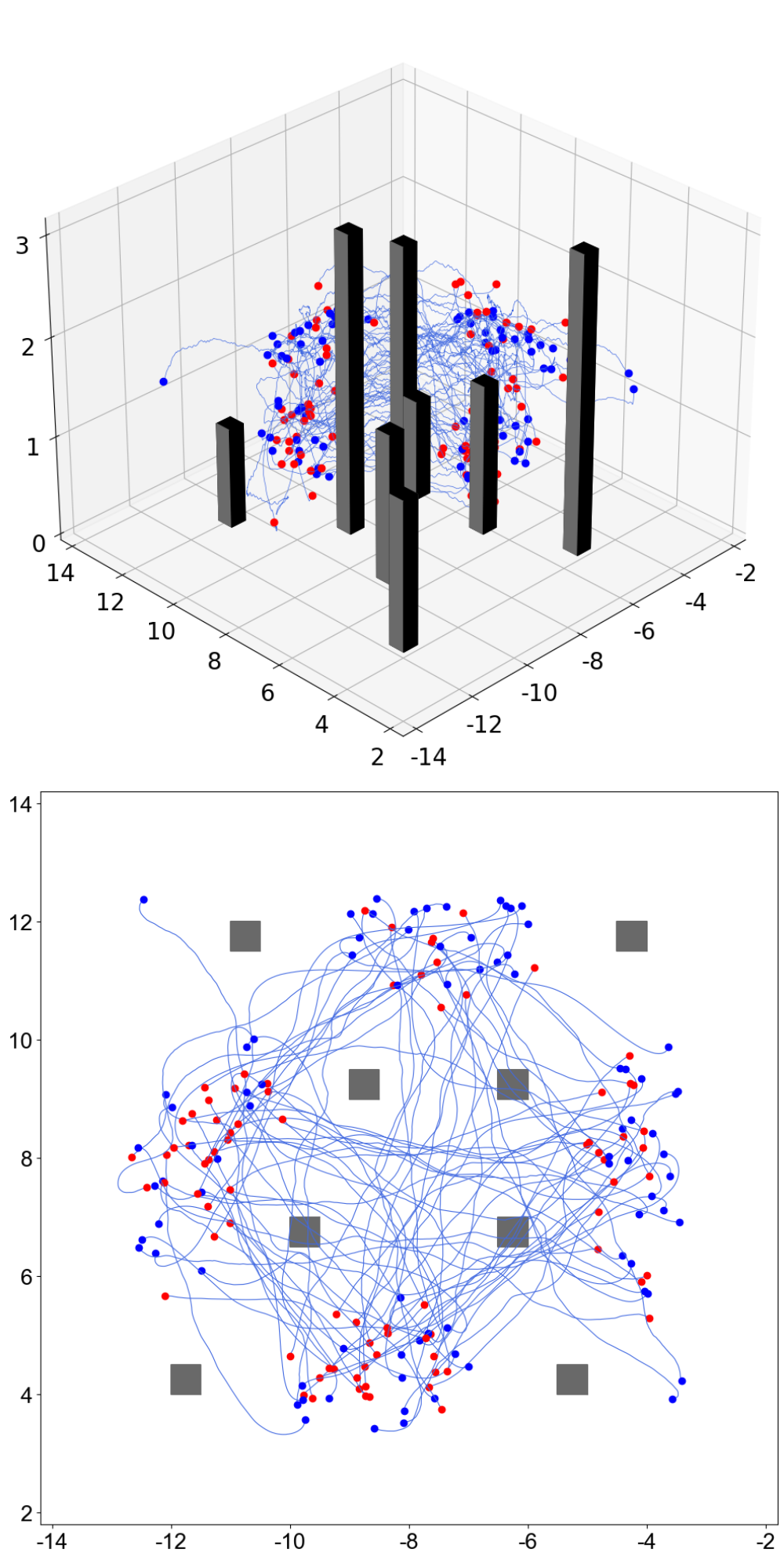}}
\subfigure[]{\label{fig:12b}\includegraphics[width=0.245\textwidth]{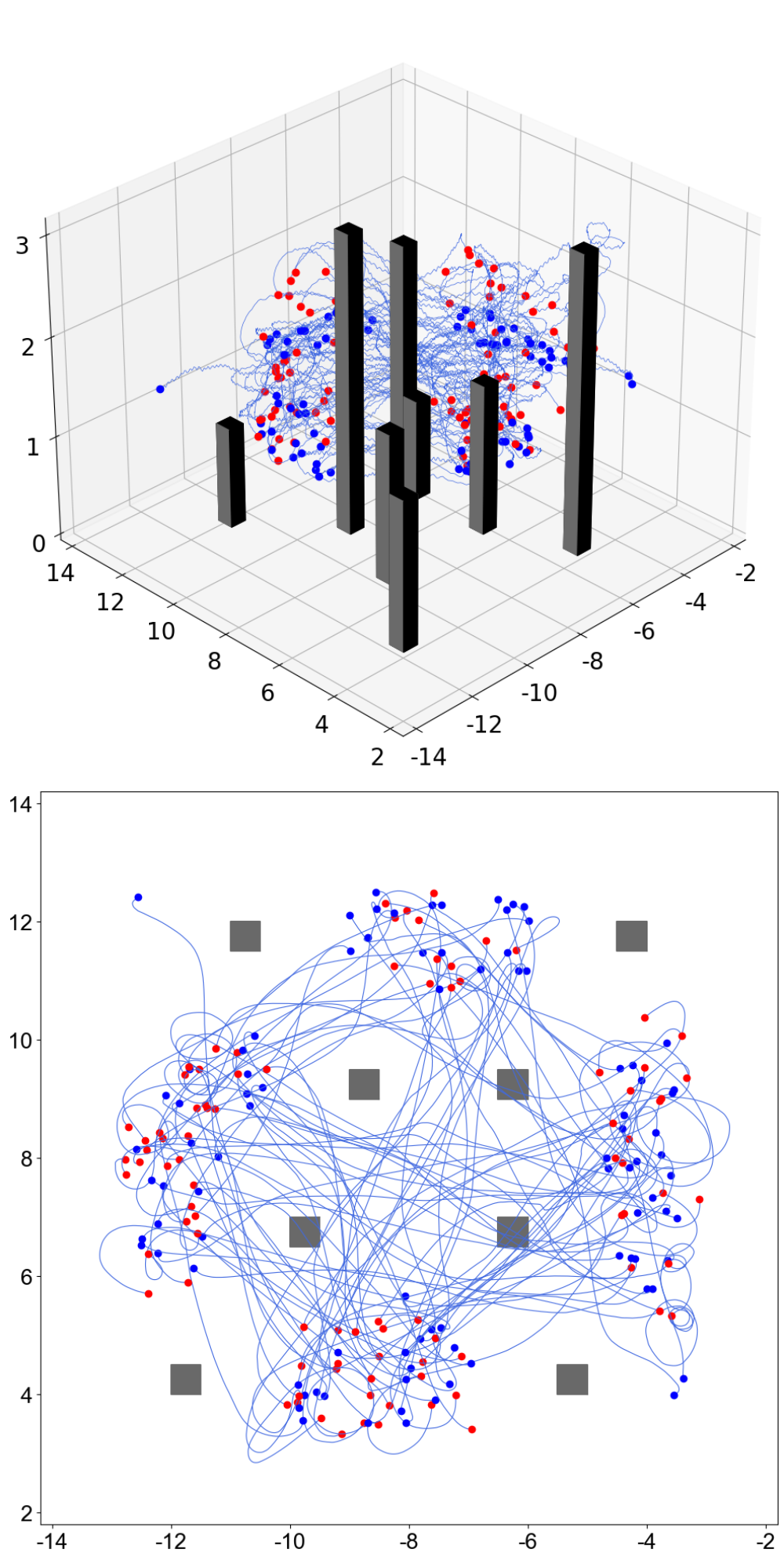}}
\subfigure[]{\label{fig:12c}\includegraphics[width=0.245\textwidth]{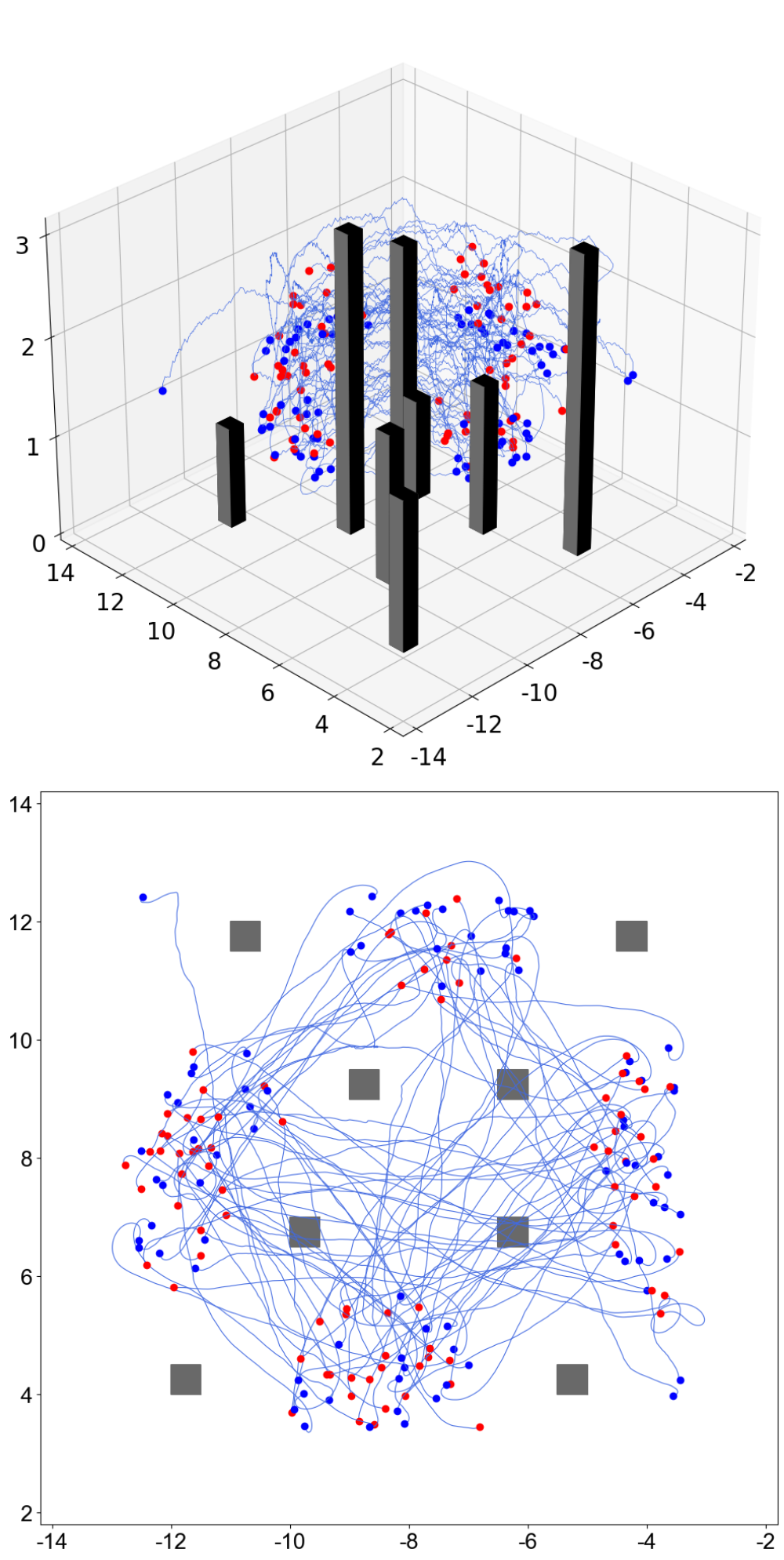}}
\subfigure[]{\label{fig:12d}\includegraphics[width=0.245\textwidth]{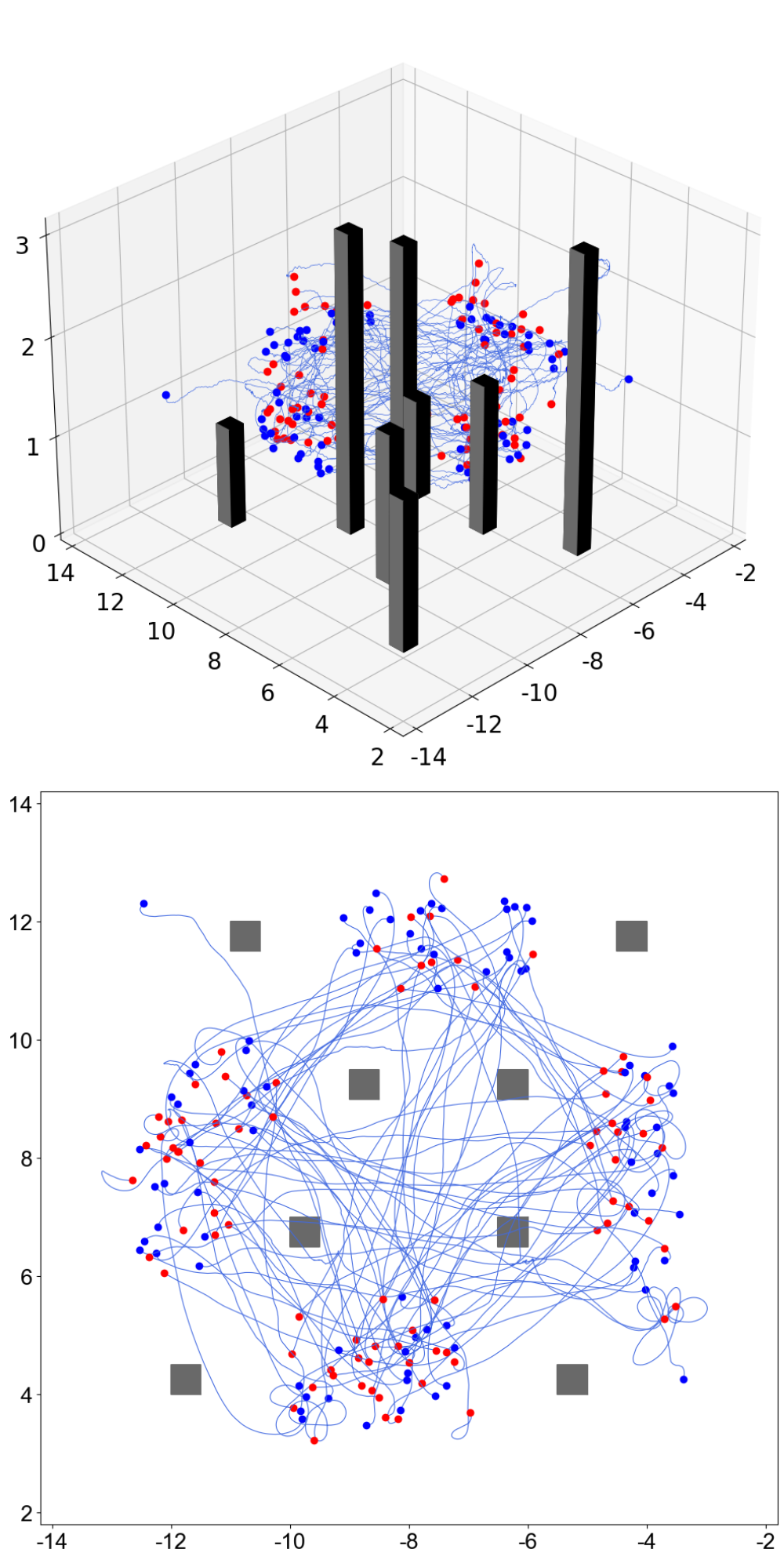}}
\subfigure[]{\label{fig:12e}\includegraphics[width=0.245\textwidth]{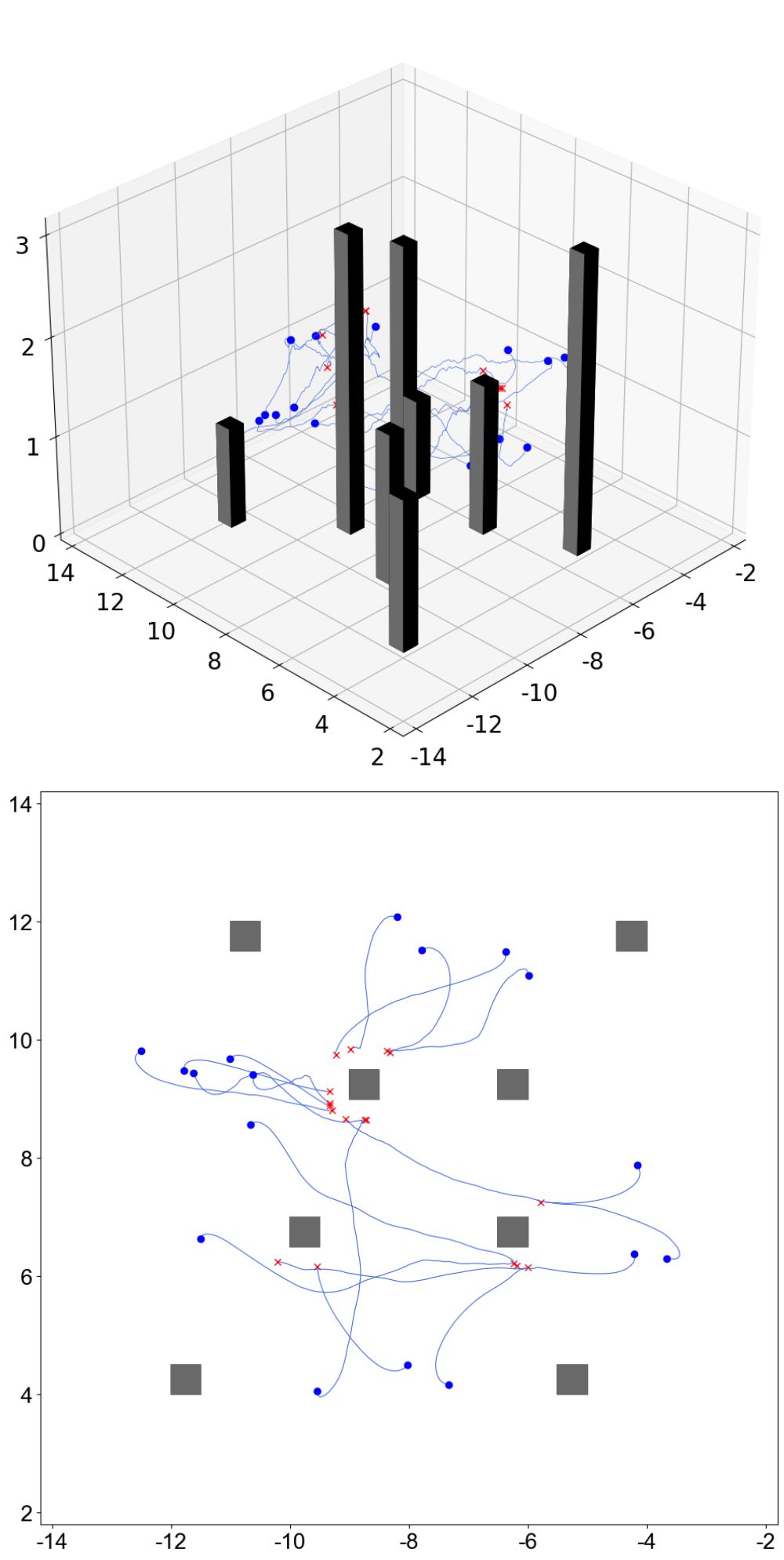}}
\subfigure[]{\label{fig:12f}\includegraphics[width=0.245\textwidth]{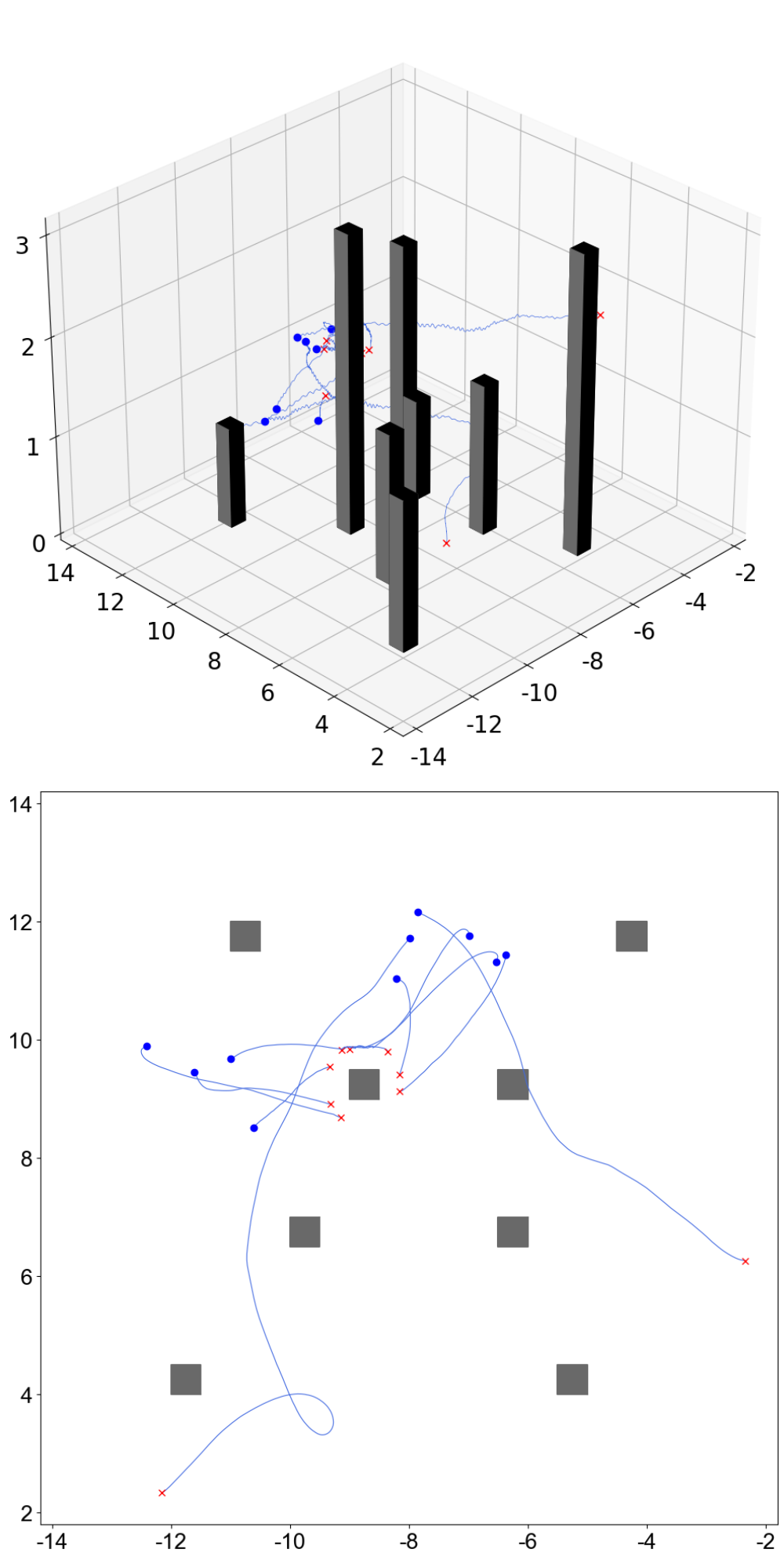}}
\subfigure[]{\label{fig:12g}\includegraphics[width=0.245\textwidth]{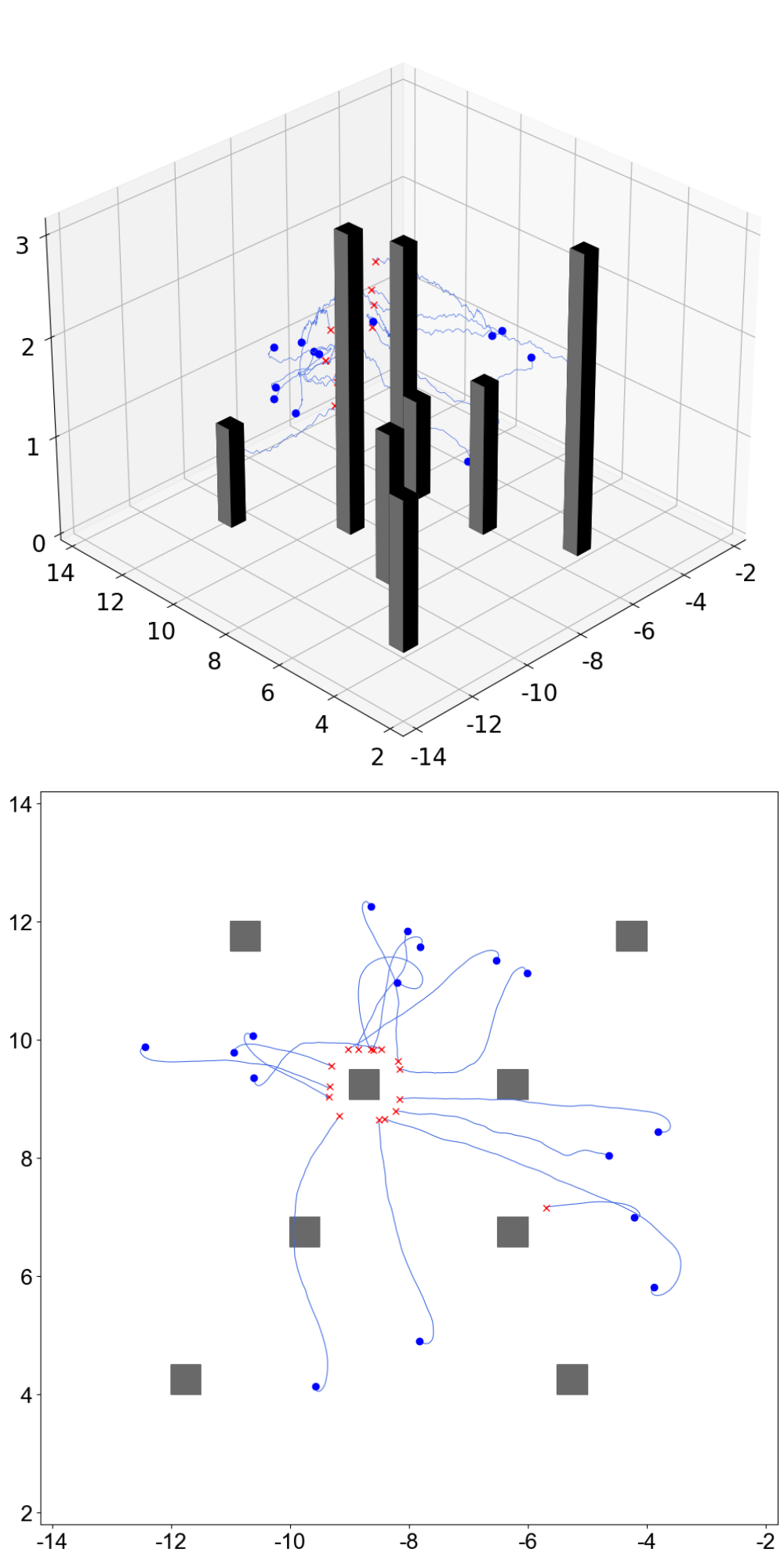}}
\subfigure[]{\label{fig:12h}\includegraphics[width=0.245\textwidth]{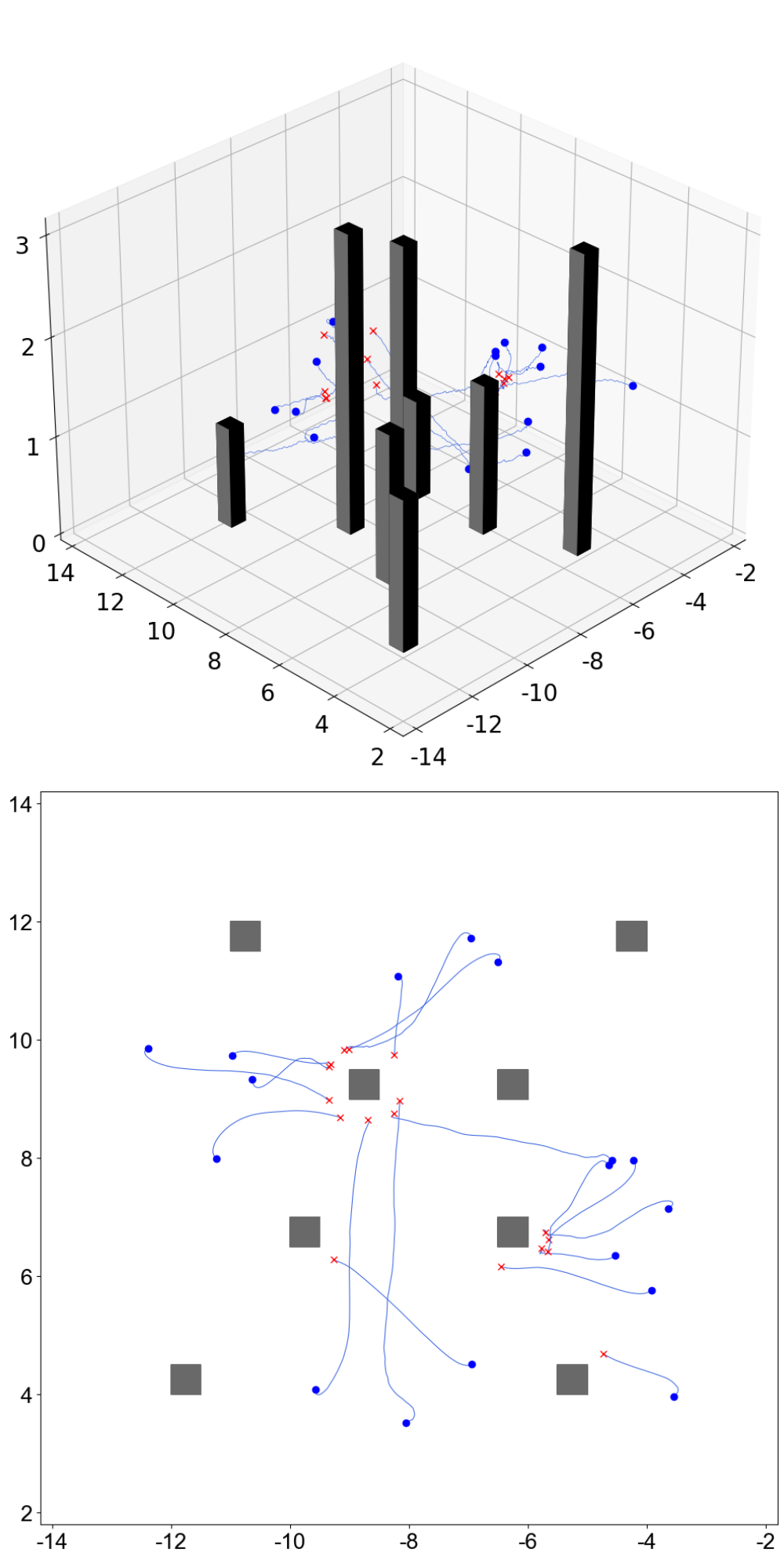}}
\caption{Success and failure trajectories in evaluation environment \#1. (a)-(d): Success trajectories over 200 random start and goal points of the four tained models obtained by original DDPG, HDDPG-STBP, HDDPG-BPTT and HDDPG-SLAYER (3D and top view). (e)-(h): Corresponding failure trajectories (3D and top view). (a) DDPG, success. (b) HDDPG-STBP, success. (c) HDDPG-BPTT, success. (d) HDDPG-SLAYER, success. (e) DDPG, failure. (f) HDDPG-STBP, failure. (g) HDDPG-BPTT, failure. (h) HDDPG-SLAYER, failure.}
\label{fig:12}
\vspace{-0mm}
\end{figure*}

\begin{figure*}[htbp]
\vspace{-0mm}
\centering
\subfigure[]{\label{fig:13a}\includegraphics[width=0.245\textwidth]{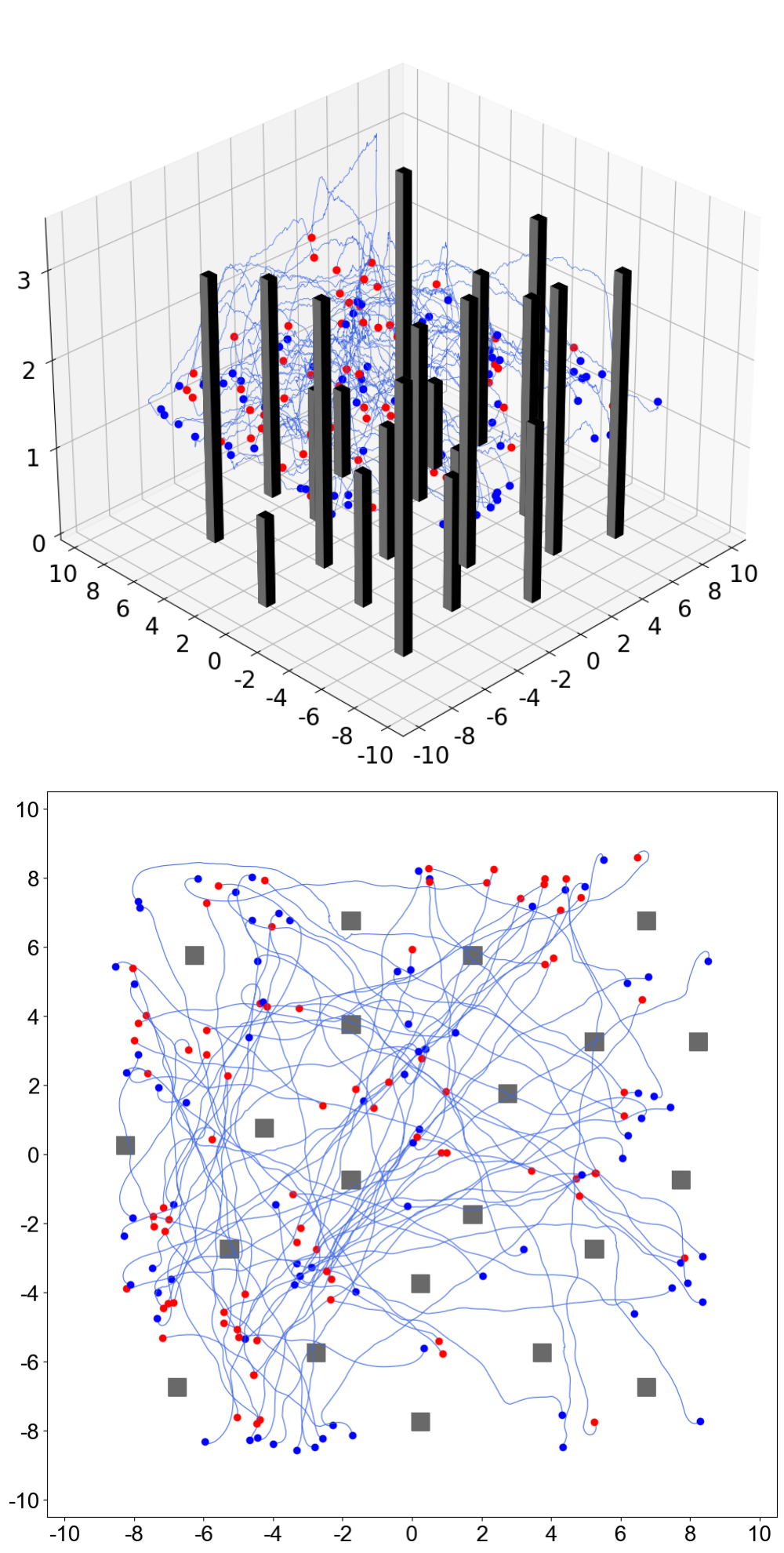}}
\subfigure[]{\label{fig:13b}\includegraphics[width=0.245\textwidth]{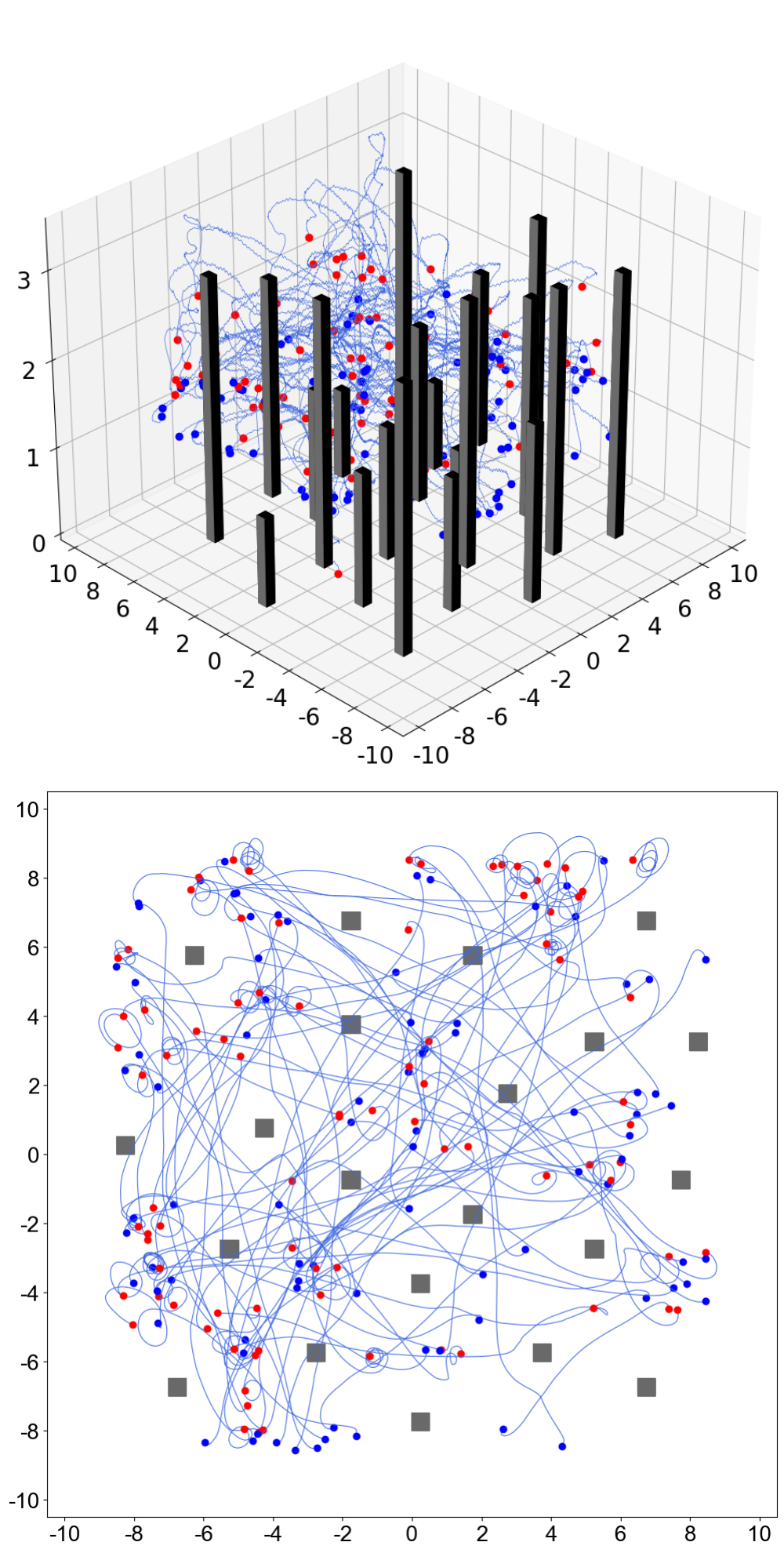}}
\subfigure[]{\label{fig:13c}\includegraphics[width=0.245\textwidth]{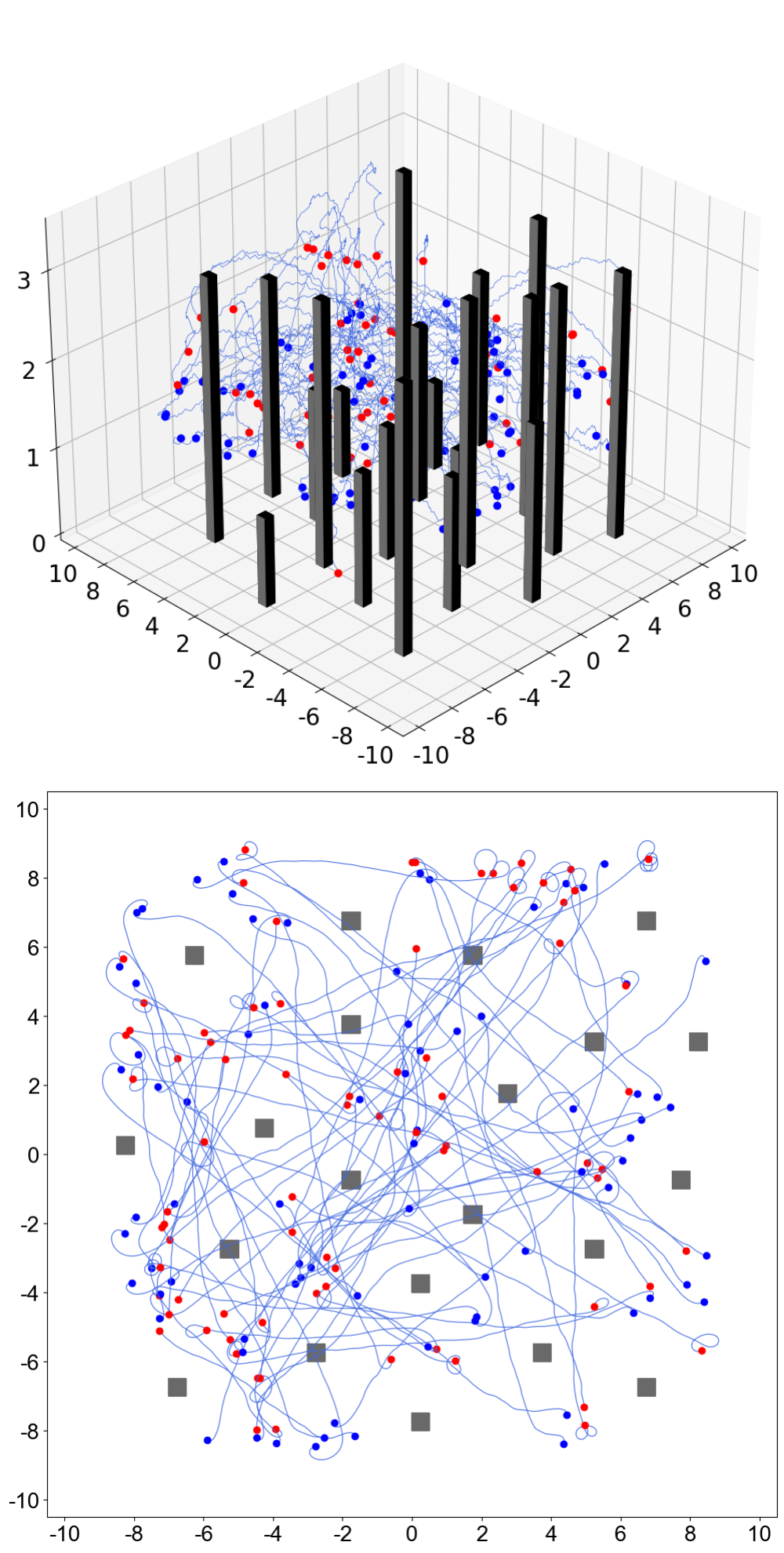}}
\subfigure[]{\label{fig:13d}\includegraphics[width=0.245\textwidth]{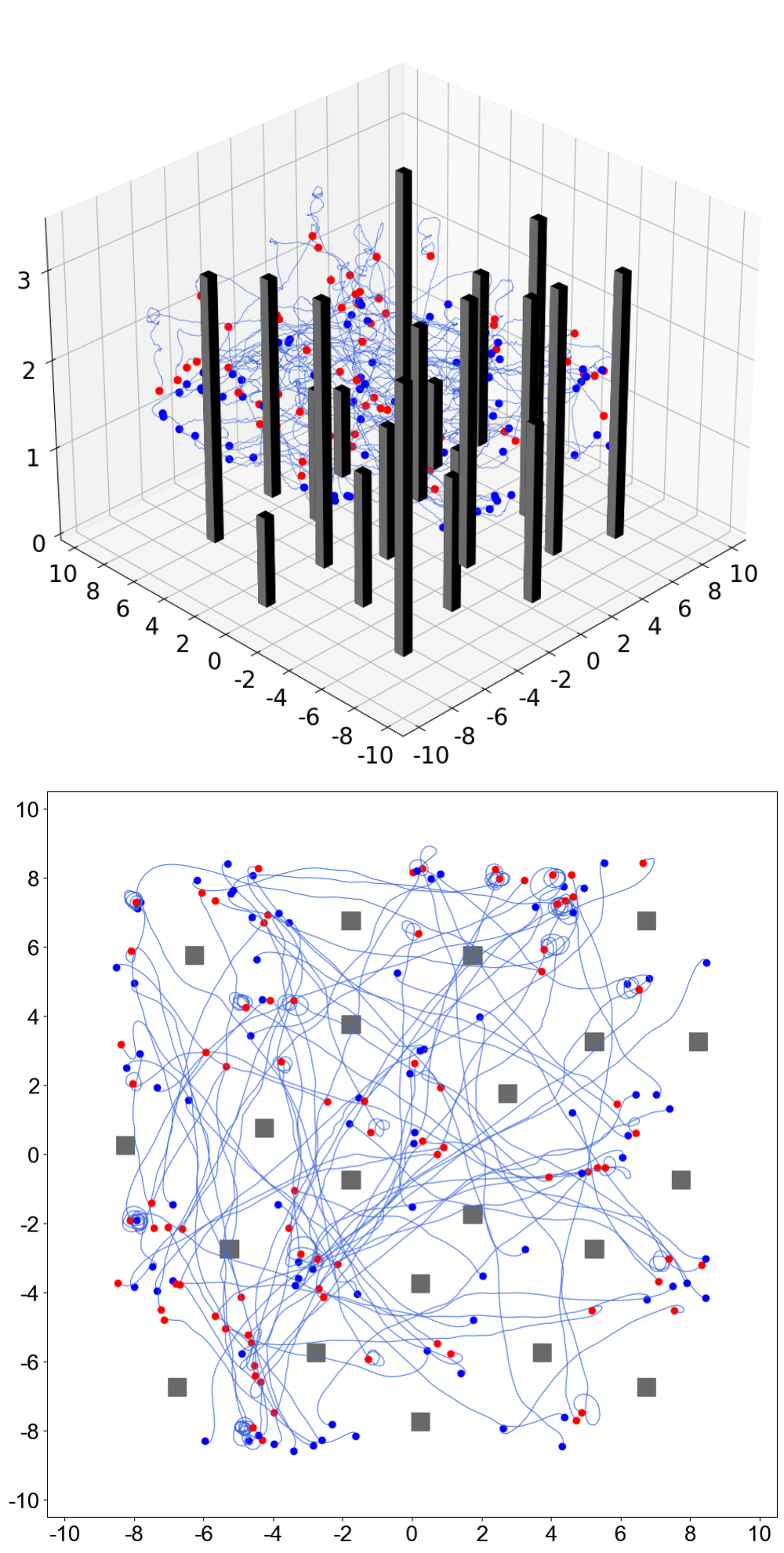}}
\subfigure[]{\label{fig:13e}\includegraphics[width=0.245\textwidth]{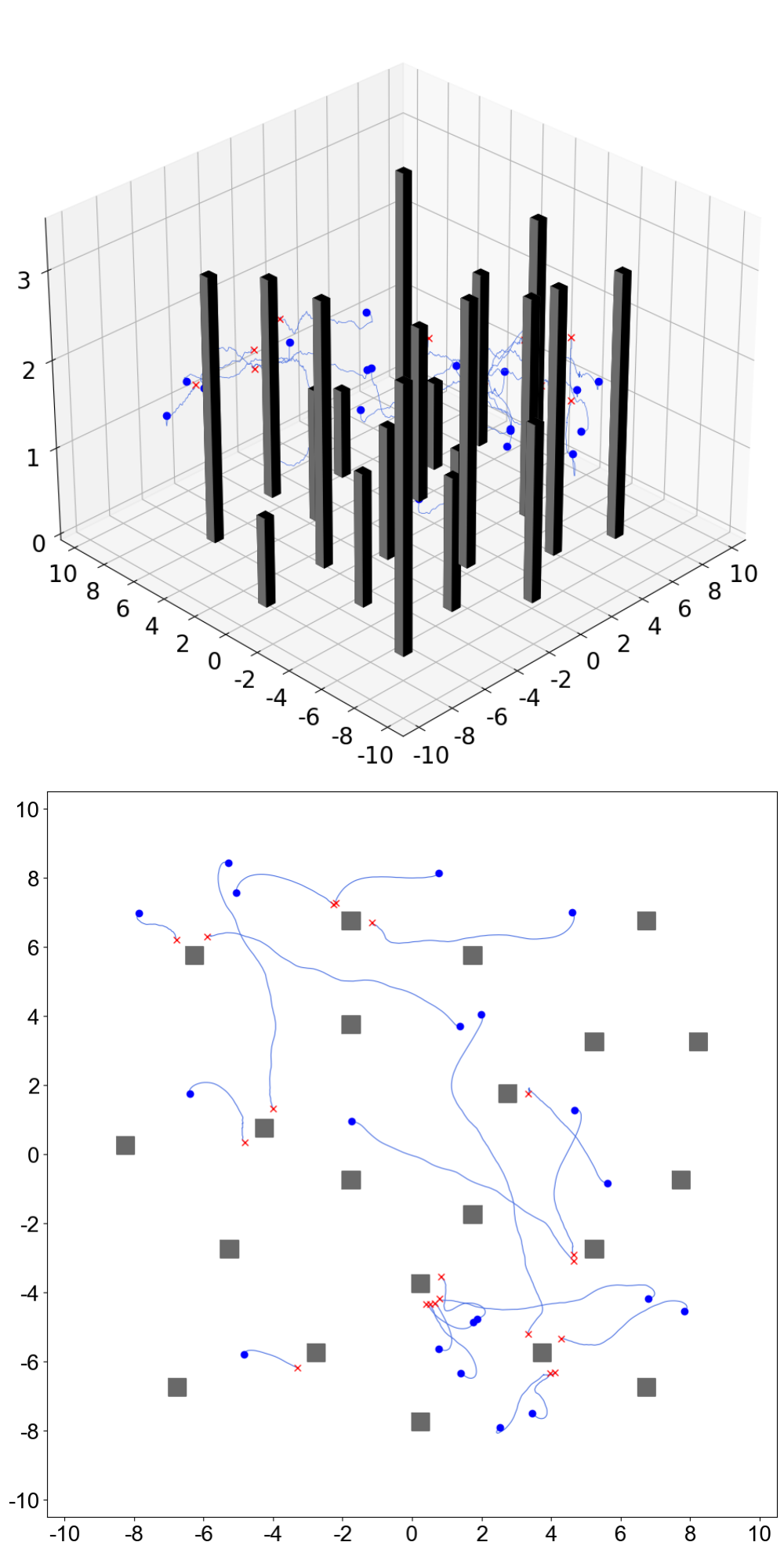}}
\subfigure[]{\label{fig:13f}\includegraphics[width=0.245\textwidth]{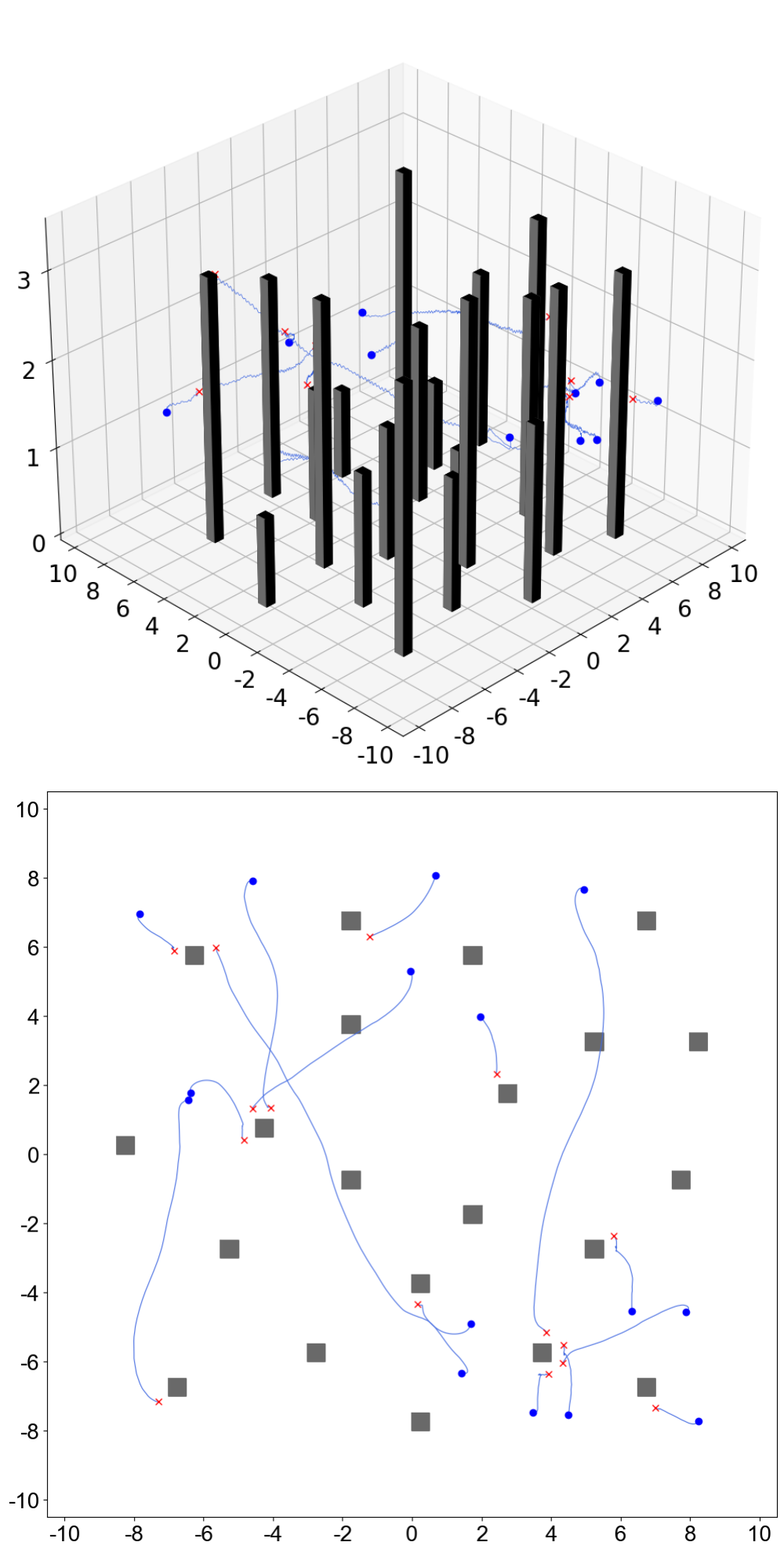}}
\subfigure[]{\label{fig:13g}\includegraphics[width=0.245\textwidth]{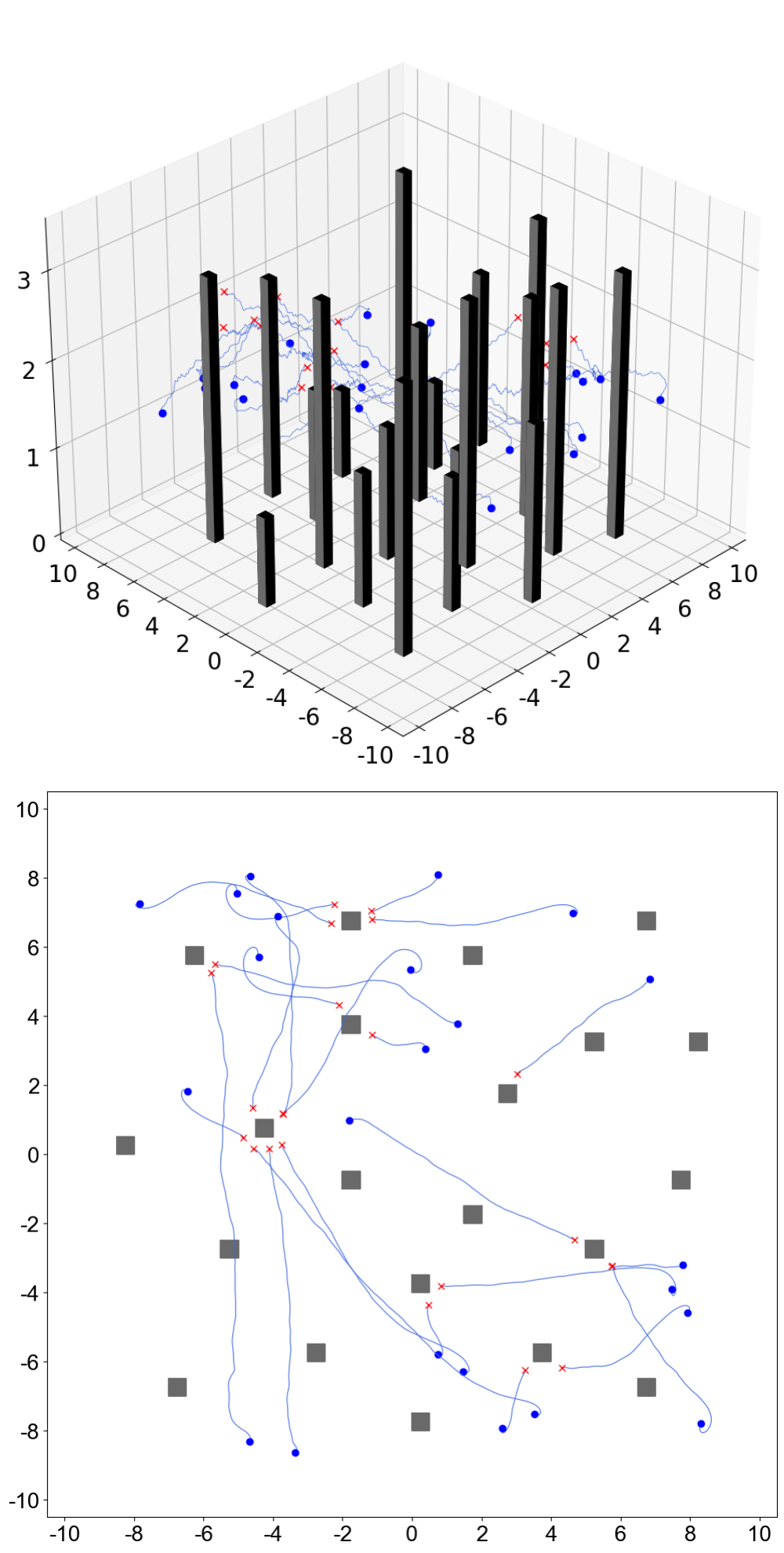}}
\subfigure[]{\label{fig:13h}\includegraphics[width=0.245\textwidth]{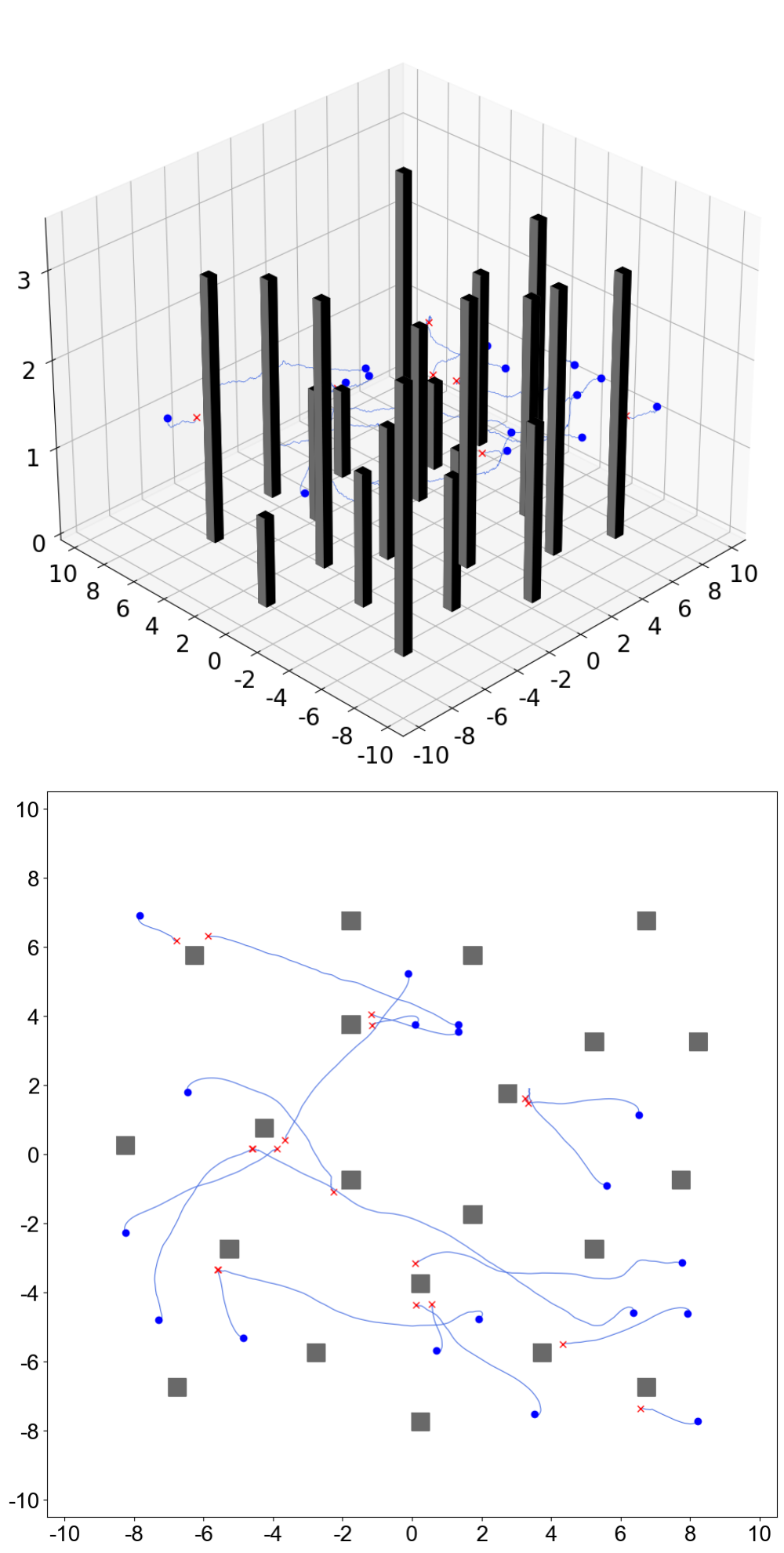}}
\caption{Success and failure trajectories in evaluation environment \#2. (a)-(d): Success trajectories over 200 random start and goal points of the four trained models obtained by original DDPG, HDDPG-STBP, HDDPG-BPTT and HDDPG-SLAYER (3D and top view). (e)-(h): Corresponding failure trajectories (3D and top view). (a) DDPG, success. (b) HDDPG-STBP, success. (c) HDDPG-BPTT, success. (d) HDDPG-SLAYER, success. (e) DDPG, failure. (f) HDDPG-STBP, failure. (g) HDDPG-BPTT, failure. (h) HDDPG-SLAYER, failure.}
\label{fig:13}
\vspace{-0mm}
\end{figure*}

\subsubsection{Average Distance and Average Speed}
The average distance and average speed in evaluation environment \#1 are shown in Fig. \ref{fig:11a} and \ref{fig:11b}. In terms of average distance, the average distance of the path generated by original DDPG is the shortest. The average distance of HDDPG-STBP and HDDPG-BPTT is higher than that of original DDPG, thus the model obtained based on these two frameworks will cause the MAV to fly slightly upward compared with the model obtained by original DDPG. But these two frameworks can have faster average speed. These two frameworks trade off a portion of the distance for faster flight speed. The flying height of HDDPG-SLAYER is basically the same as that of original DDPG, so the average distance of HDDPG-SLAYER is the smallest among the three SNN training frameworks, but the average speed is also the lowest. The average distance and average speed in evaluation environment \#2 are shown in Fig. \ref{fig:11c} and \ref{fig:11d}. As the evaluation environment expanded, the average distance for each method increased. In evaluation environment \#2, the average distance and average speed showed the same regularity as in evaluation environment \#1.

\subsection{Flight Trajectory Analysis in Evaluation}
In this section, we show and analyze the navigation trajectories (both successful and failed trajectories) of each method in two evaluation environments.

In Fig. \ref{fig:12}, we show the success and failure trajectories of the four methods in evaluation environment \#1. In Fig. \ref{fig:12a}-\ref{fig:12d} of the successful trajectories, the highest heights of most trajectories of original DDPG are concentrated between 1.5m-2m, and the highest heights of individual trajectories reach between 2m-2.5m. Most of the trajectories of HDDPG-STBP have the highest heights between 2m and 2.5m, and two trajectories have the highest heights between 2.5m and 3m. The maximum heights of the trajectories of HDDPG-BPTT are concentrated between 2m and 3m. The trajectories of HDDPG-SLAYER are similar to that of original DDPG. In evaluation environment \#1, the trajectories of original DDPG and HDDPG-SLAYER are the most stable, the average maximum height of the trajectories of HDDPG-STBP are slightly higher, and the average maximum height of the trajectories of HDDPG-BPTT are the highest. In the failure trajectories in Fig. \ref{fig:12e}-\ref{fig:12h}, the reasons for the failure of each method’s trajectories due to collisions are similar: (1) The visual sensor carried by the MAV is depth camera. Compared with lidar, the field of view of the depth camera is smaller and only forward-looking, so it is easy to turn and collide without perceiving the obstacle; (2) the global geometric information is lacking, so that the trajectory is close to the obstacle sometimes. So the MAV lack enough space to avoid obstacles.

In Fig. \ref{fig:13}, we show the success and failure trajectories of the four methods in evaluation environment \#2. Due to the enlargement of the whole environment, each method has the situation that the maximum height of the trajectories rise. In Fig. \ref{fig:13a}-\ref{fig:13d} of the successful trajectories, the highest heights of most trajectories of original DDPG and HDDPG-SLAYER are concentrated between 2m-2.5m, and the highest heights of individual trajectories will reach between 2.5m-3m, The highest heights of the trajectories of HDDPG-STBP and HDDPG-BPTT are concentrated between 2.5m-3m, and there are very few trajectories with the highest heights appearing between 3m-3.5m. In evaluation environment \#2, the trajectories of original DDPG and HDDPG-SLAYER are the most stable, and the average maximum heights of the trajectories of HDDPG-STBP and HDDPG-BPTT is slightly higher. In the failure trajectories Fig. \ref{fig:13e}-\ref{fig:13h}, besides the reasons mentioned above, there is another reason that the increase in size and number of obstacles in evaluation environment \#2 makes the MAV face more complex situations than the training environment. Faced with individual states, it is difficult for the actor network to map to appropriate actions, resulting in collisions.

\section{Conclusions}
\label{sec:conclusions}
In this paper, we propose a neuromorphic method combining deep reinforcement learning and spiking neural network, which is applied to the MAV 3D visual navigation for the first time and achieves comparable navigation performance. Our neuromorphic reinforcement learning framework adopts actor-critic network architecture. Combining with the deep critic network, we train the spiking actor network directly by STBP, BPTT and SLAYER three training frameworks respectively. Then, we evaluate the impact of different time steps of three training frameworks. The experimental results show that in two unfamiliar evaluation environments, the spiking actor network trained with the STBP framework can achieve a success rate better than that of the artificial neural network. The spiking actor network shows better robust trained with the SLAYER framework. Although our method can successfully achieve navigation in unfamiliar environments, but due to the lack of global geometric information that the map can provide in traditional navigation algorithms, it is difficult for MAV to achieve a global path that satisfies geometric constraints. In the future, we will consider adding geometric information to the reward function to achieve a better flight path and using a lower-level control method to enable the MAV to achieve higher-speed navigation. Furthermore, we will try to combine neuromorphic vision sensors (such as event cameras \cite{gallego2020event} \cite{kong2022event} \cite{wu2021novel}) to enable neuromorphic 3D visual navigation of MAV.

\bibliographystyle{IEEEtran}
\bibliography{mybibfile}

%\begin{thebibliography}{1}

%\bibitem{ams}
%{\it{Mathematics into Type}}, American Mathematical Society. Online available: 

%\bibitem{oxford}
%T.W. Chaundy, P.R. Barrett and C. Batey, {\it{The Printing of Mathematics}}, Oxford University Press. London, 1954.

%\bibitem{lacomp}{\it{The \LaTeX Companion}}, by F. Mittelbach and M. Goossens

%\bibitem{mmt}{\it{More Math into LaTeX}}, by G. Gr\"atzer

%\bibitem{amstyle}{\it{AMS-StyleGuide-online.pdf,}} published by the American Mathematical Society

%\bibitem{Sira3}
%H. Sira-Ramirez. ``On the sliding mode control of nonlinear systems,'' \textit{Systems \& Control Letters}, vol. 19, pp. 303--312, 1992.

%\bibitem{Levant}
%A. Levant. ``Exact differentiation of signals with unbounded higher derivatives,''  in \textit{Proceedings of the 45th IEEE Conference on Decision and Control}, San Diego, California, USA, pp. 5585--5590, 2006.

%\bibitem{Cedric}
%M. Fliess, C. Join, and H. Sira-Ramirez. ``Non-linear estimation is easy,'' \textit{International Journal of Modelling, Identification and Control}, vol. 4, no. 1, pp. 12--27, 2008.

%\bibitem{Ortega}
%R. Ortega, A. Astolfi, G. Bastin, and H. Rodriguez. ``Stabilization of food-chain systems using a port-controlled Hamiltonian description,'' in \textit{Proceedings of the American Control Conference}, Chicago, Illinois, USA, pp. 2245--2249, 2000.

%\end{thebibliography}

\begin{IEEEbiography}[{\includegraphics[width=1in,height=1.25in,clip,keepaspectratio]{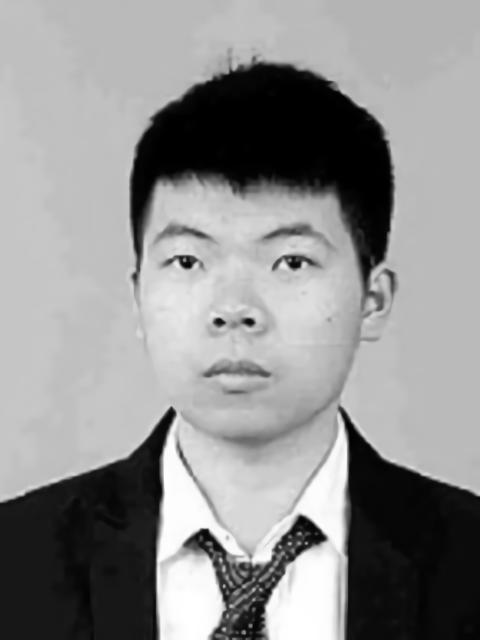}}]{Junjie Jiang}
received the B.S. degree in automation from Northeastern University at Qinhuangdao, China, in 2020. He is currently pursuing the M.S. degree in robot science and engineering with Northeastern University, China. His research interests include spiking neural network, robot visual navigation, and reinforcement learning.\end{IEEEbiography}

\begin{IEEEbiography}[{\includegraphics[width=1in,height=1.25in,clip,keepaspectratio]{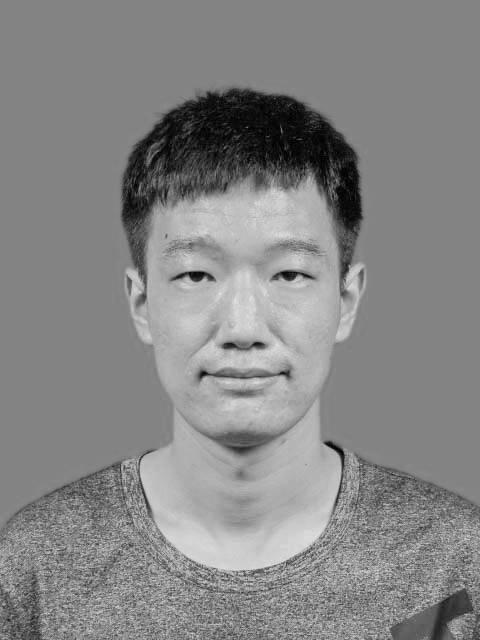}}]{Delei Kong}
received the B.S. degree in automation from Henan Polytechnic University, China, in 2018, and the M.S. degree in control engineering from Northeastern University, China, in 2021. Since 2021, he has been a research assistant with Northeastern University, China. His research interests include event-based vision, robot visual navigation, and neuromorphic computing.\end{IEEEbiography}

\begin{IEEEbiography}[{\includegraphics[width=1in,height=1.25in,clip,keepaspectratio]{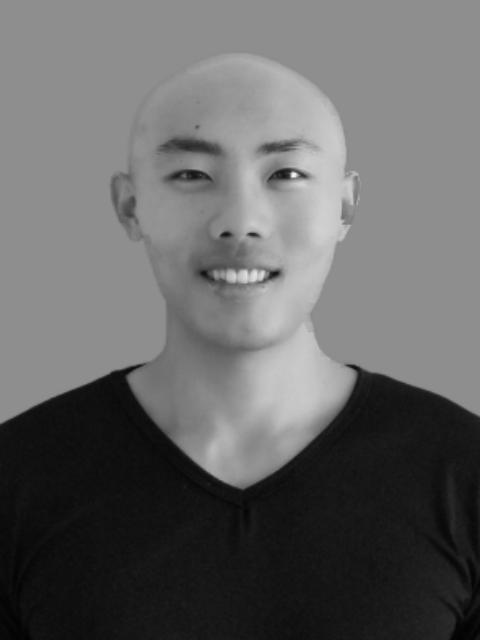}}]{Kuanxu Hou}
received the B.S. degree in robot engineering from Northeastern University, China, in 2020. He is currently pursuing the M.S. degree in robot science and engineering with Northeastern University, China. His research interests include event-based vision, visual place recognition, and deep learning.\end{IEEEbiography}

\begin{IEEEbiography}[{\includegraphics[width=1in,height=1.25in,clip,keepaspectratio]{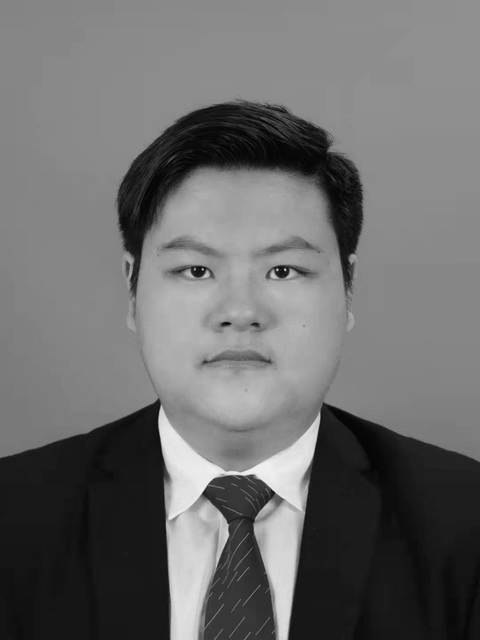}}]{Xinjie Huang}
received the B.S. degree in robot engineering from Northeastern University, China, in 2021. He is currently pursuing the M.S. degree in robot science and engineering with Northeastern University, China. His research interests include event-based vision, visual SLAM, and deep learning.\end{IEEEbiography}

\begin{IEEEbiography}[{\includegraphics[width=1in,height=1.25in,clip,keepaspectratio]{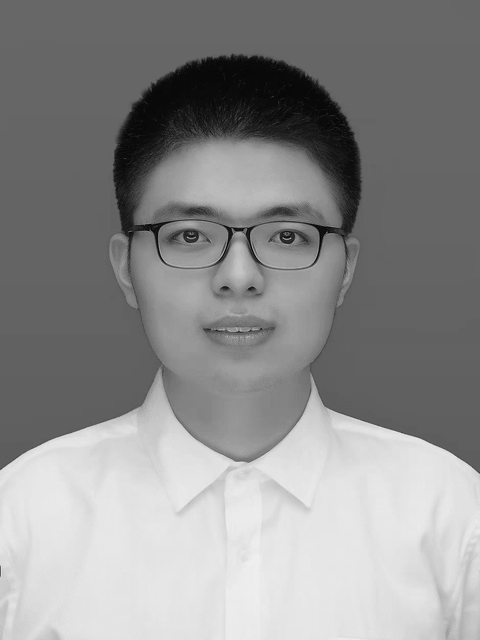}}]{Hao Zhuang}
received the B.S. degree in robot engineering from Northeastern University, China, in 2021. He is currently pursuing the M.S. degree in robot science and engineering with Northeastern University, China. His research interests include event-based vision, visual-inertial odometry, and deep learning.\end{IEEEbiography}

\begin{IEEEbiography}[{\includegraphics[width=1in,height=1.25in,clip,keepaspectratio]{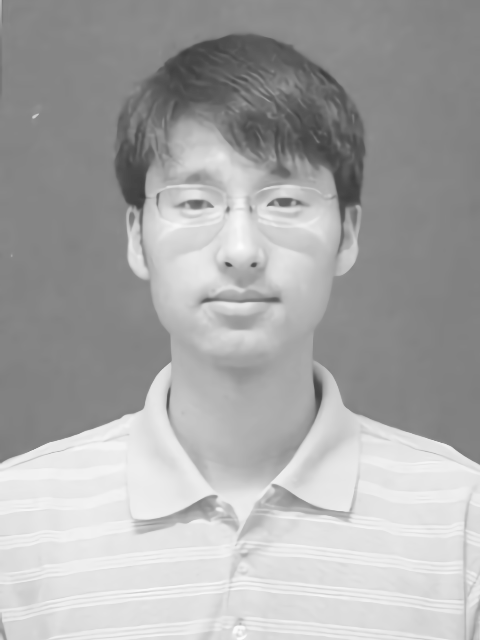}}]{Zheng Fang}
(Member, IEEE) received the B.S. degree in automation and the Ph.D. degree in pattern recognition and intelligent systems from Northeastern University, China, in 2002 and 2006, respectively. He was a Postdoctoral Research Fellow of Carnegie Mellon University from 2013 to 2015. He is currently a Professor with the Faculty of Robot Science and Engineering, Northeastern University. His research interests include visual/laser SLAM, and perception and autonomous navigation of various mobile robots. He has published over 60 papers in well-known journals or conferences in robotics and computer vision, including JFR, TPAMI, ICRA, IROS, BMVC, etc. \end{IEEEbiography}

\end{document}